\documentclass[sigconf]{acmart}%,anonymous,review]{acmart}
\usepackage{graphicx} % Required for inserting images

\usepackage{amsmath}
\usepackage{amsthm} % for \proof{}
\usepackage{listings} % code
\usepackage{makecell}
\usepackage{multirow}

\usepackage{subcaption}
\usepackage{stfloats} % allows [b] placement of figure*

\usepackage{hyperref} % clickable \ref links

\usepackage{placeins}

\usepackage{pgfplots}
\pgfplotsset{compat=1.18}
\usepgfplotslibrary{groupplots}

\acmConference{}{}{}

\newcommand{\code}[1]{\texttt{\small #1}}

\title{Vectorized Dynamic Histograms for Sparse Oblique Forests}

\author{Ariel Lubonja}
\email{ariel@cs.jhu.edu}
\affiliation{
  \institution{Johns Hopkins University}
  \city{Baltimore}
  \state{Maryland}
  \country{USA}
}

\author{Jungsang Yoon}
\email{jyoon72@jh.edu}
\affiliation{
  \institution{Johns Hopkins University}
  \city{Baltimore}
  \state{Maryland}
  \country{USA}
}

\author{Haoyin Xu}
\email{hx@jhu.edu}
\affiliation{
  \institution{Johns Hopkins University}
  \city{Baltimore}
  \state{Maryland}
  \country{USA}
}

\author{Yue Wan}
\email{ywan23@jh.edu}
\affiliation{
  \institution{Johns Hopkins University}
  \city{Baltimore}
  \state{Maryland}
  \country{USA}
}

\author{Yilin Xu}
\email{yxu233@jh.edu}
\affiliation{
  \institution{Johns Hopkins University}
  \city{Baltimore}
  \state{Maryland}
  \country{USA}
}

\author{Richard Stotz}
\email{richardstotz@google.com}
\affiliation{
  \institution{Google}
  \city{Zurich}
  \country{Switzerland}
}

\author{Mathieu Guillame-Bert}
\email{gbm@google.com}
\affiliation{
  \institution{Google}
  \city{Zurich}
  \country{Switzerland}
}

\author{Joshua T. Vogelstein}
\email{jovo@jhu.edu}
\affiliation{
  \institution{Johns Hopkins University}
  \city{Baltimore}
  \state{Maryland}
  \country{USA}
}

\author{Randal Burns}
\email{randal@cs.jhu.edu}
\affiliation{
  \institution{Johns Hopkins University}
  \city{Baltimore}
  \state{Maryland}
  \country{USA}
}

\begin{document}

\begin{abstract}

% \TODO{SPO-GBT results ran with $\sqrt{F}$ instead of $F$! Accuracy is updated. runtime in progress on ydf-1}

Sparse oblique (SPO), part of the top-ranked configuration of Google's Yggdrasil Decision Forests (YDF), improve the accuracy while maintaining interpretability of Random Forests (RF) and Gradient Boosted Trees (GBT) by scanning a sparse linear combination of features instead of a single feature. Because projections are sampled at runtime, training is significantly slower, as pre-run optimizations such as presorting cannot be used.
We overhaul SPO training in YDF, first by fixing inefficiencies in the released version, achieving 2--5$\times$ speedup, then by devising two novel methods, one aimed at GBTs and the other at RFs: (1) Hierarchical AVX2 and AVX-512-vectorized histogram filling speeds up typical GBT-depth trees by up to 1.6$\times$ in SPO-GBT and 2$\times$ for SPO-RF, and (2) runtime-dynamic histograms between vectorized histograms and exact splits, speeds up deeper, RF depth-typical trees (depths 16 to purity) by up to 1.6$\times$ with no detectable effect on accuracy. Our optimizations bring SPO-RF training time on par with YDF's axis-aligned RF. We extensively evaluate on 8 natural and 11 synthetic datasets of up to 10.5 million rows and 1.6 million features, from depth 6 to purity, and open-source the implementation.

\end{abstract}

\maketitle

%% RB GPU implementation is different from RAPIDS CATboost XGboost because it is node based. contribution -- histogramming on the GPU.
%% comment -- per node computation could do more complex things, like MORF.




\section{Introduction}

% Para 1: Both RF and GBT are important even in the modern LLM world (compressed to five sentences).
Gradient Boosted Trees (GBT) and Random Forests (RF) remain critical components in machine learning ensembles. Though tabular foundation models outperform tuned tree ensembles when datasets are smaller than 50k rows and 2000 features~\cite{hollmann_accurate_2025,qu2026tabiclv2,grinsztajn2026tabpfn,tabarena,kong2026tabfm}, GBTs perform best on high-dimensional datasets with millions to billions of rows~\cite{ye2026closer}. RF offers several benefits to GBTs at the expense of inference latency. They are highly parallelizable, robust to hyperparameter selection~\cite{probst2019tunability} and give an unbiased error estimate and feature importances without holding out data, and adding trees does not overfit~\cite{breiman}. Due to their low computation cost, trees have a very strong cost-to-performance ratio, so are heavily used in resource usage-aware deployments~\cite{maier2026hapens}.

Sparse oblique (SPO) splitting further improves the expressivity and accuracy of GBT and RF~\cite{tomita2020sparse}. While RF selects a random $\sqrt{d}$ subset of features at each node and evaluates which feature best separates the data, sparse oblique forests sample $\sqrt{d}$ random projections, consisting of a sparse linear combination of features. The stochasticity and greater expressivity of the projections makes sparse oblique trees robust to noise and on average leads to more balanced trees~\cite{tomita2020sparse}, and its interpretability makes it a strong candidate for biomedical benchmarks~\cite{might2}.

% SPO motivation without the AutoML figure: YDF KDD'23 Sec. 1 (AutoML share), Sec. 3.11 + Appendix C.1 (rank1 template = SPARSE_OBLIQUE for RF and GBT), Sec. 5 (benchmark hp slower to train).
Yggdrasil Decision Forests (YDF)~\cite{ydf} is Google's open-source implementation of decision forests, serving tens of millions of predictions per second at Google~\cite{ydf}, and offers a wide variety of GBT and RF learners, including Sparse Oblique. YDF is integrated in Google's internal AutoML systems, where it is chosen in roughly a quarter of real-world use cases~\cite{ydf}. Sparse oblique splits are in YDF's top-ranked benchmark configuration for both RF and GBT, and ranks above every untuned XGBoost~\cite{chen2016xgboost}, LightGBM~\cite{ke2017lightgbm} and scikit-learn learner~\cite{pedregosa2011scikit}. We include a similar comparison, adding CatBoost~\cite{prokhorenkova2018catboost} (see Table~\ref{tab:auc-vs-libraries-per-dataset-auc}). The YDF authors report that this configuration is significantly slower to train than their defaults (~\cite{ydf}, Table~\ref{tab:train-time-vs-libraries}). 

% Both the improved accuracy and higher runtime is unsurprising, as Sparse Oblique combines sparse feature subsets at runtime, which is equivalent to evaluating more features and feature combinations. - in next paragraph

% Our paper targets YDF as it is, to the best of our knowledge, the fastest existing Sparse Oblique implementation.

% \{\textit{Theme: performance engineering YDF. Separate into Interesting: Dynamic, Vectorization \& other work: HWY, which we adopted \& fixes. Focus on Sparse Oblique. Will say these apply to AA (no presort)}} - improved!! :)

Given the importance of SPO methods in YDF, we study how to reduce or eliminate training time overhead. Unlike a regular, feature axis-aligned (AA) decision tree, where the feature set remains constant throughout training, oblique random projections must be performed at runtime. At each node, projection features are randomly sampled, their linear combination is calculated, and the resulting new ``feature'' is scanned for the best split. There are combinatorially many such splits in the number of features. The evaluation of this larger feature space gives SPO its accuracy benefit, but precludes many optimization strategies such as pre-sorting features for exact splits~\cite{ydf,wright2017ranger}, histogram subtraction~\cite{chen2016xgboost,ke2017lightgbm}, or oblivious trees~\cite{prokhorenkova2018catboost}.

% \{\textit{Ensure writing sounds like Performance overhaul of SPO-YDF, like abstract.}}

% we need to mention (briefly) that Highway sort was added, we adopted it, we made plenty of PRs, and the baseline is highly optimized
A deep performance study of SPO-RF and SPO-GBT revealed bottlenecks during training. We contributed the fixes to YDF (Floyd's projections sampling, caching projection evaluator, a Fisher-Yates shuffle, among others) as a series of pull requests (see Appendix~\ref{app:pull-requests} for full list). Independently, Google replaced \texttt{std::sort} with Highway VQSort~\cite{wassenberg_vectorized_2022}, a highly optimized sorting library that heavily leverages SIMD instructions, which we adopted. Together these set a new baseline for this paper that is 2-5$\times$ faster than the open-source version with which we started. In this baseline, the majority of runtime is spent sorting nodes with {\em exact} splits: Highway sort is faster than every prior method, including all histogram variants and \texttt{std::sort} (Table~\ref{tab:stdsort-vs-highway}). All speedups in this paper are measured against this stronger baseline. Against \texttt{std::sort} previously, dynamic selection alone achieves 2$\times$ (Appendix~\ref{sec:stdsort-vs-highway}). %\{\textit{[low-priority] update fig in appendix w/ HWY \& Vectorized Random}}.

% Dynamic charts look better on EvalProjection only, but there's no way to fairly compare EvalProj to Axis-Aligned, since AA pays the read cost inside EP - it doesn't have a separate ApplyProjection(). Chose NodeTrain
% HIGGS doesn't show benefit w/ HWY
\begin{figure*}[t]
\centering
\newcommand{\placeholderpanel}[1]{
  \fbox{\begin{minipage}[b][0.62\linewidth][c]{\dimexpr\linewidth-2\fboxsep-2\fboxrule\relax}
    \centering\ttfamily\small #1
  \end{minipage}}}
\begin{subfigure}[b]{0.32\textwidth}
  \centering
  \includegraphics[width=\linewidth]{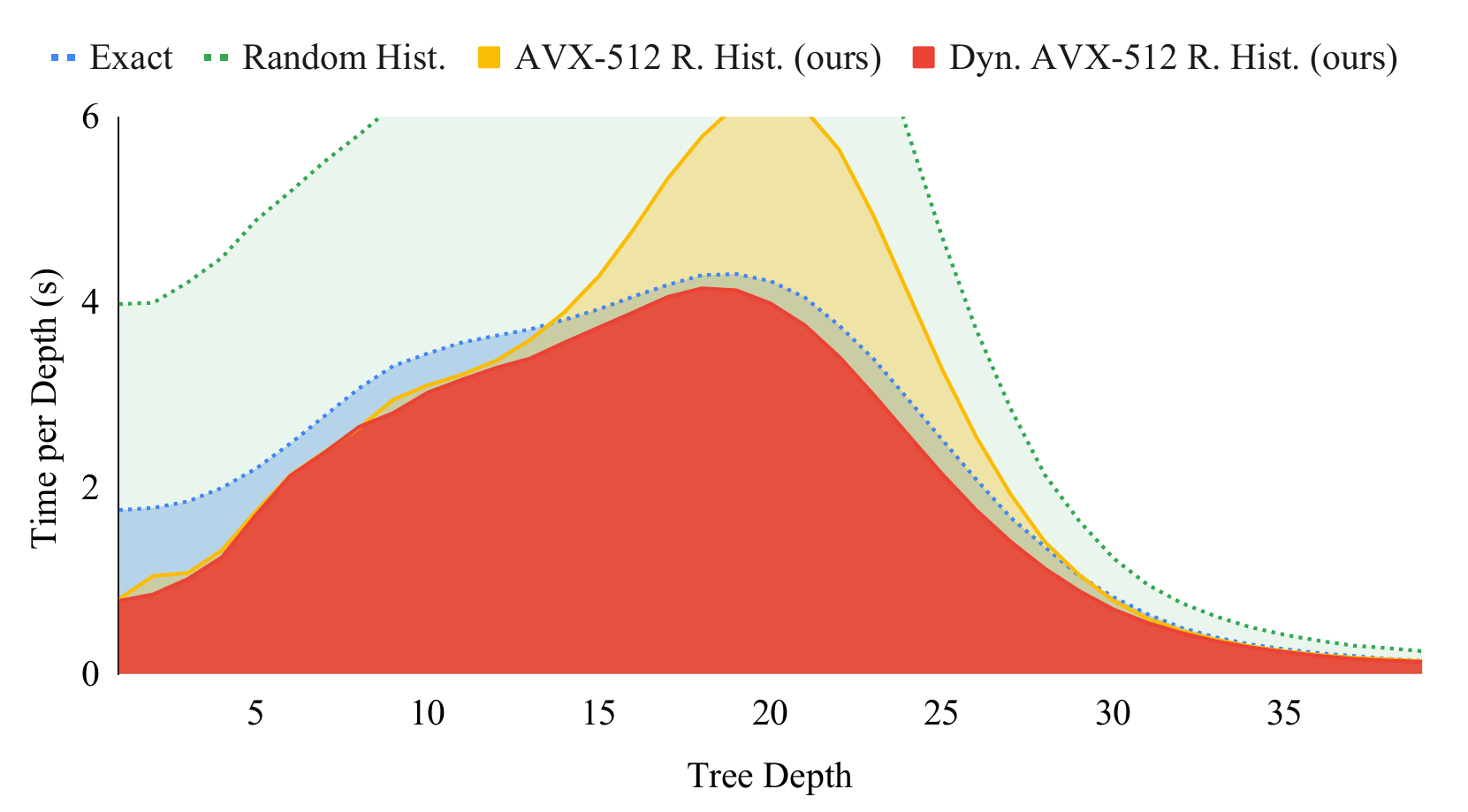}
  \caption{SPO-GBT}
  \label{fig:motivation-nodetrain-gbt}
\end{subfigure}
\hfill
\begin{subfigure}[b]{0.32\textwidth}
  \centering
  \includegraphics[width=\linewidth]{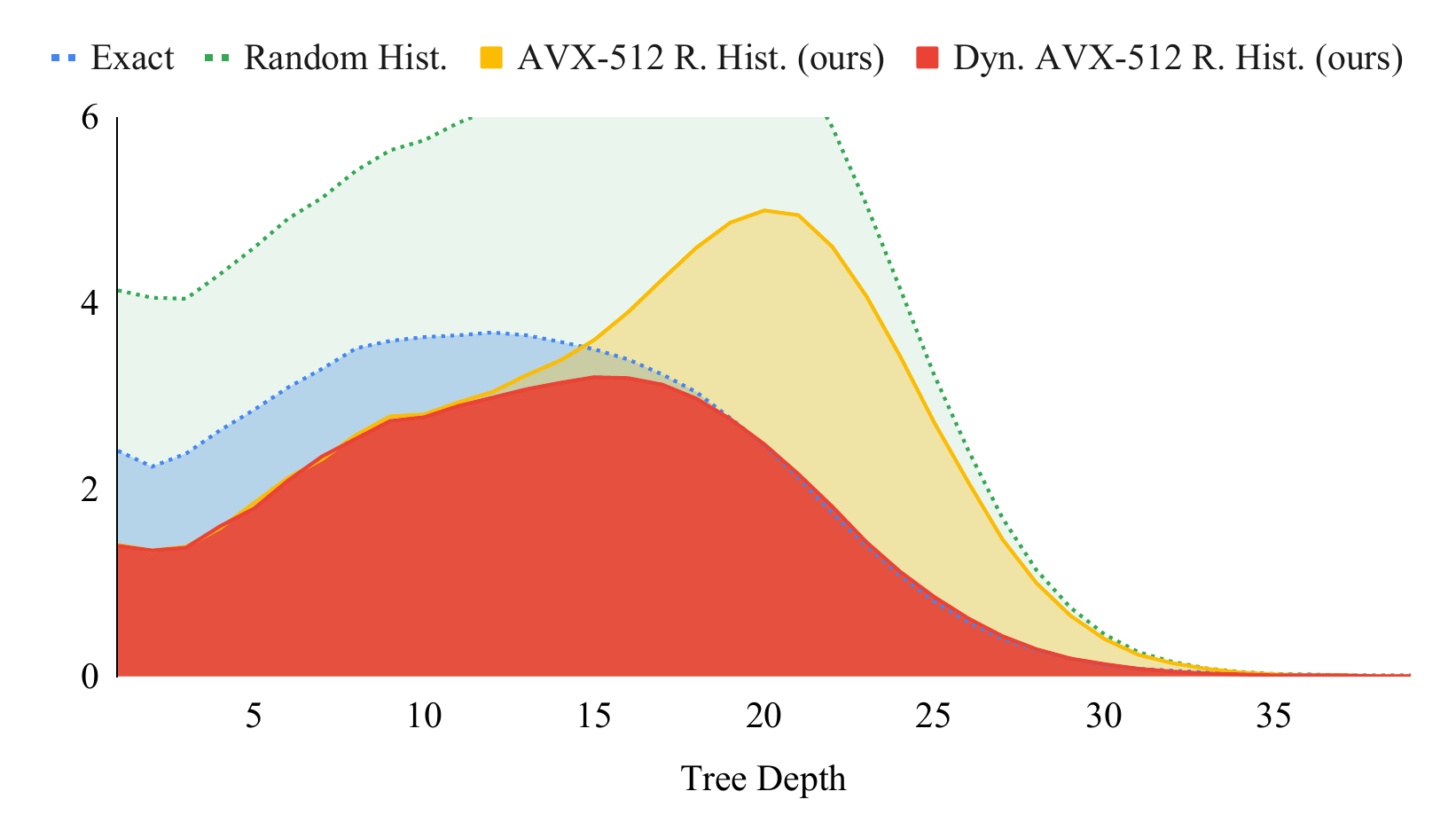}
  \caption{SPO-RF}
  \label{fig:motivation-nodetrain-rf}
\end{subfigure}
\hfill
\begin{subfigure}[b]{0.32\textwidth}
  \centering
  \includegraphics[width=\linewidth]{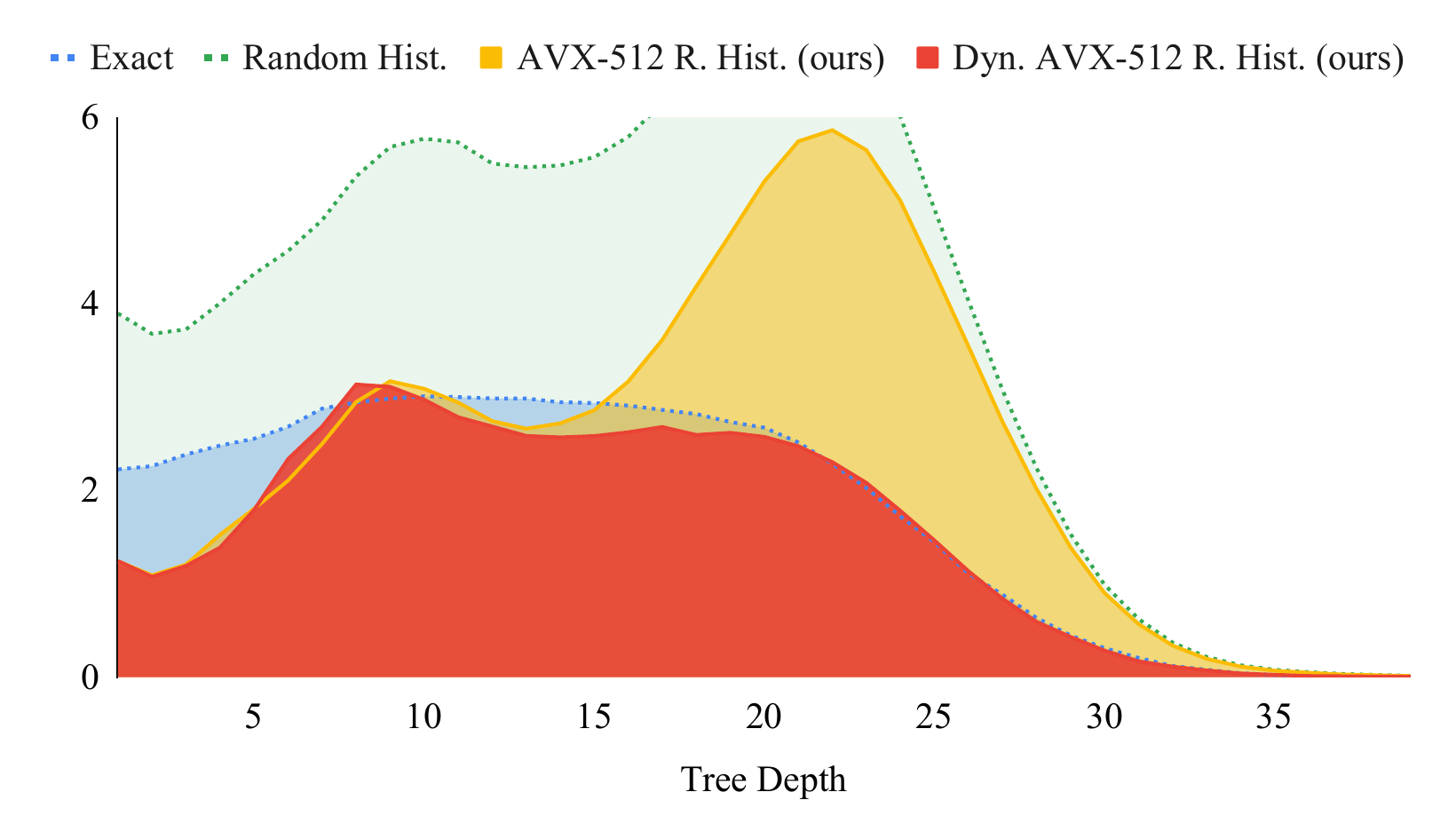}
  \caption{AA-RF}
  \label{fig:motivation-nodetrain-aarf}
\end{subfigure}
\caption{Training time by tree depth on YouTube-8M, one panel per model family: SPO-GBT, SPO-RF and axis-aligned RF (AA-RF). Each panel compares the existing Exact and Random Histogram (256-bins) splitting with our Vectorized and Dynamic-Vectorized (256-bin) methods.}
\Description{Three line charts of \texttt{NodeTrain} time, one per model family, plotting per-depth training time against tree depth for four splitting methods.}
\label{fig:dynamic-motivation-nodetrain}
\end{figure*}

Training time not only depends on tree depth, but is skewed to shallower or deeper levels depending on the splitting method:  very large nodes near the root disproportionately affect the $O(n\log n)$ sorting cost, while large numbers of small nodes punish the $O(k)$ startup cost in histogram's $O(n\log k + k)$ complexity. GBTs are trained to depth 6, often to leverage a fast inference engine, such as QuickScorer~\cite{lucchese_quickscorer_2015}, which has a limit on maximum number of nodes. Training to depth 16-24 is common for Random Forests, and training to purity (full depth) is useful for posterior estimates~\cite{might1,might2}. The two families thus stress opposite ends of the depth range, and we contribute one optimization for each. Based on this observation, we contribute:

\emph{\textbf{(1) Vectorization of histogram construction}}. Histogram construction is asymptotically more efficient than sorting and should be faster when there are a sufficient number of samples. Yet with the powerful Highway sort replacing \texttt{std::sort}, the baseline histogram implementation using \texttt{std::upper\_bound()} is much slower than vectorized Highway sort (Figure~\ref{fig:dynamic-motivation-nodetrain}). We propose a hierarchical vectorized histogram filling implementation that leverages AVX2 and AVX-512, and delivers up to $2\times$ speedup over the SPO-RF Exact and $1.6\times$ over SPO-GBT, and beats scalar Random Hist. at all tree levels, while being mathematically identical.

\emph{\textbf{(2) Runtime-adaptive histograms.}} In deeper trees, the number of active nodes grows near-exponentially, and peaks at around halfway to purity. Though these nodes have few samples in them, histogram approaches must still pay the $O(k)$ startup cost of bin generation and scanning that Exact avoids (Fig.~\ref{fig:dynamic-motivation-nodetrain}). We propose a Dynamic switching method that adaptively chooses between vectorized histograms and exact splits depending on the number of samples at each node, with no detectable effect on accuracy (Table~\ref{tab:accuracy-rf}). This achieves up to 1.6$\times$ speedup when training deeper trees, bringing SPO performance on par with Axis-Aligned in YDF (Table~\ref{tab:train-time-vs-libraries}).

Neither optimization is tied to a depth or a model family. Vectorized filling accelerates every histogram split, and the dynamic switch acts on node size rather than depth, so both apply to GBT and RF alike, and to axis-aligned splits whenever features are not pre-sorted (Figure~\ref{fig:dynamic-motivation-nodetrain}(c)). At depth 6 over millions of rows no node is small enough to switch, so dynamic selection reduces to vectorized histograms, and at purity, all nodes switch. We therefore report each family in the setting it is used in practice, GBT at depth 6 and RF from depth 16 to purity. We provide extensive results on a variety of both natural and synthetic datasets (Table ~\ref{tab:all-datasets}), we sweep the full depth range in Table~\ref{tab:train-time-by-depth}. In addition, we provide a heatmap of predicted speedups by dataset size based on our evaluation (Figure~\ref{fig:speedup-heatmap}). An open-source implementation is available at (anonymized).

% \TODO{treat runtime comparison agianst XGboost and Catboost as apples-to-oranges}

% \TODO{include natural results table in main text (that supports numbers above). they're in Appendix}

%TODO RB Exact is a bad name for the baseline.

% only show three lines Exact, Random Hist and AVX-512.  Maybe rename Exact baseline?  Show results to level 8. State in caption that default for all major packages are 6 (XGBoost, CatBoost, LightGBM).

%and adopting vectorized sorting~\cite{}
%when building. This raiseed the baseline of for this paper by a factor of 2x-5x training time for larger datasets. 
%The first was the construction of linear combinations, which is a vector sum of a sparese scan

%Para 3: nature of SO methods.
%they create a tradeoff.  no class/regression inference time penalty. better accuracy and more robust. downside has been training time

% \{[Low priority] GBT and RF have similar computational properties. They only differ in parallelization. GBT has enough work up to depth 6 to keep CPU busy}

\section{Related Work}

% Para 1 -- oblique trees. Point: the split feature is synthesized at the node.
Oblique decision trees split on linear combinations of features rather than on a single feature. Breiman's original random forest paper proposed Forest-RC, which draws random linear combinations of a few inputs at each node~\cite{breiman}. OC1 searches for hyperplanes by hill climbing with random perturbations~\cite{murthy1994system}, canonical correlation forests derive the projection at each node from CCA~\cite{rainforth2015canonical}, and TAO learns sparse oblique trees by alternating optimization over the node hyperplanes~\cite{carreira2018alternating}, with bivariate trees as the two-feature special case~\cite{kairgeldin_bivariate_2024}. Sparse projection oblique random forests (SPO-RF)~\cite{tomita2020sparse} sample sparse random projections at each node and are implemented in treeple~\cite{treeple} and in YDF~\cite{ydf}, whose sparse oblique splits we optimize. SPO-RF is also the base learner of MIGHT, which trains forests to purity to obtain calibrated posteriors for biomedical screening~\cite{might1,might2}. Common to all of these methods is the fact that the feature being split exists only once the node's projection has been drawn and therefore cannot be presorted/pre-quantized before training.

% Para 2 -- how axis-aligned libraries find splits, and why none of it transfers. Merges the old presort and GPU paragraphs.
Axis-aligned learners find splits either exactly or with histograms. Exact splitting sorts the candidate feature's values within the node or scans a pre-sorted index of the whole column: ranger~\cite{wright2017ranger}, XGBoost's exact method~\cite{chen2016xgboost} and YDF~\cite{ydf} pre-sort, and scikit-learn's forests sort in the node~\cite{pedregosa2011scikit}. Histogram splitting, the default in LightGBM~\cite{ke2017lightgbm}, CatBoost~\cite{prokhorenkova2018catboost}, XGBoost's histogram method~\cite{chen2016xgboost}, TF Boosted Trees~\cite{ponomareva2017tf} and Spark MLlib~\cite{meng2016mllib}, quantizes every feature once before training into a fixed number of bins, so that filling a node's histogram is a scatter-add over stored bin indices with no search in the loop. XGBoost's local proposal variant~\cite{chen2016xgboost} and YDF's histogram splits~\cite{ydf} instead recompute bin boundaries per node, from the node's quantiles, its value range, or at random. Further savings come from sharing work across nodes: histogram subtraction derives one child's histogram from the parent's~\cite{ke2017lightgbm}, CatBoost's oblivious trees split every node of a level on the same feature and bin grid~\cite{prokhorenkova2018catboost}, and GPU trainers assemble partial histograms per thread block~\cite{wenthundergbm19,xu2025scalagbm}. 

% Para 4 -- SIMD bin lookup is known (PANDA, DB range partitioning); state exactly what is ours.
% Para 4 -- SIMD bin lookup is known; state exactly what is ours.
Resolving a key against a register of boundaries with one compare and a population count is established: per tree level in $k$-ary search~\cite{schlegel2009k} and in cache-resident range indexes~\cite{polychroniou_comprehensive_2014}, and in PANDA as a two-level SIMD scan over histogram boundaries~\cite{patwary_panda_2016}. All three build their boundaries once, then bin everything against them. Sparse oblique splits draw new boundaries for every projection at every node, so the structure is rebuilt per histogram and may serve only a few hundred lookups. Our lookup is sized for this: $B = W^2$ bins, two compare-and-count steps, one extra register, and an index, mathematically identical to \texttt{std::upper\_bound} (Section~\ref{v2:method:vectorization}).

% Para 3 -- per-node algorithm selection: precedent and the sharpened novelty claim.
To our knowledge, no tree learner switches between exact and histogram splitting per node, which is the choice our dynamic histograms make.

% Para 5 -- Highway. Library facts only; the 2-3x number lives in the appendix table.
Highway~\cite{wassenberg_vectorized_2022} is a C++ library of portable SIMD operations with runtime dispatch across x86, ARM and RISC-V targets. Its vectorized quicksort, VQSort, is the sort behind YDF's exact splits and the baseline for every speedup in this paper.

% Para 6 -- inference-side work, one sentence.
Inference performance optimization of RF and GBT includes: trees being compiled into C++~\cite{cho2018treelite,rapids} to improve memory layout and reduce branching and into tensor operations~\cite{nakandala2020tensor}. Cache-efficient memory layouts have been applied to nodes in multiple trees to reduce random I/O~\cite{browne2022forestpacking,madhyastha2021blockset}.

\section{Method}
\label{v2:method}

We describe the per-node computation of sparse oblique splitting and its cost, then the two optimizations: a vectorized bin lookup (Section~\ref{v2:method:vectorization}) and a per-node choice between histogram and exact splitting (Section~\ref{v2:method:dynamic}). Throughout, $n$ is the number of active samples at a node, $d$ the number of features, $P$ the projections per node, $k$ the histogram bins and $w$ the lanes of a vector register.

\subsection{Sparse Oblique Workflow and Its Cost}
\label{v2:method:workflow}

\begin{figure*}[thbp]
    \centering
    \includegraphics[width=1\linewidth]{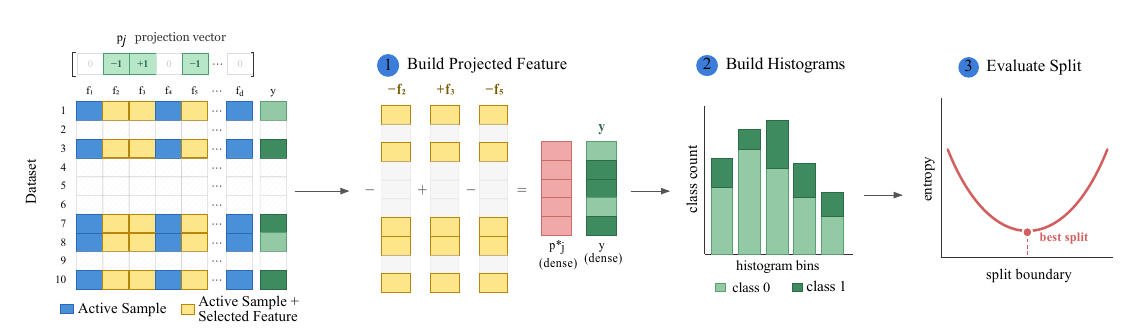}

    \caption{Workflow at each tree node. Histogram splitting of a random linear combination of features requires sparse access in both rows and columns, computing a vector sum, building histograms and evaluating split boundaries.}
    \Description{Computation performed at each tree node includes: Sampling a projection, applying that projection, which involves sparse access across both rows and columns, building histograms and evaluating splits.}
    \label{fig:computenode}
\end{figure*}

YDF stores the data column by column and never copies it; a node is a sorted list of row indices, and each split partitions the parent's list in place. For RF the root holds a bootstrap sample of $n$ rows drawn with replacement, about 63\% of them distinct. At each node SPO draws $P=\lceil\sqrt{d}\,\rceil$ projections, capped at 1000~\cite{tomita2020sparse}, each a sparse vector with Binomial$(d,\,1.5/d)$ non-zeros, i.e.\ 1.5 on average, features chosen without replacement and weights in $\{-1,1\}$. Every projection passes through four stages (Figure~\ref{fig:computenode}):
\begin{enumerate}
\item \emph{Sample Projection}: draw the non-zero features and weights.
\item \emph{Build Projected Feature}: gather the selected columns at the active rows and accumulate the weighted sum into $n$ values. Sparse in both dimensions and increasingly scattered with depth.
\item \emph{Build Histograms}, or sort: bin each value and count its class, or sort the values for exact splitting.
\item \emph{Evaluate Split}: score every candidate threshold by information gain. Keep the best over all $P$ projections.
\end{enumerate}

\emph{Exact} splitting packs each (value, label) pair into a 64-bit key, sorts the $n$ keys with Highway's VQSort~\cite{wassenberg_vectorized_2022} and scans them once, considering the midpoint between every pair of adjacent distinct values: $O(n\log n)$ per projection. \emph{Random Histogram} splitting draws $k$ thresholds uniformly between the node's minimum and maximum value, sorts them, bins each value by binary search over the thresholds, and scans the $k$ candidates: $O(n\log k)$ for the lookups plus an $O(k)$ term independent of $n$ for drawing and sorting the thresholds, initializing the counters and scanning the bins. Because the thresholds are random rather than equally spaced, the bin index has no closed form, unlike learners that quantize each feature once before training~\cite{ke2017lightgbm,prokhorenkova2018catboost}. The search runs once per value, per projection, per node.

\begin{figure}[t]
    \centering
    \includegraphics[width=\linewidth]{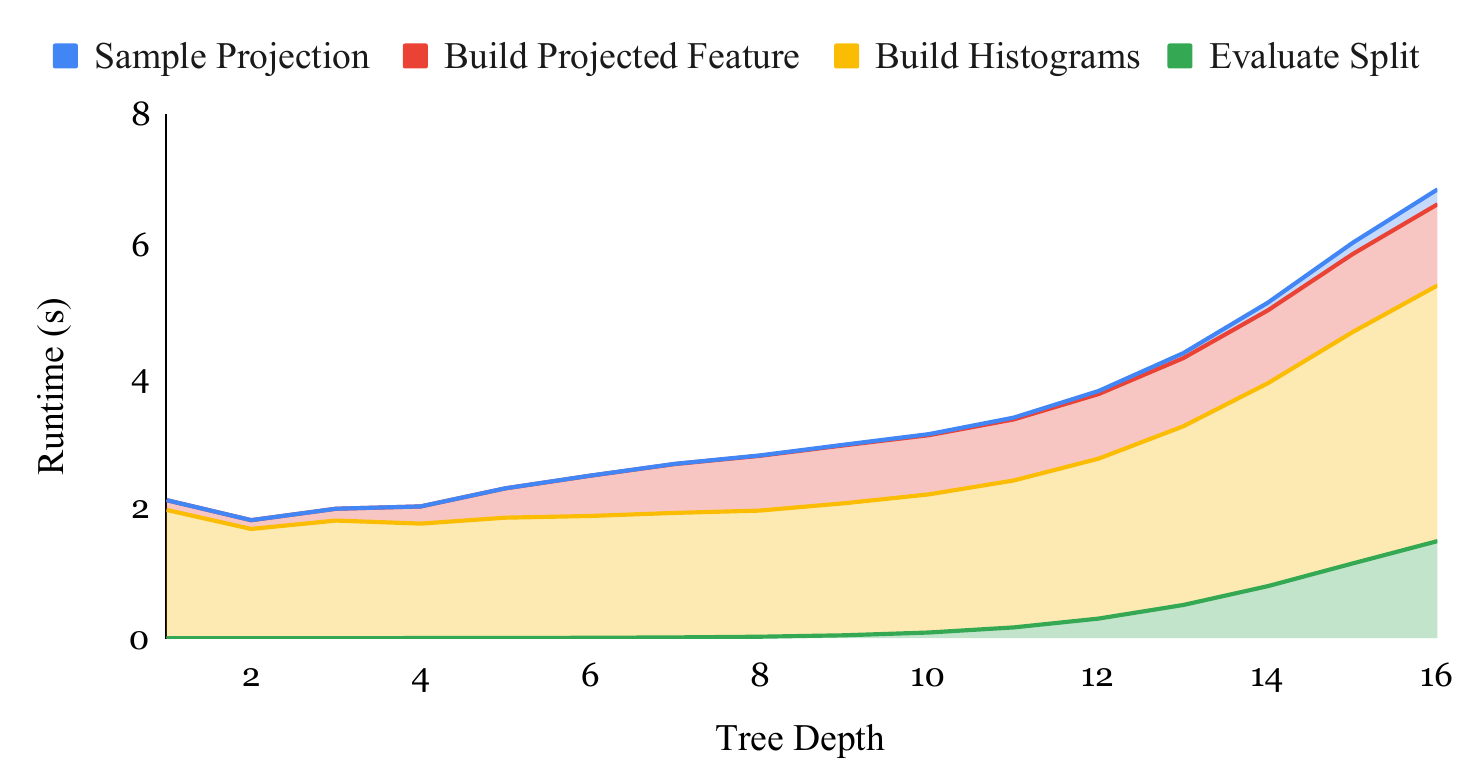}
    \caption{Comparative runtime of the different components of computation for histogram splitting. The dataset has 1M samples and 4096 features.}
    \Description{Share of histogram runtime (in dynamic mode) attributable to each of 4 subfunctions. Plot reveals Build Histograms requires the majority of the time, necessitating Vectorization. }
    \label{fig:decompose}
\end{figure}

Figure~\ref{fig:decompose}, from our per-node timers, breaks Random Histogram training time down by stage and depth on 1M rows and 4096 features. % this is for scalar, but vectorized looks identical, except Build Histogram section is bigger
Building histograms dominates at every depth. Its cost per depth is flat while nodes are large, since each depth bins all $n$ samples once per projection, and grows from depth 10 on, where the number of nodes doubles per level and the $O(k)$ work per projection grows with it. Evaluate Split, also $O(k)$ per projection, appears at the same depth. The two optimizations follow: in shallow trees the per-value lookup dominates, in deep trees the per-projection fixed cost does.

\subsection{Vectorization of Histogram Filling}
\label{v2:method:vectorization}

\begin{figure*}[!t]
\centering
\begin{subfigure}[b]{0.44\textwidth}
\centering
\includegraphics[width=.8\linewidth]{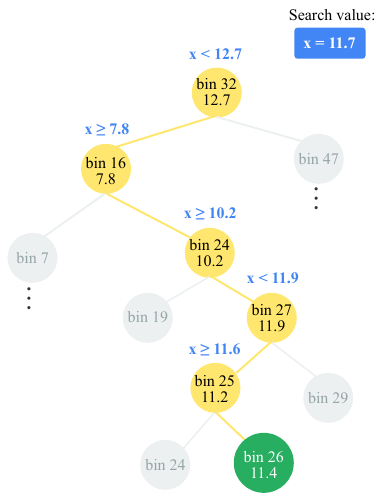}
% \caption{Illustration of Binary Search on 64 histogram bins.}
\end{subfigure}
\begin{subfigure}[b]{0.54\textwidth}
\includegraphics[width=\linewidth]{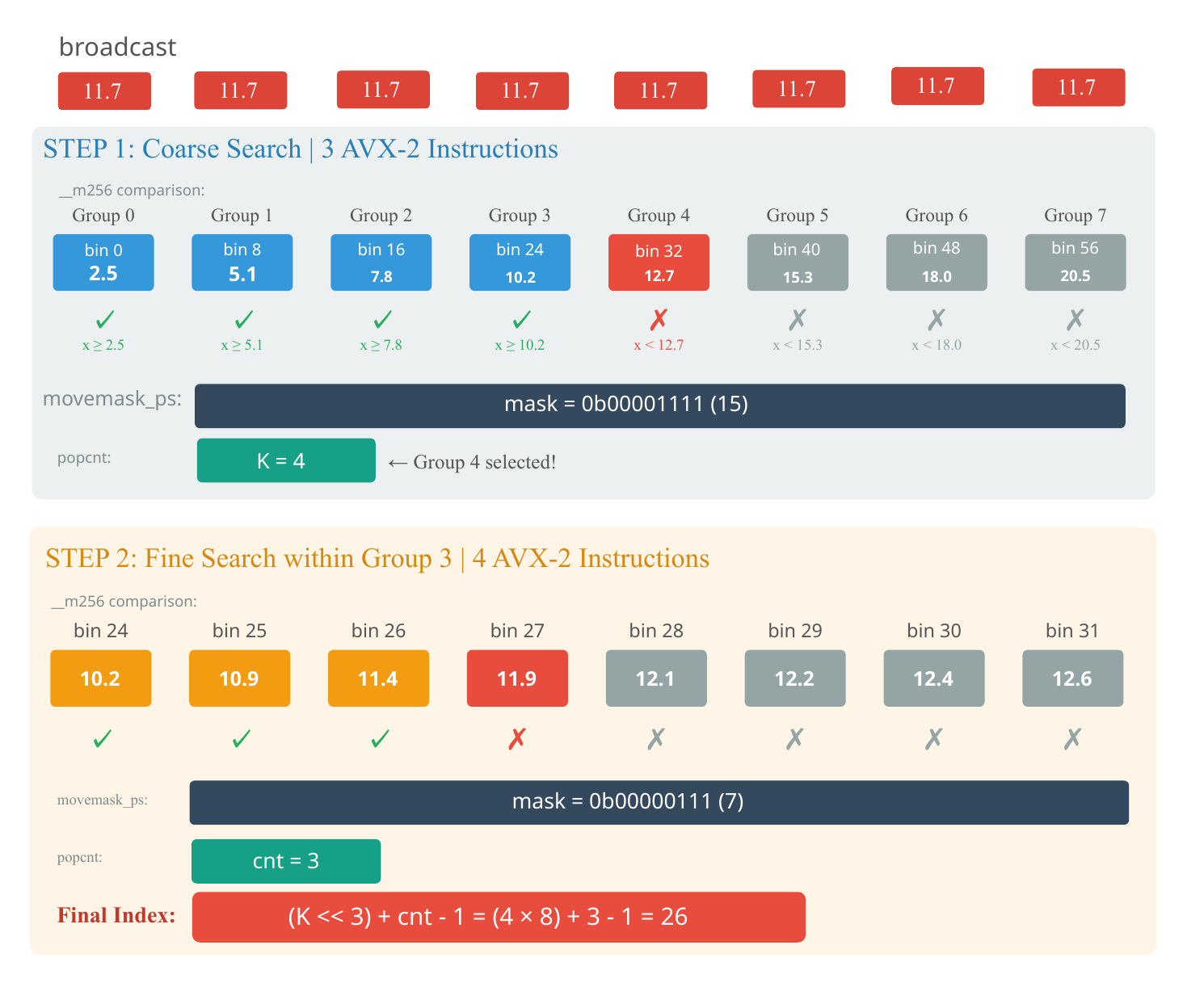}
% \caption{Illustration of conversion of Binary Search into AVX2 SIMD instructions.}
\end{subfigure}
\caption{Mapping sample points to histograms using binary search (left) and our vectorized AVX2 implementation (right).}
\label{fig:vectorization-illustration}
\Description{Illustration of mapping sample points to histograms using binary search (default, left) and our vectorized implementation (right).}
\end{figure*}

YDF bins a value with \texttt{std::upper\_bound}, a binary search over the $k$ sorted thresholds (Figure~\ref{fig:vectorization-illustration}, left): $\log_2 k$ dependent steps, each ending in a data-dependent branch that can cause mispredictions. A linear scan avoids the branches but costs $k$ comparisons per value.

We replace the search by two levels of SIMD comparisons (Figure~\ref{fig:vectorization-illustration} right). With $w$ lanes, $w=8$ for AVX2 and $w=16$ for AVX-512, we use $k=w^2$ bins. Once per histogram we copy the sorted thresholds $T[0,k)$ into an aligned array and build a coarse table $C[0,w)$ of the last threshold of each group of $w$ bins, $C[g]=T[w(g+1)-1]$. Per value $x$, one broadcast, one compare against $C$ and one \texttt{popcnt} give $K$, the number of groups whose thresholds are all at most $x$, i.e.\ the group containing $x$; a second compare against the $w$ thresholds of group $K$ and a second \texttt{popcnt} give the count $j$ of thresholds at most $x$ within it, and the bin is $wK+j-1$, or $-1$ if $x$ lies below every threshold. If $K=w$ the value belongs to the last bin.

% \begin{figure}[t]
% \begin{lstlisting}[basicstyle=\ttfamily\footnotesize, columns=fullflexible, keepspaces=true, morecomment={[l]{//}}, commentstyle=\itshape]
% // Setup, once per histogram (T[0..k) sorted, k = w*w)
% for g in 0..w-1: C[g] = T[w*(g+1) - 1]  // last of group g

% // Lookup per value x: largest threshold <= x, or -1
% vx = broadcast(x)
% K  = popcnt(mask(vx >= C[0..w)))        // group of x
% if K == w: return k - 1                 // above all
% j  = popcnt(mask(vx >= T[w*K .. w*K+w)))  // in group K
% return w*K + j - 1
% \end{lstlisting}
% \caption{Two-level SIMD bin lookup. One vector compare and popcount select the group of $w$ bins, a second pair selects the bin. $w=8$ on AVX2 (64 bins) and $w=16$ on AVX-512 (256 bins).}
% \Description{Pseudocode of the two-level SIMD bin lookup.}
% \label{v2:lst:binsearch}
% \end{figure}

The lookup is seven vector and bit-count instructions on AVX2 and five on AVX-512, whose compares write a mask register directly, with no data-dependent branch. Thresholds are drawn, sorted and used exactly as before, and the lookup returns the same index as \texttt{std::upper\_bound} for every input, which debug builds assert, so the trees are bit-identical to the scalar implementation. The kernel is selected at run time from the CPU and the bin count, with a scalar fallback, and serves the axis-aligned histogram finder unchanged. The two widths are not interchangeable: the lookup costs the same, but the $O(k)$ per-projection work is four times larger with 256 bins, so AVX-512 wins only where $n$ is large relative to $k$ and loses in deep trees (Figure~\ref{fig:AVX-512-slower}).

\subsection{Dynamic Histogramming}
\label{v2:method:dynamic}

\begin{figure}[!ht]
    \centering
    \includegraphics[width=1\linewidth]{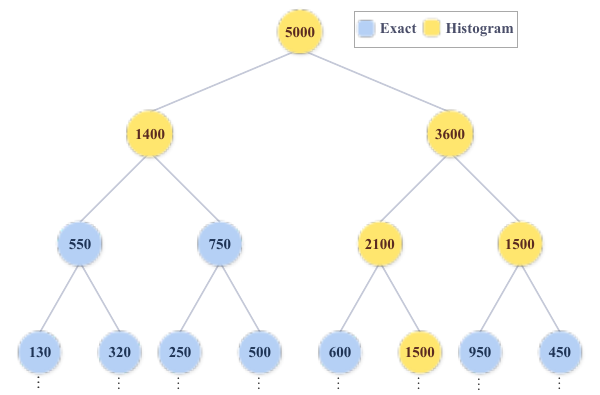}
    \caption{Illustration of the selection of exact splitting and histogram splitting as a function of the number of active samples in a tree node with a breakeven point of 1300.}
    \Description{Illustration of the selection of exact splitting and histogram splitting in an example decision tree.}
    \label{fig:levels}
\end{figure}

Histogram and exact splitting are expensive at opposite ends of the tree (Figure~\ref{fig:dynamic-motivation-nodetrain}). Near the root the $O(n\log n)$ sort dominates and histograms are faster~\cite{ydf,prokhorenkova2018catboost}, and deep in the tree the number of nodes doubles per level while samples per node halve, and the $O(k)$ work per histogram, paid at every node regardless of $n$, dominates.

We therefore choose the finder per node: before the projection loop, a node with $n<\tau$ samples uses exact splitting for all its projections, otherwise vectorized histograms. The choice costs one comparison per node, and nodes at the same depth may have differerent split types (Figure~\ref{fig:levels}). The threshold $\tau$ is a single integer hyperparameter that depends on the kernel and the sort, not on the dataset. We provide a microbenchmark to heuristically set it at the first run on a new machine. For our test, we chose $\tau=250$ for the 64-bin AVX2 kernel and $1000$ for the 256-bin AVX-512 kernel. A full runtime comparison of various thresholds is provided in Figure~\ref{fig:threshold-sensitivity}. Near the top of the tree, most nodes clear the threshold and dynamic splitting chooses as vectorized histograms, whereas below it a growing share of nodes is exact, avoiding the histogram's fixed cost.

Dynamic forests are not node-for-node identical to histogram forests: exact splitting considers every midpoint instead of $k$ random thresholds, and since the histogram finder consumes random numbers for its thresholds and the exact finder does not, projections drawn after the first switch differ as well. The split rule is unchanged at every node with $n\ge\tau$, and we find no detectable effect on accuracy (Table~\ref{tab:accuracy-rf}).

\section{Results}
\label{sec:results}

% v2 (2026-09-26): Results rewritten to follow the abstract/intro claims. New text sits
% above the previous text of each subsection, kept verbatim below a 'previous text' marker.

We report each contribution in the setting described in the introduction. Section~\ref{app:parameters} sets the protocol and the baseline. Section~\ref{sec:res-depth6} measures vectorized histogram filling on depth-6 trees, for SPO-GBT and SPO-RF. Section~\ref{sec:res-depth16} follows SPO-RF from depth 16 to 24, where the dynamic switch takes over from vectorization, and Section~\ref{sec:res-purity} completes the depth range at purity and compares against YDF's axis-aligned RF. Section~\ref{sec:res-scalability} maps the speedup over rows and features, Section~\ref{sec:dyn} separates the two optimizations and the two bin counts, Section~\ref{app:threshold-sensitivity} varies the switch threshold, and Section~\ref{sec:res-accuracy} shows that this does not change accuracy. Table~\ref{tab:train-time-by-depth} (SPO-RF, six depths) and Table~\ref{tab:train-time-gbt-depth6} (SPO-GBT, depth 6) in the appendix hold the full timing results.

\subsection{Experimental Setup}
\label{app:parameters}

\begin{table}[t]
    \centering
    \begin{tabular}{c|c|c|c}
Dataset & Task & Rows & Feat. \\
\midrule
HIGGS~\cite{higgs_280} & bin & 11,000,000 & 28 \\
SUSY~\cite{susy_279} & bin & 5,000,000 & 18 \\
Epsilon~\cite{chang2011libsvm} & bin & 400,000 & 2,000 \\
YouTube-8M~\cite{abu2016youtube}         & mul$^*$   & 3,888,919            & 1152 \\
Shifts Weather~\cite{malinin2021shifts}   & reg$^*$   & 3,129,592            & 127 \\
ClimSim~\cite{yu2023climsim}              & reg$^*$   & 10,091,520           & 124 \\
Jane Street~\cite{jane-street-real-time-market-data-forecasting}       & reg$^*$   & 10,213,190$^\dagger$ & 82
\end{tabular}
\caption{Natural Datasets used in our experiments. * - treated as \texttt{bin}. $\dagger$ - subset used. Full datasets and their description in Table~\ref{tab:all-datasets}.}
\label{tab:natural-datasets}
\end{table}

\paragraph{Hardware and software} All timings come from one AWS m7i.metal-24xl instance (Intel Xeon Platinum 8488C, 48 physical cores, 384\,GiB, Ubuntu 24.04) running YDF compiled with Intel ICX 2025.2.1 and AVX-512 enabled, with 48 threads. Random forests train one tree per thread; gradient-boosted trees parallelize the split search within each tree. Times cover training only: dataset loading and model serialization are excluded.

\paragraph{Learners} SPO-RF: 240 trees, bootstrap sampling with replacement and a bag size equal to the number of rows, \code{min\_examples}=1, maximum depth 6, 10, 16, 20, 24 or unlimited (purity). SPO-GBT: 300 trees of depth 6, shrinkage 0.1, \code{min\_examples}=5. In GBT, every node scans all $F$ features (axis-aligned methods) or samples $F$ projections (sparse oblique methods). In RF, each node samples $\lceil\sqrt{F}\rceil$ features (axis-aligned) or sparse projections (SPO), with at most 1000. All SPO projections in turn sample 1.5 features each. All other parameters are YDF defaults.

\paragraph{Split finders and baseline} Exact splits sort the projected values with Highway VQSort and are the baseline for every speedup below: speedup $= T_{\mathrm{Exact}}/T_{\mathrm{method}}$, so values above 1 are faster. Table~\ref{tab:stdsort-vs-highway} in the appendix relates this baseline to the \texttt{std::sort} exact finder of the released YDF, which it improves by 1.8--3.5$\times$. Random histograms draw their bin boundaries per projection and use 64 bins with the scalar (\texttt{std::upper\_bound}) and AVX2 binners and 256 bins with the scalar and AVX-512 binners. The dynamic variants bin nodes with at least $\tau$ rows and sort the rest, with $\tau=250$ for AVX2/64 bins and $\tau=1000$ for AVX-512/256 bins (Section~\ref{app:threshold-sensitivity}). The scalar 64-bin binner has a breakeven threshold $\tau>50000$, and therefore appears only as an non-Dynamic baseline.

\paragraph{Datasets and protocol} Sevel natural datasets (Table~\ref{tab:natural-datasets}) and 13 Trunk shapes spanning 15k to 4.5M rows and 32 to 1.6M features (Table~\ref{tab:train-time-by-depth}). Because regression is slightly slower than classification, regression targets are thresholded at the training-set median and the multi-label YouTube-8M target is its most frequent class against the rest. 
% Tables~\ref{tab:train-time-by-depth} and \ref{tab:train-time-gbt-depth6} and Figures~\ref{fig:overall} and \ref{fig:speedup-heatmap} are single runs per cell with one seed.
Accuracy protocols are stated in Section~\ref{sec:res-accuracy}.

% ---- previous text below (kept for comparison. v2 rewrite above, 2026-09-26) ----

% \textit{\textbf{Big embeddings are an important use case at Google. Google can't give access, but we make a proxy. }} The use of tabular-embedding multimodal is picking up pace. \textit{Text + LLM embeddings is a strong GBT use case. Tabular structure gives benefit \cite{jiang2026multimodal}}. Good but small: MulTABench \cite{arazi_multabench_2026}: Multimodal Tabular Benchmark - great but small: max 114,000 rows x 245 features, AutoML Bag of Tricks~\cite{tang_bag_2024} (20k rows), AutoGluon–TimeSeries~\cite{shchur_autogluon-timeseries_2023} (100k), Multimodal Tabular Reasoning~\cite{jiang2026multimodal}, however, these are too small for a performance study. As such, we use datasets described in~\ref{tab:natural-datasets}.

\subsection{Depth 6 - GBTs}
\label{sec:res-depth6}

% fig:overall lives in the generated file (make_e2e_grid_figure.py). caption/label are set there.
% Generated by benchmarks/src/utils/make_e2e_grid_figure.py -- do not hand-edit.
% Needs \usepackage{pgfplots}\pgfplotsset{compat=1.18}\usepgfplotslibrary{groupplots}.
\definecolor{e2ec0}{HTML}{4285F4}
\definecolor{e2ec1}{HTML}{34A853}
\definecolor{e2ec2}{HTML}{0B6B3A}
\definecolor{e2ec3}{HTML}{B57BFF}
\definecolor{e2ec4}{HTML}{6A1B9A}
\definecolor{e2ec5}{HTML}{FBBC04}
\definecolor{e2ec6}{HTML}{EA4335}
\pgfplotsset{bar0/.style={ybar, fill=e2ec0, draw=e2ec0!70!black, line width=0.2pt}}
\pgfplotsset{bar1/.style={ybar, fill=e2ec1, draw=e2ec1!70!black, line width=0.2pt}}
\pgfplotsset{bar2/.style={ybar, fill=e2ec2, draw=e2ec2!70!black, line width=0.2pt}}
\pgfplotsset{bar3/.style={ybar, fill=e2ec3, draw=e2ec3!70!black, line width=0.2pt}}
\pgfplotsset{bar4/.style={ybar, fill=e2ec4, draw=e2ec4!70!black, line width=0.2pt}}
\pgfplotsset{bar5/.style={ybar, fill=e2ec5, draw=e2ec5!70!black, line width=0.2pt}}
\pgfplotsset{bar6/.style={ybar, fill=e2ec6, draw=e2ec6!70!black, line width=0.2pt}}
\begin{figure*}[!t]
    \centering
    \begin{tikzpicture}
    \begin{groupplot}[
        group style={group size=2 by 2, horizontal sep=0.75cm, vertical sep=1.05cm,
            xticklabels at=edge bottom},
        width=0.53\textwidth, height=4.5cm,
        /pgf/bar width=0.1, ybar, ybar legend,
        xmin=0.45, xmax=7.55, ymin=0, ymax=1.5, ytick={0,0.5,1,1.5},
        xtick={1,2,3,4,5,6,7},
        xticklabels={Epsilon, Jane Street, SUSY, HIGGS, YouTube-8M, Trunk 150k$\times$4096, Trunk 15k$\times$400k},
        xticklabel style={rotate=28, anchor=north east, font=\scriptsize, yshift=-1pt},
        yticklabel style={font=\scriptsize}, title style={font=\small, yshift=-4pt},
        tick align=outside, tick pos=left, axis line style={black!50},
        legend columns=3, legend cell align=left,
        legend style={font=\scriptsize, draw=gray!40, /tikz/every even column/.append style={column sep=5pt}},
        legend image code/.code={\draw[#1] (0cm,-0.08cm) rectangle (0.25cm,0.14cm);},
    ]
    \nextgroupplot[title={SPO-GBT, depth 6}]
    \addplot[bar0, /pgf/bar shift=-0.3500] coordinates {(1,1.0000) (2,1.0000) (3,1.0000) (4,1.0000) (5,1.0000) (6,1.0000) (7,1.0000)};
    \addplot[bar1, /pgf/bar shift=-0.2500] coordinates {(1,1.5563) (2,1.0293) (3,1.4808) (4,1.5305) (5,0.9780) (6,1.2973) (7,1.7768)};
    \addplot[bar2, /pgf/bar shift=-0.1500] coordinates {(1,2.1522) (2,1.3991) (3,2.0342) (4,2.0756) (5,1.3292) (6,1.8327) (7,2.6210)};
    \addplot[bar3, /pgf/bar shift=0.0000] coordinates {(1,0.7031) (2,0.5444) (3,0.6787) (4,0.6907) (5,0.5574) (6,0.5327) (7,0.7466)};
    \addplot[bar4, /pgf/bar shift=0.1000] coordinates {(1,0.7808) (2,0.5531) (3,0.6502) (4,0.6743) (5,0.5867) (6,0.5790) (7,0.8466)};
    \draw[gray!70, dashed, line width=0.5pt] (axis cs:0.45,1) -- (axis cs:7.55,1);
    \draw[white, line width=1.1pt] (axis cs:0.6880,1.380) -- (axis cs:0.8120,1.430);
    \draw[white, line width=1.1pt] (axis cs:0.6880,1.440) -- (axis cs:0.8120,1.490);
    \draw[white, line width=1.1pt] (axis cs:3.6880,1.380) -- (axis cs:3.8120,1.430);
    \draw[white, line width=1.1pt] (axis cs:3.6880,1.440) -- (axis cs:3.8120,1.490);
    \draw[white, line width=1.1pt] (axis cs:6.6880,1.380) -- (axis cs:6.8120,1.430);
    \draw[white, line width=1.1pt] (axis cs:6.6880,1.440) -- (axis cs:6.8120,1.490);
    \draw[white, line width=1.1pt] (axis cs:0.7880,1.380) -- (axis cs:0.9120,1.430);
    \draw[white, line width=1.1pt] (axis cs:0.7880,1.440) -- (axis cs:0.9120,1.490);
    \draw[white, line width=1.1pt] (axis cs:2.7880,1.380) -- (axis cs:2.9120,1.430);
    \draw[white, line width=1.1pt] (axis cs:2.7880,1.440) -- (axis cs:2.9120,1.490);
    \draw[white, line width=1.1pt] (axis cs:3.7880,1.380) -- (axis cs:3.9120,1.430);
    \draw[white, line width=1.1pt] (axis cs:3.7880,1.440) -- (axis cs:3.9120,1.490);
    \draw[white, line width=1.1pt] (axis cs:5.7880,1.380) -- (axis cs:5.9120,1.430);
    \draw[white, line width=1.1pt] (axis cs:5.7880,1.440) -- (axis cs:5.9120,1.490);
    \draw[white, line width=1.1pt] (axis cs:6.7880,1.380) -- (axis cs:6.9120,1.430);
    \draw[white, line width=1.1pt] (axis cs:6.7880,1.440) -- (axis cs:6.9120,1.490);
    \nextgroupplot[title={SPO-RF, depth 10}]
    \addplot[bar0, /pgf/bar shift=-0.3500] coordinates {(1,1.0000) (2,1.0000) (3,1.0000) (4,1.0000) (5,1.0000) (6,1.0000) (7,1.0000)};
    \addplot[bar1, /pgf/bar shift=-0.2500] coordinates {(1,1.1974) (2,0.7162) (3,0.8726) (4,0.8126) (5,0.8985) (6,1.3052) (7,1.5191)};
    \addplot[bar2, /pgf/bar shift=-0.1500] coordinates {(1,1.5118) (2,0.8300) (3,1.0348) (4,0.9466) (5,1.0719) (6,1.6716) (7,2.0709)};
    \addplot[bar3, /pgf/bar shift=0.0000] coordinates {(1,0.7087) (2,0.5889) (3,0.6161) (4,0.6315) (5,0.6454) (6,0.7319) (7,0.8538)};
    \addplot[bar4, /pgf/bar shift=0.1000] coordinates {(1,0.7287) (2,0.5952) (3,0.6135) (4,0.6423) (5,0.6508) (6,0.7791) (7,1.0362)};
    \draw[gray!70, dashed, line width=0.5pt] (axis cs:0.45,1) -- (axis cs:7.55,1);
    \draw[white, line width=1.1pt] (axis cs:6.6880,1.380) -- (axis cs:6.8120,1.430);
    \draw[white, line width=1.1pt] (axis cs:6.6880,1.440) -- (axis cs:6.8120,1.490);
    \draw[white, line width=1.1pt] (axis cs:0.7880,1.380) -- (axis cs:0.9120,1.430);
    \draw[white, line width=1.1pt] (axis cs:0.7880,1.440) -- (axis cs:0.9120,1.490);
    \draw[white, line width=1.1pt] (axis cs:5.7880,1.380) -- (axis cs:5.9120,1.430);
    \draw[white, line width=1.1pt] (axis cs:5.7880,1.440) -- (axis cs:5.9120,1.490);
    \draw[white, line width=1.1pt] (axis cs:6.7880,1.380) -- (axis cs:6.9120,1.430);
    \draw[white, line width=1.1pt] (axis cs:6.7880,1.440) -- (axis cs:6.9120,1.490);
    \nextgroupplot[title={SPO-RF, depth 20}]
    \addplot[bar0, /pgf/bar shift=-0.3500] coordinates {(1,1.0000) (2,1.0000) (3,1.0000) (4,1.0000) (5,1.0000) (6,1.0000) (7,1.0000)};
    \addplot[bar1, /pgf/bar shift=-0.2500] coordinates {(1,1.4956) (2,0.9027) (3,1.0708) (4,1.0226) (5,1.0937) (6,1.6122) (7,1.5965)};
    \addplot[bar2, /pgf/bar shift=-0.1500] coordinates {(1,2.2918) (2,1.1026) (3,1.3608) (4,1.2535) (5,1.4051) (6,2.5911) (7,2.4816)};
    \addplot[bar3, /pgf/bar shift=0.0000] coordinates {(1,0.9799) (2,0.7138) (3,0.7102) (4,0.7413) (5,0.7445) (6,1.0528) (7,1.0762)};
    \addplot[bar4, /pgf/bar shift=0.1000] coordinates {(1,1.4981) (2,0.7422) (3,0.7666) (4,0.7809) (5,0.8331) (6,1.7200) (7,1.6647)};
    \addplot[bar5, /pgf/bar shift=0.2500] coordinates {(1,0.8266) (2,0.7047) (3,0.6938) (4,0.7338) (5,0.7324) (6,0.8489) (7,0.8732)};
    \addplot[bar6, /pgf/bar shift=0.3500] coordinates {(1,0.8608) (2,0.7139) (3,0.7028) (4,0.7425) (5,0.7399) (6,0.8892) (7,0.9006)};
    \draw[gray!70, dashed, line width=0.5pt] (axis cs:0.45,1) -- (axis cs:7.55,1);
    \draw[white, line width=1.1pt] (axis cs:5.6880,1.380) -- (axis cs:5.8120,1.430);
    \draw[white, line width=1.1pt] (axis cs:5.6880,1.440) -- (axis cs:5.8120,1.490);
    \draw[white, line width=1.1pt] (axis cs:6.6880,1.380) -- (axis cs:6.8120,1.430);
    \draw[white, line width=1.1pt] (axis cs:6.6880,1.440) -- (axis cs:6.8120,1.490);
    \draw[white, line width=1.1pt] (axis cs:0.7880,1.380) -- (axis cs:0.9120,1.430);
    \draw[white, line width=1.1pt] (axis cs:0.7880,1.440) -- (axis cs:0.9120,1.490);
    \draw[white, line width=1.1pt] (axis cs:5.7880,1.380) -- (axis cs:5.9120,1.430);
    \draw[white, line width=1.1pt] (axis cs:5.7880,1.440) -- (axis cs:5.9120,1.490);
    \draw[white, line width=1.1pt] (axis cs:6.7880,1.380) -- (axis cs:6.9120,1.430);
    \draw[white, line width=1.1pt] (axis cs:6.7880,1.440) -- (axis cs:6.9120,1.490);
    \draw[white, line width=1.1pt] (axis cs:6.0380,1.380) -- (axis cs:6.1620,1.430);
    \draw[white, line width=1.1pt] (axis cs:6.0380,1.440) -- (axis cs:6.1620,1.490);
    \draw[white, line width=1.1pt] (axis cs:7.0380,1.380) -- (axis cs:7.1620,1.430);
    \draw[white, line width=1.1pt] (axis cs:7.0380,1.440) -- (axis cs:7.1620,1.490);
    \nextgroupplot[title={SPO-RF, full depth (purity)}, legend to name=e2elegendgrid]
    \addplot[bar0, /pgf/bar shift=-0.3500] coordinates {(1,1.0000) (2,1.0000) (3,1.0000) (4,1.0000) (5,1.0000) (6,1.0000) (7,1.0000)};
    \addlegendentry{Exact}
    \addplot[bar1, /pgf/bar shift=-0.2500] coordinates {(1,1.6318) (2,1.3358) (3,1.4146) (4,1.4201) (5,1.2174) (6,1.6625) (7,1.6188)};
    \addlegendentry{Random Hist. (64 bins)}
    \addplot[bar2, /pgf/bar shift=-0.1500] coordinates {(1,2.6361) (2,2.1024) (3,2.2016) (4,2.1851) (5,1.6924) (6,2.6886) (7,2.6007)};
    \addlegendentry{Random Hist. (256 bins)}
    \addplot[bar3, /pgf/bar shift=0.0000] coordinates {(1,1.1218) (2,1.0194) (3,0.9771) (4,1.0297) (5,0.8402) (6,1.1082) (7,1.0978)};
    \addlegendentry{AVX2 R. Hist. (ours)}
    \addplot[bar4, /pgf/bar shift=0.1000] coordinates {(1,1.8576) (2,1.5744) (3,1.4964) (4,1.5624) (5,1.0898) (6,1.8375) (7,1.7469)};
    \addlegendentry{AVX-512 R. Hist. (ours)}
    \addlegendimage{empty legend}
    \addlegendentry{}
    \addplot[bar5, /pgf/bar shift=0.2500] coordinates {(1,0.8753) (2,0.8087) (3,0.7856) (4,0.8464) (5,0.7526) (6,0.8506) (7,0.8638)};
    \addlegendentry{Dyn. AVX2 R. Hist. (ours)}
    \addplot[bar6, /pgf/bar shift=0.3500] coordinates {(1,0.8739) (2,0.8098) (3,0.7851) (4,0.8280) (5,0.7644) (6,0.8895) (7,0.8977)};
    \addlegendentry{Dyn. AVX-512 R. Hist. (ours)}
    \draw[gray!70, dashed, line width=0.5pt] (axis cs:0.45,1) -- (axis cs:7.55,1);
    \draw[white, line width=1.1pt] (axis cs:0.6880,1.380) -- (axis cs:0.8120,1.430);
    \draw[white, line width=1.1pt] (axis cs:0.6880,1.440) -- (axis cs:0.8120,1.490);
    \draw[white, line width=1.1pt] (axis cs:5.6880,1.380) -- (axis cs:5.8120,1.430);
    \draw[white, line width=1.1pt] (axis cs:5.6880,1.440) -- (axis cs:5.8120,1.490);
    \draw[white, line width=1.1pt] (axis cs:6.6880,1.380) -- (axis cs:6.8120,1.430);
    \draw[white, line width=1.1pt] (axis cs:6.6880,1.440) -- (axis cs:6.8120,1.490);
    \draw[white, line width=1.1pt] (axis cs:0.7880,1.380) -- (axis cs:0.9120,1.430);
    \draw[white, line width=1.1pt] (axis cs:0.7880,1.440) -- (axis cs:0.9120,1.490);
    \draw[white, line width=1.1pt] (axis cs:1.7880,1.380) -- (axis cs:1.9120,1.430);
    \draw[white, line width=1.1pt] (axis cs:1.7880,1.440) -- (axis cs:1.9120,1.490);
    \draw[white, line width=1.1pt] (axis cs:2.7880,1.380) -- (axis cs:2.9120,1.430);
    \draw[white, line width=1.1pt] (axis cs:2.7880,1.440) -- (axis cs:2.9120,1.490);
    \draw[white, line width=1.1pt] (axis cs:3.7880,1.380) -- (axis cs:3.9120,1.430);
    \draw[white, line width=1.1pt] (axis cs:3.7880,1.440) -- (axis cs:3.9120,1.490);
    \draw[white, line width=1.1pt] (axis cs:4.7880,1.380) -- (axis cs:4.9120,1.430);
    \draw[white, line width=1.1pt] (axis cs:4.7880,1.440) -- (axis cs:4.9120,1.490);
    \draw[white, line width=1.1pt] (axis cs:5.7880,1.380) -- (axis cs:5.9120,1.430);
    \draw[white, line width=1.1pt] (axis cs:5.7880,1.440) -- (axis cs:5.9120,1.490);
    \draw[white, line width=1.1pt] (axis cs:6.7880,1.380) -- (axis cs:6.9120,1.430);
    \draw[white, line width=1.1pt] (axis cs:6.7880,1.440) -- (axis cs:6.9120,1.490);
    \draw[white, line width=1.1pt] (axis cs:1.0380,1.380) -- (axis cs:1.1620,1.430);
    \draw[white, line width=1.1pt] (axis cs:1.0380,1.440) -- (axis cs:1.1620,1.490);
    \draw[white, line width=1.1pt] (axis cs:2.0380,1.380) -- (axis cs:2.1620,1.430);
    \draw[white, line width=1.1pt] (axis cs:2.0380,1.440) -- (axis cs:2.1620,1.490);
    \draw[white, line width=1.1pt] (axis cs:4.0380,1.380) -- (axis cs:4.1620,1.430);
    \draw[white, line width=1.1pt] (axis cs:4.0380,1.440) -- (axis cs:4.1620,1.490);
    \draw[white, line width=1.1pt] (axis cs:6.0380,1.380) -- (axis cs:6.1620,1.430);
    \draw[white, line width=1.1pt] (axis cs:6.0380,1.440) -- (axis cs:6.1620,1.490);
    \draw[white, line width=1.1pt] (axis cs:7.0380,1.380) -- (axis cs:7.1620,1.430);
    \draw[white, line width=1.1pt] (axis cs:7.0380,1.440) -- (axis cs:7.1620,1.490);
    \end{groupplot}
    \path (group c1r1.outer west) -- (group c1r2.outer west)
        node[midway, rotate=90, anchor=south, font=\small] {Runtime Normalized to Exact};
    \end{tikzpicture}

    \vspace{2pt}
    \pgfplotslegendfromname{e2elegendgrid}
    \caption{End-to-end training time on select datasets, normalized to the performance of SO-YDF with exact splits. Panels: SPO-GBT at depth 6; SPO-RF at depth 10, depth 20 and full depth (purity). Blocks of each cluster, left to right: Exact and scalar random histograms; our vectorized histograms; our dynamic-vectorized histograms (not run at depths 6 and 10). Bars above 1.5 are cut off (white break mark).}
    \Description{End-to-end evaluation reveals the benefit of our methods - up to 35\% reduction in runtime with only Dynamic, and up to 50\% with Vectorized Dynamic. It also reveals the similarity in performance of RF Exact to SO Exact.}
    \label{fig:overall}
\end{figure*}

% Generated by make_overall_depth_table.py --natural --reps 1 -- regenerate, don't edit.
% Generated by benchmarks/src/spaa27_experiments/make_overall_depth_table.py --natural -- do not hand-edit.
\begin{table*}[t]
    \centering
    \footnotesize
    \setlength{\tabcolsep}{3pt}
    \begin{tabular}{l|rr|rr|rr|rr|rr}
        \hline
        \multirow{2}{*}{\textbf{Dataset}} & \multicolumn{2}{c|}{\textbf{SPO-GBT, depth 6}} & \multicolumn{2}{c|}{\textbf{SPO-RF, depth 6}} & \multicolumn{2}{c|}{\textbf{SPO-RF, depth 10}} & \multicolumn{2}{c|}{\textbf{SPO-RF, depth 20}} & \multicolumn{2}{c}{\textbf{SPO-RF, purity}} \\
         & Exact & Ours & Exact & Ours & Exact & Ours & Exact & Ours & Exact & Ours \\
        \hline
        HIGGS 10.5M$\times$28 & 2622 & 1811 (1.4$\times$) & 79.6 & 44.7 (1.8$\times$) & 141 & 90.1 (1.6$\times$) & 249 & 183 (1.4$\times$) & 364 & 298 (1.2$\times$) \\
        SUSY 4.5M$\times$18 & 1019 & 692 (1.5$\times$) & 27.0 & 14.9 (1.8$\times$) & 46.1 & 28.4 (1.6$\times$) & 80.6 & 55.9 (1.4$\times$) & 118 & 90.9 (1.3$\times$) \\
        Epsilon 400k$\times$2000 & 541 & 380 (1.4$\times$) & 12.0 & 7.4 (1.6$\times$) & 23.7 & 16.8 (1.4$\times$) & 51.0 & 42.1 (1.2$\times$) & 59.9 & 50.8 (1.2$\times$) \\
        YouTube-8M 3.9M$\times$1152 & 4480 & 2497 (1.8$\times$) & 137 & 70.9 (1.9$\times$) & 235 & 151 (1.6$\times$) & 438 & 320 (1.4$\times$) & 504 & 379 (1.3$\times$) \\
        Shifts weather 3.1M$\times$127 & 856 & 552 (1.6$\times$) & 34.3 & 20.0 (1.7$\times$) & 58.1 & 39.4 (1.5$\times$) & 92.2 & 68.9 (1.3$\times$) & 93.9 & 72.0 (1.3$\times$) \\
        ClimSim 10.1M$\times$124 & 3145 & 1818 (1.7$\times$) & 146 & 73.2 (2.0$\times$) & 228 & 128 (1.8$\times$) & 298 & 189 (1.6$\times$) & 308 & 203 (1.5$\times$) \\
        Jane Street 10.2M$\times$82 & 2846 & 1549 (1.8$\times$) & 121 & 60.1 (2.0$\times$) & 202 & 119 (1.7$\times$) & 374 & 264 (1.4$\times$) & 593 & 479 (1.2$\times$) \\
        \hline
    \end{tabular}
    \caption{End-to-end training time (s) on the natural datasets: Exact (HWY) vs.\ ours (Vec. Random Hist.\ AVX2, 64 bins for SPO-GBT; Vec. Dynamic Hist.\ AVX2, 64 bins for SPO-RF). Speedup over Exact in parentheses. All depths and arms in Tables~\ref{tab:train-time-by-depth} and \ref{tab:train-time-gbt-depth6}.}
    \label{tab:train-time-natural}
\end{table*}

Depth-6 trees over millions of rows have no small nodes: on the seven large natural datasets no node falls below the switch threshold, so no dynamic versions were included (Table~\ref{tab:train-time-by-depth}).

\paragraph{SPO-GBT} With the AVX2 64-bin binner, SPO-GBT trains 1.3--1.6$\times$ faster than Exact on the seven large natural datasets (Tables~\ref{tab:train-time-gbt-depth6},~\ref{tab:train-time-natural}). HIGGS and Epsilon 1.3$\times$, SUSY and Shifts weather 1.4$\times$, YouTube-8M 1.5$\times$, ClimSim and Jane Street 1.6$\times$). The AVX-512 256-bin is within 5\% of it on every dataset (Figure~\ref{fig:overall}, SPO-GBT panel). The YDF baseline Random Hist. scalar is slower than Exact at this depth on every dataset (0.6--0.9$\times$ with 64 bins, 0.4--0.6$\times$ with 256 bins). Vectorization of histograms flips this: the vectorized binner is 1.5--2.2$\times$ (64 bins) and 1.7--2.9$\times$ (256 bins) faster than its scalar counterpart on the natural datasets while building the same trees. On the Trunk shapes the gain over Exact is 1.2--1.4$\times$ from 150k rows up and vanishes on the two 15k-row shapes (1.0$\times$), where the total work (proportional to n. rows) is also low.

\paragraph{SPO-RF} The same kernel gives 1.6--2.0$\times$ on the seven large natural datasets at depth 6 (Tables~\ref{tab:train-time-by-depth},,~\ref{tab:train-time-natural}). Epsilon 1.6$\times$, Shifts weather 1.7$\times$, HIGGS and SUSY 1.8$\times$, YouTube-8M 1.9$\times$, ClimSim and Jane Street 2.0$\times$) and 1.4--1.9$\times$ on the Trunk shapes, rising with the number of rows. Scalar histograms are again unreliable against Exact (0.7--1.6$\times$), whereas the vectorized ones beat Exact on every dataset. How SPO-GBT and SPO-RF compare with axis-aligned YDF and with XGBoost, LightGBM and CatBoost, which pre-sort or pre-quantize their fixed feature set, is reported in Table~\ref{tab:train-time-vs-libraries}.

\subsection{Depth 16 - RF, Dynamic}
\label{sec:res-depth16}

Between depth 16 and depth 24 the need for dynamic switching starts to appear (Table~\ref{tab:train-time-by-depth}, Figure~\ref{fig:overall}).

\paragraph{Depth 16} Vectorized dynamic histograms (AVX2, 64 bins) train 1.3--1.6$\times$ faster than Exact on the seven large natural datasets (Epsilon 1.3$\times$, Shifts weather, HIGGS and YouTube-8M 1.4$\times$, SUSY and Jane Street 1.5$\times$, ClimSim 1.6$\times$) and 1.2--1.4$\times$ on the Trunk shapes. The nodes of a 10M-row dataset still hold hundreds of rows at this depth, so on the natural datasets the switch fires rarely and vectorization supplies almost all of the gain: the dynamic and vectorized-only variants differ by at most 8\% (Epsilon). On the Trunk shapes with 15k--150k rows the switch already matters: vectorized-only histograms fall to 0.95--1.2$\times$ and the switch lifts them to 1.2--1.3$\times$.

\begin{figure*}[b]
\centering
\begin{subfigure}[b]{0.49\textwidth}
\centering
\includegraphics[width=\linewidth]{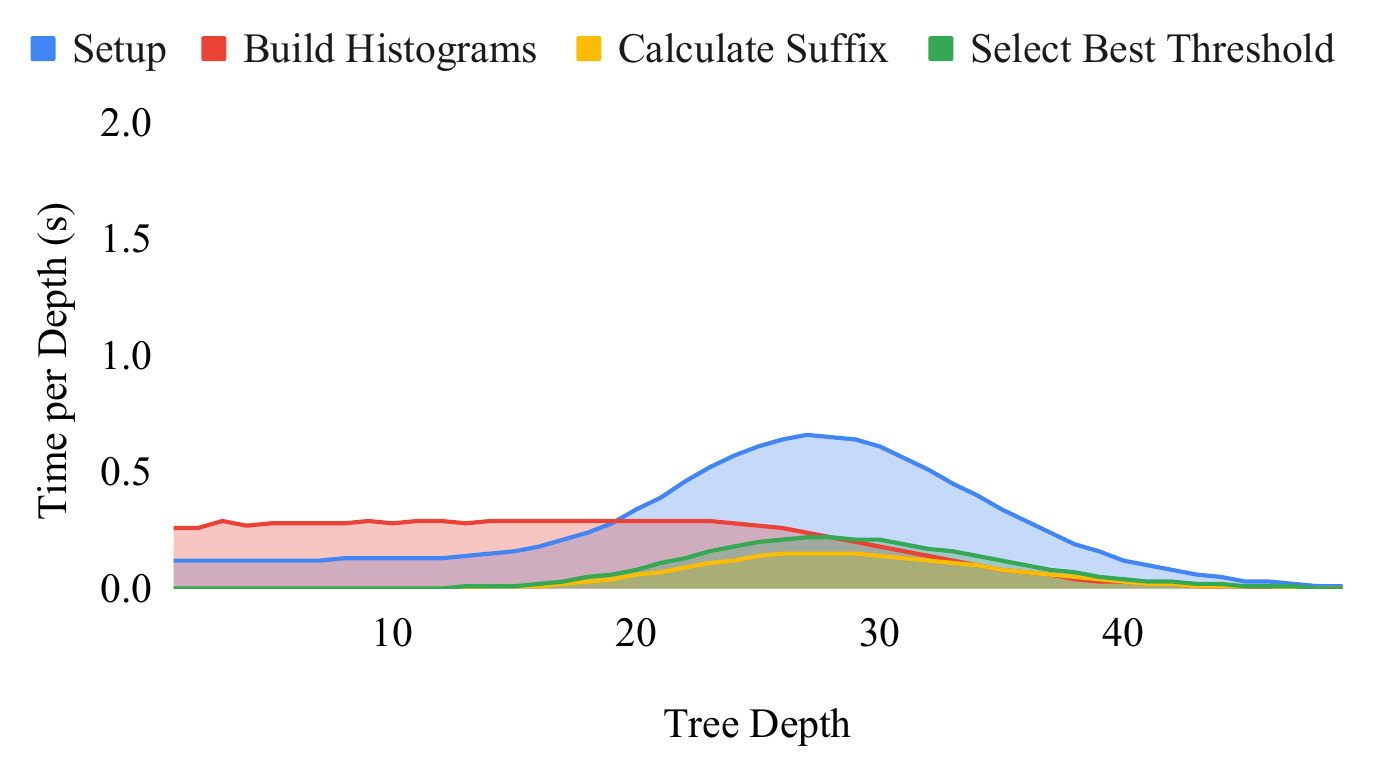}
\caption{AVX2 (64 bins)}
\end{subfigure}
\hfill
\begin{subfigure}[b]{0.49\textwidth}
\centering
\includegraphics[width=\linewidth]{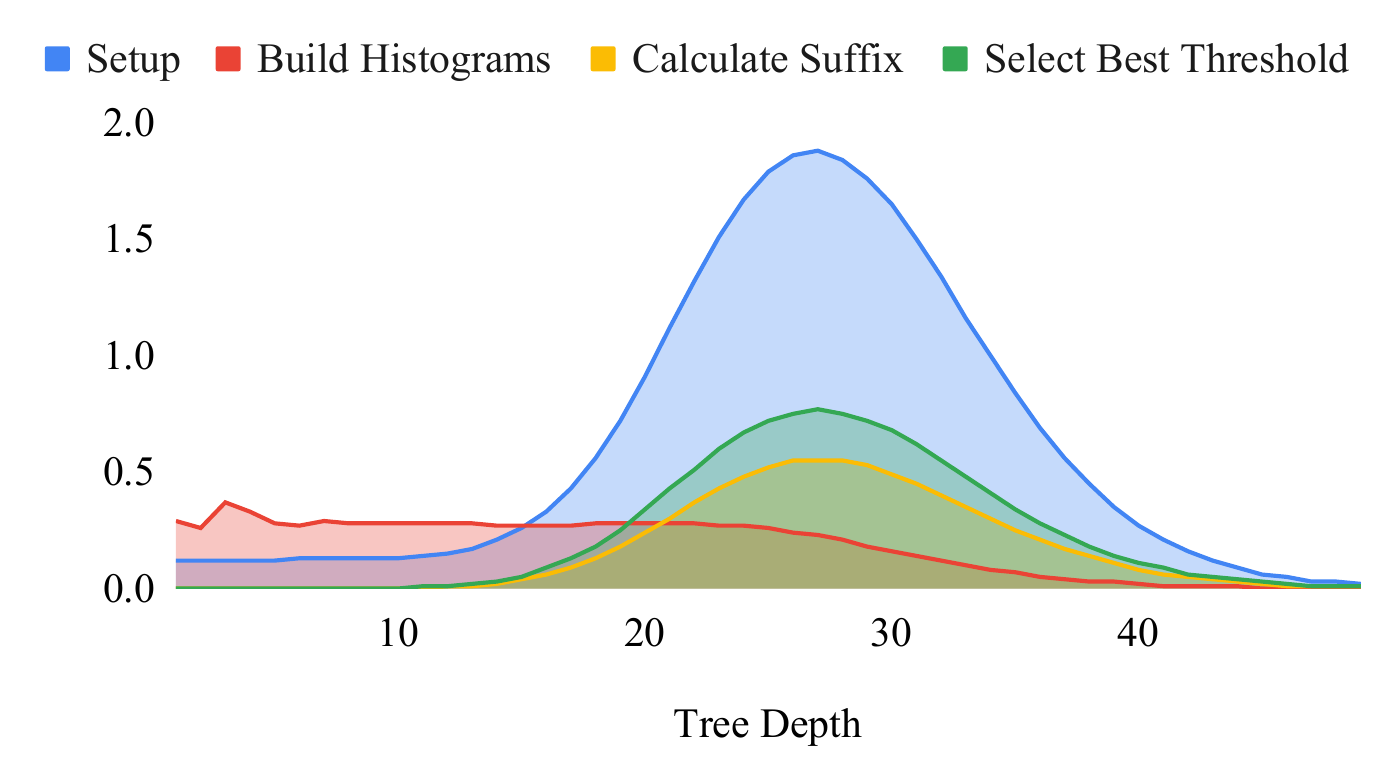}
\caption{AVX-512 (256 bins)}
\end{subfigure}
\caption{Vectorized histogram cost breakdown: AVX2 with 64 bins (left) and AVX-512 with 256 bins (right).}
\label{fig:AVX-512-slower}
\Description{Side-by-side cost breakdown of the vectorized histogram split finder with AVX2 at 64 bins and AVX-512 at 256 bins.}
\end{figure*}

\paragraph{Depths 20 and 24} Each level doubles the number of nodes and halves their size, and the per-projection cost of drawing and scanning the bins (Figure~\ref{fig:AVX-512-slower}) outweighs sorting $<100$ rows. Vectorized-only histograms fall to 0.9--1.5$\times$ Exact on the natural datasets at depth 24 (Epsilon 0.9$\times$, HIGGS 1.2$\times$, ClimSim 1.5$\times$) and to 0.9--1.1$\times$ on the Trunk shapes, with 256 bins they fall to 0.6--1.4$\times$ and 0.5--0.8$\times$. The switch keeps 1.2--1.5$\times$ on the natural datasets at depth 24 (Epsilon 1.2$\times$, HIGGS, Shifts weather and YouTube-8M 1.3$\times$, Jane Street and SUSY 1.4$\times$, ClimSim 1.5$\times$) and 1.1--1.3$\times$ on every Trunk shape, with the two bin counts within 2\% of each other.

\subsection{Training to Purity}
\label{sec:res-purity}

At purity the trend completes (Tables~\ref{tab:train-time-by-depth},~\ref{tab:train-time-natural}), Figure~\ref{fig:overall}, bottom-right). Histogram-only methods now lose to Exact on most datasets: scalar 64-bin histograms run at 0.6--1.0$\times$ Exact on the natural datasets and 0.6--0.75$\times$ on Trunk, vectorized 64-bin at 0.8--1.4$\times$ and 0.9--1.2$\times$ (HIGGS 0.95$\times$, Jane Street 0.98$\times$, Epsilon 0.90$\times$. Only ClimSim, Shifts weather and YouTube-8M stay above 1.1$\times$), and AVX-512 256-bin at 0.5--1.2$\times$. Dynamic switching makes the vectorized histogram a net win on all 21 datasets: 1.2--1.5$\times$ on the seven large natural datasets (Epsilon, HIGGS and Jane Street 1.2$\times$; SUSY, Shifts weather and YouTube-8M 1.3$\times$; ClimSim 1.5$\times$) and 1.1--1.3$\times$ on Trunk. Relative to vectorized-only histograms the switch is worth 1.06--1.31$\times$ on the natural datasets (HIGGS 1.29$\times$, Epsilon 1.31$\times$) and 1.14--1.32$\times$ on Trunk with 64 bins, and 1.3--2.2$\times$ with 256 bins, where it removes the whole bin-count penalty.

\paragraph{Parity with axis-aligned RF} Table~\ref{tab:train-time-vs-libraries} compares full-depth forests on HIGGS, SUSY and Epsilon with YDF's axis-aligned RF, which pre-sorts its fixed features. The Exact SPO-RF baseline is 1.13--1.24$\times$ slower than AA-RF (363 vs.\ 292\,s on HIGGS, 120 vs.\ 104\,s on SUSY, 59 vs.\ 52\,s on Epsilon). Vectorized dynamic histograms close the gap: 296\,s (AVX2) and 288\,s (AVX-512) on HIGGS against 292\,s, and 3--16\% faster than AA-RF on SUSY (92 and 90\,s vs.\ 104\,s) and Epsilon (50.5 and 50.7\,s vs.\ 52.1\,s). Sparse oblique forests thus no longer cost training time over axis-aligned ones in YDF, at the accuracy of Section~\ref{sec:res-accuracy}.

% \paragraph{Cost of the last levels} Growing from depth 24 to purity adds 2--7\% to the dynamic variant's time on Epsilon, YouTube-8M, Shifts weather and ClimSim, but 37--55\% on HIGGS, SUSY and Jane Street, whose trees keep splitting far below depth 24; Exact pays 30--41\% on the same three. The deepest levels are cheap per node, not in aggregate.

% ---- previous text below (kept for comparison; v2 rewrite above, 2026-09-26) ----

% \noindent {\bf Datasets}: For performance experiments, we choose the largest and widest datasets for two-class RF-classification from the UCI Machine Learning Library and from Kaggle (Table \ref{tab:datasets-used}). The benefit of both dynamic histograms and vectorization scales with the number of samples (rows). Thus, high count, low-dimensional datasets are a reasonable proxy for large-wide data. The GPU acceleration also scales with the number features and benefits from data that are both wide and large. We  include synthetic Trunk data~\cite{trunk}: a generator that scales features and samples according to a $p$-dimensional multivariate Gaussian, while generating 2 balanced classes \cite{tomita2020sparse}.

% For accuracy experiments, we include data from the OpenML CC18 benchmarks~\cite{bischl2017openml} that include categorical and text data. None of these are large enough for meaningful performance experiments.

\subsection{Dataset Scalability}
\label{sec:res-scalability}

\begin{figure*}[t]
    \centering
    % Generated by benchmarks/src/spo_vs_gbt/make_rowcol_map_depths.py -- regenerate, never hand-edit.
    \input{figures/results/speedup_heatmap_rows_features.tex}
    \caption{Speedup heatmap by dataset dimensions, simulated using Trunk synthetic data (circles), of our SPO-RF Dyn-Vec method over Exact. Also added are natural datasets HIGGS, SUSY, Epsilon, YouTube-8M, Shifts weather, ClimSim and Jane Street (squares). Each marker is labelled
    with its measured speedup. The background shading interpolates the speedup by
    inverse-distance weighting: every point averages its three nearest measured points, weighted by the inverse square
    of its distance.}
    \Description{Three scatter panels, one per maximum depth, with features on the x axis and rows on the y axis, both log scale; points colored and annotated by speedup over a shaded interpolated background.}
    \label{fig:speedup-heatmap}
\end{figure*}

Figure~\ref{fig:speedup-heatmap} places every dataset by rows and features and colors it by the speedup of vectorized dynamic histograms (AVX2, 64 bins) over Exact at depth 6, depth 16 and purity; the shading interpolates between the measured points and predicts the speedup for a dataset of a given size. Two regularities of Table~\ref{tab:train-time-by-depth} shape the map. First, the speedup grows with the number of rows at every depth. On the Trunk shapes at depth 6 it rises from 1.4$\times$ at 15k rows through 1.6$\times$ at 150k and 1.6--1.7$\times$ at 1--1.5M to 1.9$\times$ at 4.5M, and the 10M-row natural datasets reach 1.8--2.0$\times$; at purity the same progression reads 1.1$\times$, 1.2$\times$, 1.2--1.3$\times$, 1.25$\times$ and 1.2--1.5$\times$. Second, the number of features barely matters: with 150k rows the depth-6 speedup is 1.56, 1.56, 1.55 and 1.60$\times$ at 4096, 40k, 160k and 400k features, and with 1.5M rows 1.71, 1.71 and 1.69$\times$ at 4096, 16384 and 40k. The feature count scales the projection sampling and evaluation, which every split finder pays alike, whereas the choice between histogram and sort acts on the $n$ projected values of a node. The widest shapes, 15k$\times$400k and 15k$\times$1.6M, are the least favourable (1.4$\times$ at depth 6, 1.1$\times$ at purity) because their split search is a minor share of the runtime (Section~\ref{sec:res-depth6}).

% ---- previous text below (kept for comparison; v2 rewrite above, 2026-09-26) ----

% Examining speedup reveals that SO-YDF training is compute bound on the CPU. Figure \ref{fig:speedup} shows near perfect scaling for vectorized dynamic histograms. This is somewhat surprising given that deep trees perform sparse memory access to feature columns. The results at 32 threads on 16 cores show no additional speedup and actually lose a little performance, likely from threads interfering in processor caches. There is no scaling when the number of threads exceed the number of CPU cores. % actually there's a small 20% benefit from hyperthreading. not included in results

% When using the GPU, we are still CPU bound for this datasets. On this benchmark, GPU occupancy never exceeded 70\%. Larger datasets shift the bottleneck to the GPU.

\subsection{Ablation - Dynamic Vectorized Histograms}
\label{sec:dyn}

Table~\ref{tab:train-time-by-depth} contains the two single-change comparisons that separate the contributions, and the two bin counts side by side.

\paragraph{Vectorization alone} The scalar and vectorized binners with the same bin count build the same histograms and the same trees, so their ratio is a pure kernel speedup. On the natural datasets it is 1.2--1.8$\times$ (64 bins, AVX2) and 1.3--2.2$\times$ (256 bins, AVX-512) at depth 6 and 1.3--1.5$\times$ and 1.3--1.6$\times$ at purity; on Trunk 1.4--2.1$\times$ and 1.4--2.6$\times$. It never drops below 1.2$\times$ at any depth on any dataset. Vectorized histograms therefore beat scalar ones at every tree level; whether they beat Exact depends on the depth (Sections~\ref{sec:res-depth6}--\ref{sec:res-purity}).

\paragraph{Dynamic switching alone} With the binner held fixed, on the seven large natural datasets the switch changes nothing at depths 6 and 10 (within 3\%), adds up to 8\% at depth 16, 3--28\% at depth 24 (HIGGS 5\%, SUSY 8\%, Epsilon 28\%) and 6--31\% at purity with 64 bins. With 256 bins, whose all-histogram variant carries four times the bin setup and scan, it adds 1.15--2.1$\times$ at depth 24 and 1.3--2.2$\times$ at purity. Where every node has hundreds of rows the switch is idle and costs nothing; where nodes are small it recovers what histograms lose.

% \paragraph{64 vs.\ 256 bins} At depth 6 the AVX-512 256-bin binner matches the AVX2 64-bin one within 5\% (Tables~\ref{tab:train-time-gbt-depth6} and \ref{tab:train-time-by-depth}). Deeper, its all-histogram variant falls behind, 0.5--1.2$\times$ Exact at purity against 0.8--1.4$\times$ for 64 bins on the natural datasets. Figure~\ref{fig:AVX-512-slower} shows why: the compare-and-count lookup costs the same for both, but drawing, sorting and scanning four times as many bin boundaries is a per-projection cost that does not shrink with the node, so it dominates once nodes are small. The switch removes exactly that cost, and the two dynamic variants agree within 4\% on the large natural datasets and within 8\% on Trunk at every depth. Accuracy does not separate the bin counts either (Table~\ref{tab:hist-vs-exact-summary}), so with the switch in place the bin count is a free parameter.

% ---- previous text below (kept for comparison; v2 rewrite above, 2026-09-26) ----

\subsection{Dynamic Threshold Sensitivity Study}
\label{app:threshold-sensitivity}

\begin{figure}[t]
    \centering
    % Generated by benchmarks/src/utils/make_breakeven_figure.py -- do not hand-edit.
\definecolor{brk0}{HTML}{4285F4}
\definecolor{brk1}{HTML}{EA4335}
\definecolor{brk2}{HTML}{FBBC05}
\definecolor{brk3}{HTML}{EDA100}
\definecolor{brk4}{HTML}{E87BA4}
\definecolor{brk5}{HTML}{008300}
\definecolor{brk6}{HTML}{34A853}
\begin{tikzpicture}
\begin{axis}[width=\linewidth, height=5.2cm, xmode=log, xmin=100, xmax=10000,
  ymin=1.05, ymax=1.36, xtick={100,200,500,1000,2000,5000,10000}, log ticks with fixed point,
  xlabel={Dynamic Switch Threshold},
  ylabel={Speedup over Exact}, tick label style={font=\scriptsize},
  label style={font=\scriptsize}, grid=none, axis line style={gray!60}, tick style={draw=none},
  every axis plot/.append style={line width=0.7pt, mark size=1.1pt},
  legend style={font=\scriptsize, draw=none, fill=none, at={(0.5,1.02)}, anchor=south, legend columns=3,
    /tikz/every even column/.append style={column sep=4pt}}, legend cell align=left,
  scaled y ticks=false, yticklabel style={/pgf/number format/fixed, /pgf/number format/precision=2}]
\fill[gray!14] (axis cs:800,1.05) rectangle (axis cs:1200,1.36);
\fill[gray!30] (axis cs:300,1.05) rectangle (axis cs:500,1.36);
\addplot[gray!70, dashed, line width=0.6pt, forget plot] coordinates {(100,1) (10000,1)};
\addplot[color=brk0, mark=none, solid, forget plot] coordinates {(100,1.1750) (200,1.2216) (300,1.2491) (400,1.2501) (500,1.2514) (600,1.2515) (700,1.2535) (800,1.2544) (900,1.2547) (1000,1.2523) (1200,1.2563) (1400,1.2546) (1600,1.2534) (1800,1.2496) (2000,1.2466) (2500,1.2471) (3000,1.2427) (3500,1.2390) (4000,1.2319) (4500,1.2313) (5000,1.2311) (6000,1.2328) (7000,1.2304) (8000,1.2195) (9000,1.2305) (10000,1.2257)};
\addplot[color=black, mark=x, mark size=2.6pt, only marks, forget plot] coordinates {(1200,1.2563)};
\addplot[color=brk1, mark=none, solid, forget plot] coordinates {(100,1.2072) (200,1.2639) (300,1.2808) (400,1.2819) (500,1.2978) (600,1.2987) (700,1.2998) (800,1.3006) (900,1.2956) (1000,1.2990) (1200,1.2912) (1400,1.2951) (1600,1.2929) (1800,1.2906) (2000,1.2891) (2500,1.2805) (3000,1.2786) (3500,1.2731) (4000,1.2714) (4500,1.2721) (5000,1.2695) (6000,1.2630) (7000,1.2564) (8000,1.2532) (9000,1.2411) (10000,1.2480)};
\addplot[color=black, mark=x, mark size=2.6pt, only marks, forget plot] coordinates {(800,1.3006)};
\addplot[color=brk2, mark=none, solid, forget plot] coordinates {(100,1.0690) (200,1.1397) (300,1.1691) (400,1.1758) (500,1.1836) (600,1.1875) (700,1.1900) (800,1.1877) (900,1.1943) (1000,1.1931) (1200,1.1873) (1400,1.1918) (1600,1.1853) (1800,1.1842) (2000,1.1821) (2500,1.1620) (3000,1.1550) (3500,1.1751) (4000,1.1712) (4500,1.1633) (5000,1.1663) (6000,1.1609) (7000,1.1578) (8000,1.1536) (9000,1.1527) (10000,1.1501)};
\addplot[color=black, mark=x, mark size=2.6pt, only marks, forget plot] coordinates {(900,1.1943)};
\addplot[color=brk6, mark=none, solid, forget plot] coordinates {(100,1.2589) (200,1.2890) (300,1.3010) (400,1.3129) (500,1.3131) (600,1.3150) (700,1.3125) (800,1.3187) (900,1.3205) (1000,1.3085) (1200,1.3160) (1400,1.3053) (1600,1.3077) (1800,1.3068) (2000,1.3009) (2500,1.2978) (3000,1.2940) (3500,1.2923) (4000,1.2914) (4500,1.2884) (5000,1.2863) (6000,1.2873) (7000,1.2785) (8000,1.2742) (9000,1.2741) (10000,1.2698)};
\addplot[color=black, mark=x, mark size=2.6pt, only marks, forget plot] coordinates {(900,1.3205)};
\addplot[color=brk0, mark=none, dash pattern=on 2pt off 1.5pt, forget plot] coordinates {(100,1.2521) (200,1.2671) (300,1.2742) (400,1.2745) (500,1.2701) (600,1.2676) (700,1.2616) (800,1.2672) (900,1.2641) (1000,1.2571) (1200,1.2598) (1400,1.2553) (1600,1.2491) (1800,1.2473) (2000,1.2481) (2500,1.2398) (3000,1.2362) (3500,1.2364) (4000,1.2321) (4500,1.2270) (5000,1.2330) (6000,1.2240) (7000,1.2232) (8000,1.2200) (9000,1.2256) (10000,1.2168)};
\addplot[color=black, mark=x, mark size=2.6pt, only marks, forget plot] coordinates {(400,1.2745)};
\addplot[color=brk1, mark=none, dash pattern=on 2pt off 1.5pt, forget plot] coordinates {(100,1.3136) (200,1.3289) (300,1.3359) (400,1.3319) (500,1.3281) (600,1.3254) (700,1.3105) (800,1.3166) (900,1.3117) (1000,1.3149) (1200,1.3056) (1400,1.3005) (1600,1.3012) (1800,1.3005) (2000,1.2943) (2500,1.2891) (3000,1.2805) (3500,1.2784) (4000,1.2750) (4500,1.2648) (5000,1.2601) (6000,1.2601) (7000,1.2555) (8000,1.2485) (9000,1.2468) (10000,1.2487)};
\addplot[color=black, mark=x, mark size=2.6pt, only marks, forget plot] coordinates {(300,1.3359)};
\addplot[color=brk2, mark=none, dash pattern=on 2pt off 1.5pt, forget plot] coordinates {(100,1.1960) (200,1.2175) (300,1.2212) (400,1.2237) (500,1.2208) (600,1.2148) (700,1.2163) (800,1.2086) (900,1.2059) (1000,1.2052) (1200,1.2017) (1400,1.1947) (1600,1.1931) (1800,1.1941) (2000,1.1852) (2500,1.1832) (3000,1.1818) (3500,1.1737) (4000,1.1658) (4500,1.1727) (5000,1.1657) (6000,1.1632) (7000,1.1583) (8000,1.1540) (9000,1.1513) (10000,1.1481)};
\addplot[color=black, mark=x, mark size=2.6pt, only marks, forget plot] coordinates {(400,1.2237)};
\addplot[color=brk6, mark=none, dash pattern=on 2pt off 1.5pt, forget plot] coordinates {(100,1.3162) (200,1.3314) (300,1.3233) (400,1.3267) (500,1.3317) (600,1.3250) (700,1.3244) (800,1.3223) (900,1.3251) (1000,1.3248) (1200,1.3161) (1400,1.3107) (1600,1.3210) (1800,1.3214) (2000,1.3091) (2500,1.3052) (3000,1.2974) (3500,1.2988) (4000,1.2925) (4500,1.2925) (5000,1.2955) (6000,1.2911) (7000,1.2774) (8000,1.2832) (9000,1.2824) (10000,1.2781)};
\addplot[color=black, mark=x, mark size=2.6pt, only marks, forget plot] coordinates {(500,1.3317)};
\addlegendimage{color=brk0, solid, line width=0.7pt, mark=none}
\addlegendentry{HIGGS}
\addlegendimage{color=brk1, solid, line width=0.7pt, mark=none}
\addlegendentry{SUSY}
\addlegendimage{color=black, dash pattern=on 2pt off 1.5pt, line width=0.7pt, mark=none}
\addlegendentry{AVX2}
\addlegendimage{color=brk2, solid, line width=0.7pt, mark=none}
\addlegendentry{Epsilon}
\addlegendimage{color=brk6, solid, line width=0.7pt, mark=none}
\addlegendentry{YouTube-8M}
\addlegendimage{color=black, solid, line width=0.7pt, mark=none}
\addlegendentry{AVX-512}
\end{axis}
\end{tikzpicture}
    \caption{Full-depth train speedup over Exact by a linesearch over Dynamic Thresholds. Above the threshold, Random-Vectorized is used, and below, Exact. Shaded bands: the range of the AVX-512 maxima (light, 800--1200 rows) and of the AVX2 maxima (dark, 300--500 rows).}
    \Description{Line chart of training speedup over all-exact against the switch threshold on a logarithmic axis from 100 to 10000 for four datasets and two vectorized binners, with an X marking each curve's maximum and two shaded bands marking the AVX-512 and AVX2 maxima ranges.}
    \label{fig:threshold-sensitivity}
\end{figure}

Figure~\ref{fig:threshold-sensitivity} varies the switch threshold $\tau$ from 100 to 10000 rows on HIGGS, SUSY, Epsilon and YouTube-8M at full depth. The optimum lies at 300--500 rows for AVX2 with 64 bins (1.22--1.34$\times$) and at 800--1200 rows for AVX-512 with 256 bins (1.19--1.32$\times$): the larger bin count raises the fixed cost per histogram and with it the node size at which a histogram beats a sort. The curves are flat. Every AVX2 curve stays within 2\% of its maximum between 200 and 1000 rows and every AVX-512 curve between 400 and 2500; pushing $\tau$ to 10000, which sorts all but the largest nodes, still retains 93--98\% of the best speedup, and $\tau=100$ costs 1--2\% (AVX2) and 5--10\% (AVX-512). The fixed values used throughout, $\tau=250$ (AVX2) and $\tau=1000$ (AVX-512), are within 1\% of every per-dataset optimum, so no per-machine tuning was needed for the results of this section. %The scalar 64-bin binner moves the breakeven to about 4600 rows (sweep not shown), which is why it is reported only in its all-histogram form. no, this threshold is >50k. dropped

% \subsection{\TODO{Roofline} Pre-Vec v. Vec (std::sort \& Random v. HWY \& Vec Random}

% AL add. Optionally add Roofline of May 2025

% auto-generated by analyze.py:make_paper_tables -- regenerate, do not hand-edit
\begin{table*}[t]
\centering
\footnotesize
\setlength{\tabcolsep}{2.2pt}
\caption{Training time comparison to popular libraries using GBT (Boosting) and Random Forests (bagging). These are not like-for-like comparisons (see footnote), but are included in expectation of high interest in such a comparison.}
% Important Ariel Sept-24-2026. Confirmed these numbers. XGboost says "XGBRFClassifier is deprecated and "does not implement a conventional random forest"
% LGBM runs single-threaded, parallelizes within a tree
\label{tab:train-time-vs-libraries}
\begin{tabular}{l|cc|cccc|cccc}
\hline
\multicolumn{11}{l}{\textit{Random Forest (full-depth)}} \\
\hline
\multirow{3}{*}{\textbf{Dataset}} & \multicolumn{2}{c|}{\textbf{SPO-RF Baselines}} & \multicolumn{4}{c|}{\textbf{SPO-RF Our Methods}} & \multicolumn{4}{c}{\textbf{Other RF}} \\
 & \textbf{Exact} & \textbf{Random Hist.} & \textbf{Vec. Random Hist.} & \textbf{Vec. Random Hist.} & \textbf{Vec. Dynamic Hist.} & \textbf{Vec. Dynamic Hist.} & \textbf{AA-RF} & \textbf{XGBoost}$^\dagger$ & \textbf{LightGBM}$^\dagger$ & \textbf{CatBoost}$^\dagger$ \\
 & \scriptsize HWY & \scriptsize scalar, 64 bins & \scriptsize AVX2, 64 bins & \scriptsize AVX-512, 256 bins & \scriptsize AVX2, 64 bins \& HWY & \scriptsize AVX-512, 256 bins \& HWY & \scriptsize Exact & \scriptsize RF mode & \scriptsize RF mode & \scriptsize RF mode \\
\hline
HIGGS & 363.0 & 524.2 & 382.3 & 560.1 & 296.2 & \textbf{287.5} & 292.4 & 3000.9 & 2598.8 & --- \\
SUSY & 119.6 & 171.3 & 119.4 & 175.6 & 91.8 & \textbf{90.2} & 104.3 & 1048.6 & 2281.1 & --- \\
Epsilon & 59.1 & 95.1 & 68.6 & 109.8 & \textbf{50.5} & 50.7 & 52.1 & OOM & OOM & --- \\
\hline
\end{tabular}
\par\vspace{6pt}
\begin{tabular}{l|cc|cccc|cccc}
\hline
\multicolumn{11}{l}{\textit{Gradient Boosted Trees (depth 6)}} \\
\hline
\multirow{3}{*}{\textbf{Dataset}} & \multicolumn{2}{c|}{\textbf{SPO-GBT Baselines}} & \multicolumn{4}{c|}{\textbf{SPO-GBT Our Methods}} & \multicolumn{4}{c}{\textbf{Other GBT}} \\
 & \textbf{Exact} & \textbf{Random Hist.} & \textbf{Vec. Random Hist.} & \textbf{Vec. Random Hist.} & \textbf{Vec. Dynamic Hist.} & \textbf{Vec. Dynamic Hist.} & \textbf{AA-GBT}$^*$ & \textbf{XGBoost}$^*$ & \textbf{LightGBM}$^*$ & \textbf{CatBoost}$^*$ \\
 & \scriptsize HWY & \scriptsize scalar, 64 bins & \scriptsize AVX2, 64 bins & \scriptsize AVX-512, 256 bins & \scriptsize AVX2, 64 bins \& HWY & \scriptsize AVX-512, 256 bins \& HWY & \scriptsize Exact & \scriptsize  & \scriptsize  & \scriptsize  \\
\hline
HIGGS & 2621.9 & 4012.8 & 1810.8 & 1768.0 & --- & --- & 746.6 & 21.8 & \textbf{18.2} & 35.9 \\
SUSY & 1018.9 & 1508.8 & 691.5 & 662.5 & --- & --- & 267.6 & 7.1 & \textbf{6.1} & 14.7 \\
Epsilon & 540.9 & 841.8 & 380.3 & 422.4 & --- & --- & 490.0 & 56.1 & 52.7 & \textbf{24.0} \\
\hline
\end{tabular}
\\[2pt]
\parbox{0.92\linewidth}{\footnotesize $^\dagger$ RF implementations of this method are not directly comparable. XGBoost 3.4.1 warns that \texttt{XGBRFClassifier} is deprecated and ``does not implement a conventional random forest'', and LightGBM only parallelizes within a tree, similar to YDF-GBT, even in Random Forest mode. This explains the slow performance of the two methods. CatBoost does not support Random Forest (Bagged) mode.\\
$^*$ These methods benefit from features being fixed throughout training (no random projections), and can use optimizations like pre-sorting. It is unclear to us why YDF's AA-GBT does not benefit to a similar extent.}
\end{table*}

% tab:train-time-vs-libraries is input from comparison_with_libraries.tex, next to the
% Training time discussion; these two are accuracy only.
% Dropped 2026-09-09 (user request): mean accuracy/AUC + mean-rank suite table.
%\input{tables/table_auc_vs_libraries_summary}
% \input{tables/table_auc_vs_libraries_large_datasets}

% \FloatBarrier
\subsection{Accuracy}
\label{sec:res-accuracy}

% \TODO{SPO-GBT mistakenly used $\sqrt{\mathbf{F}}$ projections instead of $\mathbf{F}$, putting it at an unfair disadvantage}

Vectorization changes no model: the scalar and vectorized binners bin identically and produce identical forests, which is why their accuracy columns coincide in Table~\ref{tab:auc-vs-libraries-per-dataset-auc}. The dynamic switch does change the model, since nodes below $\tau$ take an exact split where an all-histogram forest would take a binned one. Table~\ref{tab:hist-vs-exact-summary} measures the effect on held-out accuracy against Exact on 34 OpenML CC18 binary tasks (10-fold cross-validation $\times$ 3 seeds) plus HIGGS and SUSY. For SPO-RF the mean difference of the dynamic forest to Exact is $+5\times10^{-3}$ with a 95\% interval of $[-4, +14]\times10^{-3}$ that includes zero; it wins on 13 datasets, ties on 17 and loses on 6, and only two per-dataset differences survive Holm correction. For SPO-GBT the difference is $+1\times10^{-3}$ $[-1, +3]$ (9 wins, 20 ties, 7 losses), with electricity and adult significantly below Exact by 1.6 and 0.1 points. The dynamic forests are non-inferior to Exact within $2.5\times10^{-3}$ (RF) and $1\times10^{-3}$ (GBT) accuracy at $\alpha=0.05$, and the all-histogram forests fall in the same intervals. On the large datasets the SPO-RF finders agree to three decimals (HIGGS 0.762 and SUSY 0.802 for Exact and Dynamic alike) and the SPO-GBT ones within 0.2 points. Against XGBoost, LightGBM, CatBoost and axis-aligned YDF (Table ~\ref{tab:train-time-vs-libraries}, ~\ref{tab:auc-vs-libraries-per-dataset-auc}), the RF and GBT learners are very similar, and the dynamic vectorized configuration ties the exact baseline on 34 of 36 datasets.

% \input{comparison_with_libraries}

% Auto-generated by make_paper_accuracy_tables.py -- do not edit.
% Deltas are (arm - Exact) held-out test accuracy in units of 10^-3;
% positive = histogram arm better. Aggregates over per-dataset means of
% paired per-(seed,fold) deltas; margins = smallest margin passing at alpha=0.05.
\begin{table*}
\centering
\caption{Held-out test accuracy of random-histogram split finding vs.\ exact splits, aggregated over 36 datasets (34 OpenML CC18 binary tasks, 10-fold CV $\times$ 3 seeds + HIGGS and SUSY). $\Delta$ = method $-$ Exact in units of $10^{-3}$ accuracy; CIs are per-dataset paired $t$ intervals. W/T/L counts Wins/Ties/Losses ($|\Delta| \le 10^{-3}$ is a tie). LOO $\Delta$ = leave-one-out, dropping the most influential dataset. Margin = smallest one-sided non-inferiority margin passing at $\alpha{=}0.05$; Sig.\ = datasets with a Holm-corrected per-dataset Wilcoxon difference.}
\label{tab:hist-vs-exact-summary}
\begin{tabular}{llrccccc}
\toprule
Learner & Split finder & $n$ & $\Delta$ [95\% CI] ($\times 10^{-3}$) & LOO $\Delta$ Acc. [95\% CI] & W/T/L & Margin ($\times 10^{-3}$) & Sig. \\
\midrule
SPO-RF & Dynamic (64 bins, $\tau{=}250$) & 36 & +5.2 [-3.7, +14.2] & +0.8 [+0.0, +1.7] & 13/17/6 & 2.5 & 2 \\
 & Random, 64 bins & 36 & +6.5 [-5.6, +18.5] & +0.6 [-0.4, +1.6] & 12/19/5 & 5 & 2 \\
 & Random, 256 bins & 34 & +7.3 [-5.4, +20.1] & +1.1 [-0.0, +2.3] & 12/15/7 & 5 & 0 \\
\midrule
SPO-GBT & Dynamic (64 bins, $\tau{=}250$) & 36 & +0.8 [-1.9, +3.6] & -0.1 [-2.1, +1.9] & 11/19/6 & 2.5 & 2 \\
 & Random, 64 bins & 36 & +0.7 [-1.8, +3.1] & -0.4 [-1.8, +1.1] & 11/17/8 & 2.5 & 2 \\
 & Random, 256 bins & 34 & +0.8 [-2.2, +3.7] & -0.5 [-2.0, +1.0] & 11/13/10 & 2.5 & 0 \\
\bottomrule
\end{tabular}
\end{table*}

% \FloatBarrier

\section{Conclusions and Discussion}

We overhaul sparse oblique (SPO) training in YDF. Fixing inefficiencies in the released implementation and adopting Highway VQSort raised the baseline by 2--5$\times$, and against it we contribute two novel methods. AVX2 and AVX-512 vectorized histogram filling speeds up depth-6 SPO-GBT by 1.3--1.6$\times$ and SPO-RF by up to 2$\times$ on the large natural datasets, where scalar histograms lose to sorting. Dynamic switching between vectorized histograms and exact splits by node size retains 1.2--1.6$\times$ from depth 16 to purity, where histogram-only methods fall behind Exact, with no detectable effect on accuracy on 36 datasets. Together they bring SPO-RF training time on par with YDF's axis-aligned RF. Both act on node size rather than depth or model family, so they extend to axis-aligned splits without pre-sorting; batching the many small nodes of deep trees into single GPU kernels is future work.

\begin{acks}
\textbf{Use of generative AI.} In accordance with the ACM Policy on Authorship,
we disclose that generative AI tools were used to assist with
editing and revising the prose of this paper and with developing the
experimental infrastructure (benchmark setup, data-preparation and
result-processing scripts, and plotting code). All methods, algorithms, and
optimizations described in this paper were conceived and designed solely by the
authors. The authors verified all AI-assisted code and text and take full
responsibility for the accuracy of the content, references, data, and figures.
\end{acks}
% This design is not suitable for a full GPU-based implementation of training. Launching kernels for  nodes with few active samples. We are exploring techniques to run multiple nodes in a single kernel to realize a full-offloading to the GPU.

\bibliographystyle{plain}
\bibliography{main}

\appendix

\section{Appendix}

% Datasets table first: it is the only [t] float here, so it shares a page with the text spill-over.
\begin{table*}[t]
\centering
\caption{Datasets used, by source. Categorical columns are numbered ordinally, and NaNs are column mean-imputed. Task: bin = binary classification, mul = multi-class classification, reg = regression. Multi-class classification problems was converted to 1-vs-all binary classification. Eight datasets appear in both CC18 and TabArena. $^\dagger$ datasets were truncated to fit into memory.}
\label{tab:all-datasets}
\footnotesize
\setlength{\tabcolsep}{4pt}
\begin{tabular}[t]{@{}l c r r@{}}
\toprule
Dataset & Task & Rows & Feat. \\
\midrule
\multicolumn{4}{@{}l}{\textit{Large Natural Datasets}} \\
HIGGS~\cite{higgs_280} & bin & 11,000,000 & 28 \\
SUSY~\cite{susy_279} & bin & 5,000,000 & 18 \\
Epsilon~\cite{chang2011libsvm} & bin & 400,000 & 2,000 \\
YouTube-8M~\cite{abu2016youtube}         & mul$^*$   & 3,888,919            & 1152 \\
Shifts Weather~\cite{malinin2021shifts}   & reg$^*$   & 3,129,592            & 127 \\
ClimSim~\cite{yu2023climsim}              & reg$^*$   & 10,091,520$^\dagger$           & 124 \\
Jane Street~\cite{jane-street-real-time-market-data-forecasting}       & reg$^*$   & 10,213,190$^\dagger$ & 82
\\\\
\multicolumn{4}{@{}l}{\textit{Synthetic}} \\
Trunk~\cite{trunk} & bin & - & - \\\\
\multicolumn{4}{@{}l}{\textit{OpenML CC18~\cite{bischl2017openml}}} \\
adult & bin & 48,842 & 14 \\
bank-marketing & bin & 45,211 & 16 \\
banknote-authentication & bin & 1,372 & 4 \\
Bioresponse & bin & 3,751 & 1,776 \\
blood-transfusion-service-center & bin & 748 & 4 \\
breast-w & bin & 699 & 9 \\
churn & bin & 5,000 & 20 \\
climate-model-simulation-crashes & bin & 540 & 18 \\
credit-approval & bin & 690 & 15 \\
credit-g & bin & 1,000 & 20 \\
cylinder-bands & bin & 540 & 37 \\
diabetes & bin & 768 & 8 \\
dresses-sales & bin & 500 & 12 \\
electricity & bin & 45,312 & 8 \\
ilpd & bin & 583 & 10 \\
Internet-Advertisements & bin & 3,279 & 1,558 \\
jm1 & bin & 10,885 & 21 \\
kc1 & bin & 2,109 & 21 \\
kc2 & bin & 522 & 21 \\
kr-vs-kp & bin & 3,196 & 36 \\
madelon & bin & 2,600 & 500 \\
nomao & bin & 34,465 & 118 \\
numerai28.6 & bin & 96,320 & 21 \\
ozone-level-8hr & bin & 2,534 & 72 \\
pc1 & bin & 1,109 & 21 \\
pc3 & bin & 1,563 & 37 \\
pc4 & bin & 1,458 & 37 \\
PhishingWebsites & bin & 11,055 & 30 \\
phoneme & bin & 5,404 & 5 \\
qsar-biodeg & bin & 1,055 & 41 \\
spambase & bin & 4,601 & 57 \\
tic-tac-toe & bin & 958 & 9 \\
wdbc & bin & 569 & 30 \\
wilt & bin & 4,839 & 5 \\
\bottomrule
\end{tabular}
\hspace{1.5em}
\begin{tabular}[t]{@{}l c r r@{}}
\toprule
Dataset & Task & Rows & Feat. \\
\midrule
\multicolumn{4}{@{}l}{\textit{TabArena~\cite{tabarena}}} \\
airfoil\_self\_noise & reg & 1,503 & 4 \\
used-Fiat-500 & reg & 1,538 & 6 \\
APSFailure & bin & 76,000 & 170 \\
bank-marketing & bin & 45,211 & 5 \\
Bank\_Customer\_Churn & bin & 10,000 & 6 \\
Bioresponse & bin & 3,751 & 1,776 \\
blood-transfusion & bin & 748 & 4 \\
churn & bin & 5,000 & 15 \\
coil2000\_insurance\_policies & bin & 9,822 & 80 \\
concrete\_compressive & reg & 1,030 & 8 \\
credit-g & bin & 1,000 & 7 \\
credit\_card\_default & bin & 30,000 & 20 \\
customer\_satisf.\_airline & bin & 129,880 & 5 \\
diabetes & bin & 768 & 8 \\
Diabetes130US & bin & 71,518 & 8 \\
diamonds & reg & 53,940 & 6 \\
E-CommereShippingData & bin & 10,999 & 6 \\
Fitness\_Club & bin & 1,500 & 3 \\
Food\_Delivery\_Time & reg & 45,451 & 6 \\
GiveMeSomeCredit & bin & 150,000 & 10 \\
hazelnut-spread-contam. & bin & 2,400 & 30 \\
healthcare\_insurance & reg & 1,338 & 3 \\
heloc & bin & 10,459 & 23 \\
houses & reg & 20,640 & 8 \\
HR\_Analytics\_Job\_Change & bin & 19,158 & 2 \\
in\_vehicle\_coupon & bin & 12,684 & 2 \\
Is-this-a-good-customer & bin & 1,723 & 5 \\
jm1 & bin & 10,885 & 21 \\
Marketing\_Campaign & bin & 2,240 & 16 \\
miami\_housing & reg & 13,776 & 14 \\
online\_shoppers\_intention & bin & 12,330 & 10 \\
physiochemical\_protein & reg & 45,730 & 9 \\
polish\_bankruptcy & bin & 5,910 & 64 \\
qsar-biodeg & bin & 1,054 & 36 \\
QSAR-TID-11 & reg & 5,742 & 1,024 \\
QSAR\_fish\_toxicity & reg & 907 & 6 \\
seismic-bumps & bin & 2,584 & 11 \\
superconductivity & reg & 21,263 & 81 \\
taiwanese\_bankruptcy & bin & 6,819 & 94 \\
wine\_quality & reg & 6,497 & 11 \\\\
\multicolumn{4}{@{}l}{\textit{TabReD~\cite{tabred}}} \\
cooking-time & reg & 319,986 & 186 \\
ecom-offers & bin & 160,057 & 113 \\
homesite-insurance & bin & 260,753 & 253 \\
weather & reg & 423,795 & 100 \\
\bottomrule
\end{tabular}
\label{tab:datasets-used}
\end{table*}

% Generated by benchmarks/evaluation/spo_vs_gbt/make_overall_depth_table.py -- regenerate, don't edit.
% Generated by benchmarks/src/spaa27_experiments/make_overall_depth_table.py -- do not hand-edit.
\begin{table*}[p]
    \centering
    \footnotesize
    \setlength{\tabcolsep}{2.5pt}
    \begin{tabular}{l|ccc|cccc}
        \hline
        \multirow{3}{*}{\textbf{Dataset}} & \multicolumn{3}{c|}{\textbf{Baselines}} & \multicolumn{4}{c}{\textbf{Our Methods}} \\
         & \textbf{Exact} & \textbf{Random Hist.} & \textbf{Random Hist.} & \textbf{Vec. Random Hist.} & \textbf{Vec. Random Hist.} & \textbf{Vec. Dynamic Hist.} & \textbf{Vec. Dynamic Hist.} \\
         & \scriptsize HWY & \scriptsize scalar, 64 bins & \scriptsize scalar, 256 bins & \scriptsize AVX2, 64 bins & \scriptsize AVX-512, 256 bins & \scriptsize AVX2, 64 bins \& HWY & \scriptsize AVX-512, 256 bins \& HWY \\
        \hline
        \multicolumn{8}{l}{\textit{Depth 6}} \\
        HIGGS 10.5M$\times$28 & 79.6 & 59.3 & 70.1 & 43.3 (1.8$\times$) & 43.9 (1.8$\times$) & 44.7 (1.8$\times$) & 44.3 (1.8$\times$) \\
        SUSY 4.5M$\times$18 & 27.0 & 21.3 & 25.4 & 14.9 (1.8$\times$) & 14.9 (1.8$\times$) & 14.9 (1.8$\times$) & 15.1 (1.8$\times$) \\
        Epsilon 400k$\times$2000 & 12.0 & 13.6 & 17.3 & 7.4 (1.6$\times$) & 7.9 (1.5$\times$) & 7.4 (1.6$\times$) & 7.5 (1.6$\times$) \\
        GiveMeSomeCredit 150k$\times$10 & 0.3 & 0.4 & 0.4 & 0.3 (1.1$\times$) & 0.3 (1.1$\times$) & 0.3 (1.0$\times$) & 0.3 (1.2$\times$) \\
        YouTube-8M 3.9M$\times$1152 & 137 & 105 & 127 & 70.7 (1.9$\times$) & 74.6 (1.8$\times$) & 70.9 (1.9$\times$) & 74.1 (1.8$\times$) \\
        Shifts weather 3.1M$\times$127 & 34.3 & 27.9 & 33.2 & 20.2 (1.7$\times$) & 20.0 (1.7$\times$) & 20.0 (1.7$\times$) & 19.7 (1.7$\times$) \\
        ClimSim 10.1M$\times$124 & 146 & 98.2 & 117 & 72.7 (2.0$\times$) & 72.9 (2.0$\times$) & 73.2 (2.0$\times$) & 71.9 (2.0$\times$) \\
        Jane Street 10.2M$\times$82 & 121 & 74.5 & 86.7 & 60.2 (2.0$\times$) & 60.7 (2.0$\times$) & 60.1 (2.0$\times$) & 60.1 (2.0$\times$) \\
        Trunk 1M$\times$32 & 5.2 & 5.2 & 6.3 & 3.2 (1.6$\times$) & 3.3 (1.6$\times$) & 3.3 (1.6$\times$) & 3.2 (1.6$\times$) \\
        Trunk 1M$\times$512 & 18.8 & 20.0 & 22.8 & 11.2 (1.7$\times$) & 11.1 (1.7$\times$) & 10.9 (1.7$\times$) & 11.0 (1.7$\times$) \\
        Trunk 1M$\times$2048 & 35.1 & 36.1 & 45.1 & 21.2 (1.7$\times$) & 21.8 (1.6$\times$) & 20.9 (1.7$\times$) & 21.6 (1.6$\times$) \\
        Trunk 15k$\times$400k & 5.0 & 7.3 & 9.7 & 3.5 (1.4$\times$) & 3.9 (1.3$\times$) & 3.5 (1.4$\times$) & 3.5 (1.4$\times$) \\
        Trunk 15k$\times$1.6M & 8.7 & 12.0 & 16.0 & 6.1 (1.4$\times$) & 6.7 (1.3$\times$) & 6.0 (1.5$\times$) & 6.3 (1.4$\times$) \\
        Trunk 150k$\times$4096 & 5.6 & 7.2 & 9.3 & 3.6 (1.6$\times$) & 3.8 (1.5$\times$) & 3.6 (1.5$\times$) & 3.9 (1.4$\times$) \\
        Trunk 150k$\times$40k & 17.2 & 22.0 & 28.8 & 11.2 (1.5$\times$) & 11.8 (1.5$\times$) & 11.0 (1.6$\times$) & 11.4 (1.5$\times$) \\
        Trunk 150k$\times$160k & 32.9 & 43.4 & 56.2 & 21.2 (1.6$\times$) & 21.7 (1.5$\times$) & 21.2 (1.6$\times$) & 22.4 (1.5$\times$) \\
        Trunk 150k$\times$400k & 51.8 & 67.5 & 88.1 & 32.5 (1.6$\times$) & 34.1 (1.5$\times$) & 32.4 (1.6$\times$) & 33.7 (1.5$\times$) \\
        Trunk 1.5M$\times$4096 & 78.1 & 75.6 & 94.7 & 45.8 (1.7$\times$) & 47.3 (1.7$\times$) & 45.8 (1.7$\times$) & 47.4 (1.6$\times$) \\
        Trunk 1.5M$\times$16384 & 155 & 150 & 187 & 89.8 (1.7$\times$) & 90.2 (1.7$\times$) & 90.9 (1.7$\times$) & 91.6 (1.7$\times$) \\
        Trunk 1.5M$\times$40k & 243 & 234 & 293 & 145 (1.7$\times$) & 145 (1.7$\times$) & 144 (1.7$\times$) & 140 (1.7$\times$) \\
        Trunk 4.5M$\times$4096 & 296 & 236 & 290 & 158 (1.9$\times$) & 159 (1.9$\times$) & 158 (1.9$\times$) & 163 (1.8$\times$) \\
        \hline
        \multicolumn{8}{l}{\textit{Depth 10}} \\
        HIGGS 10.5M$\times$28 & 141 & 116 & 138 & 89.7 (1.6$\times$) & 91.5 (1.5$\times$) & 90.1 (1.6$\times$) & 91.2 (1.5$\times$) \\
        SUSY 4.5M$\times$18 & 46.1 & 40.8 & 49.1 & 28.5 (1.6$\times$) & 28.6 (1.6$\times$) & 28.4 (1.6$\times$) & 28.5 (1.6$\times$) \\
        Epsilon 400k$\times$2000 & 23.7 & 28.2 & 35.6 & 16.7 (1.4$\times$) & 17.7 (1.3$\times$) & 16.8 (1.4$\times$) & 17.1 (1.4$\times$) \\
        GiveMeSomeCredit 150k$\times$10 & 0.5 & 0.6 & 0.8 & 0.5 (1.0$\times$) & 0.5 (1.0$\times$) & 0.4 (1.1$\times$) & 0.5 (1.1$\times$) \\
        YouTube-8M 3.9M$\times$1152 & 235 & 211 & 251 & 151 (1.5$\times$) & 153 (1.5$\times$) & 151 (1.6$\times$) & 153 (1.5$\times$) \\
        Shifts weather 3.1M$\times$127 & 58.1 & 53.4 & 62.5 & 39.1 (1.5$\times$) & 39.0 (1.5$\times$) & 39.4 (1.5$\times$) & 38.6 (1.5$\times$) \\
        ClimSim 10.1M$\times$124 & 228 & 175 & 206 & 129 (1.8$\times$) & 127 (1.8$\times$) & 128 (1.8$\times$) & 127 (1.8$\times$) \\
        Jane Street 10.2M$\times$82 & 202 & 144 & 167 & 119 (1.7$\times$) & 120 (1.7$\times$) & 119 (1.7$\times$) & 119 (1.7$\times$) \\
        Trunk 1M$\times$32 & 8.7 & 9.6 & 11.9 & 5.7 (1.5$\times$) & 5.9 (1.5$\times$) & 5.8 (1.5$\times$) & 5.8 (1.5$\times$) \\
        Trunk 1M$\times$512 & 33.6 & 37.6 & 46.2 & 23.1 (1.5$\times$) & 23.7 (1.4$\times$) & 23.1 (1.5$\times$) & 23.1 (1.5$\times$) \\
        Trunk 1M$\times$2048 & 66.1 & 74.0 & 91.1 & 45.7 (1.4$\times$) & 46.6 (1.4$\times$) & 45.3 (1.5$\times$) & 46.2 (1.4$\times$) \\
        Trunk 15k$\times$400k & 10.6 & 15.9 & 21.8 & 9.1 (1.2$\times$) & 11.2 (0.9$\times$) & 8.3 (1.3$\times$) & 8.8 (1.2$\times$) \\
        Trunk 15k$\times$1.6M & 18.2 & 25.9 & 35.7 & 15.1 (1.2$\times$) & 18.9 (1.0$\times$) & 13.8 (1.3$\times$) & 14.8 (1.2$\times$) \\
        Trunk 150k$\times$4096 & 11.8 & 15.2 & 19.6 & 8.7 (1.4$\times$) & 9.3 (1.3$\times$) & 8.7 (1.4$\times$) & 8.9 (1.3$\times$) \\
        Trunk 150k$\times$40k & 36.6 & 46.5 & 59.6 & 26.8 (1.4$\times$) & 28.4 (1.3$\times$) & 26.4 (1.4$\times$) & 27.0 (1.4$\times$) \\
        Trunk 150k$\times$160k & 69.5 & 89.6 & 116 & 49.7 (1.4$\times$) & 52.8 (1.3$\times$) & 49.8 (1.4$\times$) & 50.3 (1.4$\times$) \\
        Trunk 150k$\times$400k & 107 & 137 & 180 & 74.1 (1.4$\times$) & 82.5 (1.3$\times$) & 73.8 (1.4$\times$) & 76.7 (1.4$\times$) \\
        Trunk 1.5M$\times$4096 & 143 & 155 & 192 & 99.3 (1.4$\times$) & 98.3 (1.5$\times$) & 98.0 (1.5$\times$) & 97.9 (1.5$\times$) \\
        Trunk 1.5M$\times$16384 & 282 & 304 & 378 & 190 (1.5$\times$) & 194 (1.5$\times$) & 190 (1.5$\times$) & 193 (1.5$\times$) \\
        Trunk 1.5M$\times$40k & 440 & 473 & 587 & 304 (1.4$\times$) & 309 (1.4$\times$) & 304 (1.4$\times$) & 300 (1.5$\times$) \\
        Trunk 4.5M$\times$4096 & 496 & 472 & 576 & 321 (1.5$\times$) & 324 (1.5$\times$) & 322 (1.5$\times$) & 328 (1.5$\times$) \\
        \hline
        \multicolumn{8}{l}{\textit{Depth 16}} \\
        HIGGS 10.5M$\times$28 & 211 & 204 & 243 & 150 (1.4$\times$) & 154 (1.4$\times$) & 150 (1.4$\times$) & 151 (1.4$\times$) \\
        SUSY 4.5M$\times$18 & 68.4 & 69.9 & 86.4 & 46.4 (1.5$\times$) & 48.1 (1.4$\times$) & 46.6 (1.5$\times$) & 46.4 (1.5$\times$) \\
        Epsilon 400k$\times$2000 & 41.8 & 55.3 & 78.8 & 35.2 (1.2$\times$) & 46.3 (0.9$\times$) & 32.5 (1.3$\times$) & 34.0 (1.2$\times$) \\
        GiveMeSomeCredit 150k$\times$10 & 0.8 & 1.1 & 1.4 & 0.8 (1.0$\times$) & 1.0 (0.8$\times$) & 0.8 (1.1$\times$) & 0.8 (1.1$\times$) \\
        YouTube-8M 3.9M$\times$1152 & 365 & 379 & 464 & 257 (1.4$\times$) & 266 (1.4$\times$) & 257 (1.4$\times$) & 260 (1.4$\times$) \\
        Shifts weather 3.1M$\times$127 & 82.5 & 84.8 & 103 & 60.6 (1.4$\times$) & 62.9 (1.3$\times$) & 60.5 (1.4$\times$) & 60.1 (1.4$\times$) \\
        ClimSim 10.1M$\times$124 & 271 & 232 & 276 & 168 (1.6$\times$) & 167 (1.6$\times$) & 167 (1.6$\times$) & 163 (1.7$\times$) \\
        Jane Street 10.2M$\times$82 & 301 & 258 & 304 & 205 (1.5$\times$) & 206 (1.5$\times$) & 205 (1.5$\times$) & 204 (1.5$\times$) \\
        Trunk 1M$\times$32 & 12.9 & 16.5 & 22.3 & 10.0 (1.3$\times$) & 12.2 (1.1$\times$) & 9.5 (1.4$\times$) & 9.6 (1.3$\times$) \\
        Trunk 1M$\times$512 & 56.1 & 70.3 & 93.8 & 44.7 (1.3$\times$) & 53.5 (1.0$\times$) & 42.7 (1.3$\times$) & 43.2 (1.3$\times$) \\
        Trunk 1M$\times$2048 & 114 & 141 & 187 & 89.9 (1.3$\times$) & 106 (1.1$\times$) & 86.5 (1.3$\times$) & 87.1 (1.3$\times$) \\
        Trunk 15k$\times$400k & 20.2 & 31.5 & 48.7 & 21.3 (0.9$\times$) & 31.1 (0.6$\times$) & 17.2 (1.2$\times$) & 18.1 (1.1$\times$) \\
        Trunk 15k$\times$1.6M & 36.4 & 52.3 & 79.1 & 36.2 (1.0$\times$) & 52.5 (0.7$\times$) & 30.9 (1.2$\times$) & 32.2 (1.1$\times$) \\
        Trunk 150k$\times$4096 & 21.5 & 31.3 & 46.8 & 20.4 (1.1$\times$) & 29.0 (0.7$\times$) & 17.7 (1.2$\times$) & 17.9 (1.2$\times$) \\
        Trunk 150k$\times$40k & 68.5 & 95.7 & 135 & 60.6 (1.1$\times$) & 81.7 (0.8$\times$) & 54.8 (1.2$\times$) & 56.3 (1.2$\times$) \\
        Trunk 150k$\times$160k & 135 & 183 & 257 & 115 (1.2$\times$) & 150 (0.9$\times$) & 107 (1.3$\times$) & 112 (1.2$\times$) \\
        Trunk 150k$\times$400k & 210 & 276 & 387 & 171 (1.2$\times$) & 216 (1.0$\times$) & 164 (1.3$\times$) & 168 (1.3$\times$) \\
        Trunk 1.5M$\times$4096 & 247 & 295 & 379 & 190 (1.3$\times$) & 212 (1.2$\times$) & 186 (1.3$\times$) & 187 (1.3$\times$) \\
        Trunk 1.5M$\times$16384 & 500 & 596 & 758 & 384 (1.3$\times$) & 425 (1.2$\times$) & 377 (1.3$\times$) & 380 (1.3$\times$) \\
        Trunk 1.5M$\times$40k & 779 & 918 & 1153 & 589 (1.3$\times$) & 641 (1.2$\times$) & 581 (1.3$\times$) & 585 (1.3$\times$) \\
        Trunk 4.5M$\times$4096 & 826 & 899 & 1115 & 597 (1.4$\times$) & 625 (1.3$\times$) & 595 (1.4$\times$) & 601 (1.4$\times$) \\
        \hline
    \end{tabular}
    \caption{(Part 1 of 2: depth 6, depth 10, depth 16.) End-to-end SPO-RF training time (s) by tree depth.}
    \label{tab:train-time-by-depth}
\end{table*}

\begin{table*}[p]
    \centering
    \ContinuedFloat
    \footnotesize
    \setlength{\tabcolsep}{2.5pt}
    \begin{tabular}{l|ccc|cccc}
        \hline
        \multirow{3}{*}{\textbf{Dataset}} & \multicolumn{3}{c|}{\textbf{Baselines}} & \multicolumn{4}{c}{\textbf{Our Methods}} \\
         & \textbf{Exact} & \textbf{Random Hist.} & \textbf{Random Hist.} & \textbf{Vec. Random Hist.} & \textbf{Vec. Random Hist.} & \textbf{Vec. Dynamic Hist.} & \textbf{Vec. Dynamic Hist.} \\
         & \scriptsize HWY & \scriptsize scalar, 64 bins & \scriptsize scalar, 256 bins & \scriptsize AVX2, 64 bins & \scriptsize AVX-512, 256 bins & \scriptsize AVX2, 64 bins \& HWY & \scriptsize AVX-512, 256 bins \& HWY \\
        \hline
        \multicolumn{8}{l}{\textit{Depth 20}} \\
        HIGGS 10.5M$\times$28 & 249 & 254 & 312 & 184 (1.3$\times$) & 194 (1.3$\times$) & 183 (1.4$\times$) & 185 (1.3$\times$) \\
        SUSY 4.5M$\times$18 & 80.6 & 86.3 & 110 & 57.3 (1.4$\times$) & 61.8 (1.3$\times$) & 55.9 (1.4$\times$) & 56.7 (1.4$\times$) \\
        Epsilon 400k$\times$2000 & 51.0 & 76.2 & 117 & 49.9 (1.0$\times$) & 76.3 (0.7$\times$) & 42.1 (1.2$\times$) & 43.9 (1.2$\times$) \\
        GiveMeSomeCredit 150k$\times$10 & 1.0 & 1.4 & 2.0 & 1.0 (0.9$\times$) & 1.4 (0.7$\times$) & 1.0 (1.0$\times$) & 0.9 (1.0$\times$) \\
        YouTube-8M 3.9M$\times$1152 & 438 & 479 & 615 & 326 (1.3$\times$) & 365 (1.2$\times$) & 320 (1.4$\times$) & 324 (1.4$\times$) \\
        Shifts weather 3.1M$\times$127 & 92.2 & 97.1 & 123 & 71.0 (1.3$\times$) & 78.2 (1.2$\times$) & 68.9 (1.3$\times$) & 69.5 (1.3$\times$) \\
        ClimSim 10.1M$\times$124 & 298 & 256 & 318 & 190 (1.6$\times$) & 196 (1.5$\times$) & 189 (1.6$\times$) & 187 (1.6$\times$) \\
        Jane Street 10.2M$\times$82 & 374 & 338 & 413 & 267 (1.4$\times$) & 278 (1.3$\times$) & 264 (1.4$\times$) & 267 (1.4$\times$) \\
        Trunk 1M$\times$32 & 13.6 & 19.0 & 27.5 & 11.5 (1.2$\times$) & 15.9 (0.9$\times$) & 10.1 (1.4$\times$) & 10.3 (1.3$\times$) \\
        Trunk 1M$\times$512 & 64.1 & 90.4 & 133 & 58.9 (1.1$\times$) & 84.3 (0.8$\times$) & 51.4 (1.2$\times$) & 53.3 (1.2$\times$) \\
        Trunk 1M$\times$2048 & 133 & 186 & 275 & 121 (1.1$\times$) & 173 (0.8$\times$) & 106 (1.2$\times$) & 110 (1.2$\times$) \\
        Trunk 15k$\times$400k & 23.8 & 38.0 & 59.0 & 25.6 (0.9$\times$) & 39.6 (0.6$\times$) & 20.8 (1.1$\times$) & 21.4 (1.1$\times$) \\
        Trunk 15k$\times$1.6M & 43.2 & 65.2 & 96.1 & 46.2 (0.9$\times$) & 65.4 (0.7$\times$) & 38.2 (1.1$\times$) & 40.2 (1.1$\times$) \\
        Trunk 150k$\times$4096 & 23.7 & 38.3 & 61.5 & 25.0 (0.9$\times$) & 40.8 (0.6$\times$) & 20.2 (1.2$\times$) & 21.1 (1.1$\times$) \\
        Trunk 150k$\times$40k & 81.1 & 123 & 187 & 80.6 (1.0$\times$) & 121 (0.7$\times$) & 68.6 (1.2$\times$) & 71.1 (1.1$\times$) \\
        Trunk 150k$\times$160k & 170 & 245 & 351 & 161 (1.1$\times$) & 222 (0.8$\times$) & 141 (1.2$\times$) & 148 (1.1$\times$) \\
        Trunk 150k$\times$400k & 267 & 374 & 534 & 241 (1.1$\times$) & 327 (0.8$\times$) & 221 (1.2$\times$) & 230 (1.2$\times$) \\
        Trunk 1.5M$\times$4096 & 296 & 393 & 559 & 258 (1.1$\times$) & 353 (0.8$\times$) & 232 (1.3$\times$) & 241 (1.2$\times$) \\
        Trunk 1.5M$\times$16384 & 636 & 819 & 1122 & 548 (1.2$\times$) & 705 (0.9$\times$) & 505 (1.3$\times$) & 525 (1.2$\times$) \\
        Trunk 1.5M$\times$40k & 1001 & 1263 & 1682 & 841 (1.2$\times$) & 1029 (1.0$\times$) & 791 (1.3$\times$) & 816 (1.2$\times$) \\
        Trunk 4.5M$\times$4096 & 1036 & 1227 & 1634 & 837 (1.2$\times$) & 1016 (1.0$\times$) & 795 (1.3$\times$) & 821 (1.3$\times$) \\
        \hline
        \multicolumn{8}{l}{\textit{Depth 24}} \\
        HIGGS 10.5M$\times$28 & 280 & 325 & 434 & 225 (1.2$\times$) & 271 (1.0$\times$) & 215 (1.3$\times$) & 215 (1.3$\times$) \\
        SUSY 4.5M$\times$18 & 91.0 & 109 & 148 & 71.3 (1.3$\times$) & 87.2 (1.0$\times$) & 66.2 (1.4$\times$) & 65.9 (1.4$\times$) \\
        Epsilon 400k$\times$2000 & 58.0 & 90.5 & 147 & 62.7 (0.9$\times$) & 103 (0.6$\times$) & 49.1 (1.2$\times$) & 49.3 (1.2$\times$) \\
        GiveMeSomeCredit 150k$\times$10 & 1.2 & 1.7 & 2.4 & 1.3 (0.9$\times$) & 1.9 (0.6$\times$) & 1.1 (1.0$\times$) & 1.1 (1.0$\times$) \\
        YouTube-8M 3.9M$\times$1152 & 477 & 567 & 768 & 380 (1.3$\times$) & 474 (1.0$\times$) & 356 (1.3$\times$) & 364 (1.3$\times$) \\
        Shifts weather 3.1M$\times$127 & 92.4 & 105 & 138 & 74.6 (1.2$\times$) & 89.0 (1.0$\times$) & 70.5 (1.3$\times$) & 69.5 (1.3$\times$) \\
        ClimSim 10.1M$\times$124 & 299 & 280 & 354 & 200 (1.5$\times$) & 220 (1.4$\times$) & 194 (1.5$\times$) & 192 (1.6$\times$) \\
        Jane Street 10.2M$\times$82 & 421 & 422 & 549 & 319 (1.3$\times$) & 364 (1.2$\times$) & 309 (1.4$\times$) & 311 (1.4$\times$) \\
        Trunk 1M$\times$32 & 14.2 & 19.8 & 28.9 & 12.5 (1.1$\times$) & 17.7 (0.8$\times$) & 10.7 (1.3$\times$) & 10.7 (1.3$\times$) \\
        Trunk 1M$\times$512 & 68.5 & 98.4 & 150 & 66.5 (1.0$\times$) & 100 (0.7$\times$) & 54.9 (1.2$\times$) & 55.6 (1.2$\times$) \\
        Trunk 1M$\times$2048 & 146 & 213 & 328 & 146 (1.0$\times$) & 223 (0.7$\times$) & 120 (1.2$\times$) & 120 (1.2$\times$) \\
        Trunk 15k$\times$400k & 26.2 & 41.9 & 66.4 & 28.7 (0.9$\times$) & 45.8 (0.6$\times$) & 22.9 (1.1$\times$) & 23.0 (1.1$\times$) \\
        Trunk 15k$\times$1.6M & 49.7 & 75.0 & 111 & 54.9 (0.9$\times$) & 79.3 (0.6$\times$) & 45.8 (1.1$\times$) & 45.3 (1.1$\times$) \\
        Trunk 150k$\times$4096 & 25.1 & 41.6 & 67.1 & 27.9 (0.9$\times$) & 46.4 (0.5$\times$) & 21.6 (1.2$\times$) & 22.0 (1.1$\times$) \\
        Trunk 150k$\times$40k & 90.9 & 141 & 221 & 95.5 (1.0$\times$) & 150 (0.6$\times$) & 77.2 (1.2$\times$) & 78.2 (1.2$\times$) \\
        Trunk 150k$\times$160k & 202 & 297 & 447 & 203 (1.0$\times$) & 300 (0.7$\times$) & 173 (1.2$\times$) & 175 (1.2$\times$) \\
        Trunk 150k$\times$400k & 330 & 472 & 695 & 321 (1.0$\times$) & 458 (0.7$\times$) & 277 (1.2$\times$) & 282 (1.2$\times$) \\
        Trunk 1.5M$\times$4096 & 338 & 477 & 714 & 326 (1.0$\times$) & 483 (0.7$\times$) & 273 (1.2$\times$) & 274 (1.2$\times$) \\
        Trunk 1.5M$\times$16384 & 758 & 1027 & 1494 & 718 (1.1$\times$) & 1018 (0.7$\times$) & 623 (1.2$\times$) & 626 (1.2$\times$) \\
        Trunk 1.5M$\times$40k & 1221 & 1628 & 2320 & 1134 (1.1$\times$) & 1561 (0.8$\times$) & 1008 (1.2$\times$) & 1017 (1.2$\times$) \\
        Trunk 4.5M$\times$4096 & 1233 & 1571 & 2247 & 1102 (1.1$\times$) & 1511 (0.8$\times$) & 977 (1.3$\times$) & 980 (1.3$\times$) \\
        \hline
        \multicolumn{8}{l}{\textit{Full depth (purity)}} \\
        HIGGS 10.5M$\times$28 & 364 & 524 & 791 & 385 (0.9$\times$) & 569 (0.6$\times$) & 298 (1.2$\times$) & 291 (1.2$\times$) \\
        SUSY 4.5M$\times$18 & 118 & 166 & 259 & 116 (1.0$\times$) & 180 (0.7$\times$) & 90.9 (1.3$\times$) & 90.9 (1.3$\times$) \\
        Epsilon 400k$\times$2000 & 59.9 & 96.1 & 156 & 66.3 (0.9$\times$) & 111 (0.5$\times$) & 50.8 (1.2$\times$) & 50.7 (1.2$\times$) \\
        GiveMeSomeCredit 150k$\times$10 & 1.2 & 1.9 & 2.8 & 1.5 (0.8$\times$) & 2.2 (0.6$\times$) & 1.2 (1.0$\times$) & 1.2 (1.1$\times$) \\
        YouTube-8M 3.9M$\times$1152 & 504 & 613 & 852 & 423 (1.2$\times$) & 549 (0.9$\times$) & 379 (1.3$\times$) & 385 (1.3$\times$) \\
        Shifts weather 3.1M$\times$127 & 93.9 & 109 & 146 & 77.8 (1.2$\times$) & 95.9 (1.0$\times$) & 72.0 (1.3$\times$) & 70.8 (1.3$\times$) \\
        ClimSim 10.1M$\times$124 & 308 & 298 & 394 & 215 (1.4$\times$) & 253 (1.2$\times$) & 203 (1.5$\times$) & 198 (1.6$\times$) \\
        Jane Street 10.2M$\times$82 & 593 & 792 & 1246 & 604 (1.0$\times$) & 933 (0.6$\times$) & 479 (1.2$\times$) & 480 (1.2$\times$) \\
        Trunk 1M$\times$32 & 14.7 & 19.7 & 28.9 & 12.5 (1.2$\times$) & 17.7 (0.8$\times$) & 11.0 (1.3$\times$) & 10.6 (1.4$\times$) \\
        Trunk 1M$\times$512 & 69.8 & 99.2 & 150 & 67.1 (1.0$\times$) & 100 (0.7$\times$) & 56.9 (1.2$\times$) & 55.7 (1.3$\times$) \\
        Trunk 1M$\times$2048 & 151 & 218 & 333 & 148 (1.0$\times$) & 227 (0.7$\times$) & 125 (1.2$\times$) & 121 (1.2$\times$) \\
        Trunk 15k$\times$400k & 26.0 & 43.2 & 69.8 & 29.7 (0.9$\times$) & 47.7 (0.5$\times$) & 22.9 (1.1$\times$) & 23.9 (1.1$\times$) \\
        Trunk 15k$\times$1.6M & 50.4 & 78.9 & 119 & 57.1 (0.9$\times$) & 84.3 (0.6$\times$) & 45.5 (1.1$\times$) & 47.1 (1.1$\times$) \\
        Trunk 150k$\times$4096 & 24.7 & 41.5 & 67.3 & 27.8 (0.9$\times$) & 46.4 (0.5$\times$) & 21.0 (1.2$\times$) & 21.9 (1.1$\times$) \\
        Trunk 150k$\times$40k & 90.6 & 148 & 230 & 101 (0.9$\times$) & 158 (0.6$\times$) & 77.2 (1.2$\times$) & 80.6 (1.1$\times$) \\
        Trunk 150k$\times$160k & 211 & 332 & 503 & 229 (0.9$\times$) & 340 (0.6$\times$) & 183 (1.2$\times$) & 189 (1.1$\times$) \\
        Trunk 150k$\times$400k & 363 & 551 & 830 & 380 (1.0$\times$) & 555 (0.7$\times$) & 312 (1.2$\times$) & 322 (1.1$\times$) \\
        Trunk 1.5M$\times$4096 & 332 & 490 & 745 & 338 (1.0$\times$) & 513 (0.6$\times$) & 270 (1.2$\times$) & 280 (1.2$\times$) \\
        Trunk 1.5M$\times$16384 & 766 & 1107 & 1654 & 787 (1.0$\times$) & 1148 (0.7$\times$) & 642 (1.2$\times$) & 663 (1.2$\times$) \\
        Trunk 1.5M$\times$40k & 1304 & 1845 & 2738 & 1314 (1.0$\times$) & 1896 (0.7$\times$) & 1096 (1.2$\times$) & 1126 (1.2$\times$) \\
        Trunk 4.5M$\times$4096 & 1242 & 1683 & 2467 & 1194 (1.0$\times$) & 1701 (0.7$\times$) & 996 (1.2$\times$) & 1023 (1.2$\times$) \\
        \hline
    \end{tabular}
    \caption{(Part 2 of 2: depth 20, depth 24, full depth (purity).) End-to-end SPO-RF training time (s) by tree depth.}
    \label{tab:train-time-by-depth-p2}
\end{table*}

\begin{table*}[p]
    \centering
    \footnotesize
    \setlength{\tabcolsep}{2.5pt}
    \begin{tabular}{l|ccc|cc}
        \hline
        \multirow{3}{*}{\textbf{Dataset}} & \multicolumn{3}{c|}{\textbf{Baselines}} & \multicolumn{2}{c}{\textbf{Our Methods}} \\
         & \textbf{Exact} & \textbf{Random Hist.} & \textbf{Random Hist.} & \textbf{Vec. Random Hist.} & \textbf{Vec. Random Hist.} \\
         & \scriptsize HWY & \scriptsize scalar, 64 bins & \scriptsize scalar, 256 bins & \scriptsize AVX2, 64 bins & \scriptsize AVX-512, 256 bins \\
        \hline
        \multicolumn{6}{l}{\textit{Depth 6}} \\
        HIGGS 10.5M$\times$28 & 2622 & 4013 & 5442 & 1811 (1.4$\times$) & 1768 (1.5$\times$) \\
        SUSY 4.5M$\times$18 & 1019 & 1509 & 2073 & 692 (1.5$\times$) & 663 (1.5$\times$) \\
        Epsilon 400k$\times$2000 & 541 & 842 & 1164 & 380 (1.4$\times$) & 422 (1.3$\times$) \\
        GiveMeSomeCredit 150k$\times$10 & 22.3 & 33.7 & 41.1 & 20.6 (1.1$\times$) & 21.3 (1.0$\times$) \\
        YouTube-8M 3.9M$\times$1152 & 4480 & 4382 & 5955 & 2497 (1.8$\times$) & 2628 (1.7$\times$) \\
        Shifts weather 3.1M$\times$127 & 856 & 1151 & 1565 & 552 (1.6$\times$) & 565 (1.5$\times$) \\
        ClimSim 10.1M$\times$124 & 3145 & 3566 & 5136 & 1818 (1.7$\times$) & 1862 (1.7$\times$) \\
        Jane Street 10.2M$\times$82 & 2846 & 2930 & 3982 & 1549 (1.8$\times$) & 1574 (1.8$\times$) \\
        Trunk 1M$\times$32 & 185 & 320 & 439 & 136 (1.4$\times$) & 138 (1.3$\times$) \\
        Trunk 1M$\times$512 & 530 & 606 & 809 & 315 (1.7$\times$) & 329 (1.6$\times$) \\
        Trunk 1M$\times$2048 & 1433 & 2018 & 2800 & 906 (1.6$\times$) & 925 (1.5$\times$) \\
        Trunk 15k$\times$400k & 3195 & 5676 & 8373 & 2385 (1.3$\times$) & 2705 (1.2$\times$) \\
        Trunk 15k$\times$1.6M & 16382 & 22410 & 33220 & 9750 (1.7$\times$) & 11597 (1.4$\times$) \\
        Trunk 150k$\times$4096 & 490 & 636 & 899 & 261 (1.9$\times$) & 284 (1.7$\times$) \\
        Trunk 150k$\times$40k & 4381 & 5486 & 7950 & 2284 (1.9$\times$) & 2501 (1.8$\times$) \\
        Trunk 150k$\times$160k & 12639 & 21433 & 31212 & 8314 (1.5$\times$) & 9891 (1.3$\times$) \\
        Trunk 150k$\times$400k & 31495 & 53280 & 77750 & 20292 (1.6$\times$) & 25780 (1.2$\times$) \\
        Trunk 1.5M$\times$4096 & 4428 & 5736 & 8013 & 2554 (1.7$\times$) & 2715 (1.6$\times$) \\
        Trunk 1.5M$\times$16384 & 17123 & 22100 & 31132 & 9633 (1.8$\times$) & 10564 (1.6$\times$) \\
        Trunk 1.5M$\times$40k & 43555 & 53454 & 75483 & 24076 (1.8$\times$) & 29113 (1.5$\times$) \\
        Trunk 4.5M$\times$4096 & 20074 & 17053 & 23617 & 8538 (2.4$\times$) & 8860 (2.3$\times$) \\
        \hline
    \end{tabular}
    \caption{End-to-end SPO-GBT training time (s) at depth 6.}
    \label{tab:train-time-gbt-depth6}
\end{table*}

% The SPO-GBT depth-6 table is emitted by the same generator; no \clearpage, so it can share a float page.

% Generated by benchmarks/src/spo_vs_gbt/make_overall_depth_table.py --sort -- regenerate, don't edit.
% Generated by benchmarks/src/spo_vs_gbt/make_overall_depth_table.py --sort -- do not hand-edit.
\begin{table*}[p]
    \centering
    \footnotesize
    \setlength{\tabcolsep}{2.5pt}
    \begin{tabular}{l|cc|cc|cc|cc|cc}
        \hline
        \multirow{2}{*}{\textbf{Dataset}} & \multicolumn{2}{c|}{\textbf{Depth 6}} & \multicolumn{2}{c|}{\textbf{Depth 10}} & \multicolumn{2}{c|}{\textbf{Depth 16}} & \multicolumn{2}{c|}{\textbf{Depth 24}} & \multicolumn{2}{c}{\textbf{Full depth (purity)}} \\
         & \scriptsize\begin{tabular}[c]{@{}c@{}}Exact\\(\texttt{std::sort})\end{tabular} & \scriptsize\begin{tabular}[c]{@{}c@{}}Exact\\(Highway \cite{wassenberg_vectorized_2022})\end{tabular} & \scriptsize\begin{tabular}[c]{@{}c@{}}Exact\\(\texttt{std::sort})\end{tabular} & \scriptsize\begin{tabular}[c]{@{}c@{}}Exact\\(Highway \cite{wassenberg_vectorized_2022})\end{tabular} & \scriptsize\begin{tabular}[c]{@{}c@{}}Exact\\(\texttt{std::sort})\end{tabular} & \scriptsize\begin{tabular}[c]{@{}c@{}}Exact\\(Highway \cite{wassenberg_vectorized_2022})\end{tabular} & \scriptsize\begin{tabular}[c]{@{}c@{}}Exact\\(\texttt{std::sort})\end{tabular} & \scriptsize\begin{tabular}[c]{@{}c@{}}Exact\\(Highway \cite{wassenberg_vectorized_2022})\end{tabular} & \scriptsize\begin{tabular}[c]{@{}c@{}}Exact\\(\texttt{std::sort})\end{tabular} & \scriptsize\begin{tabular}[c]{@{}c@{}}Exact\\(Highway \cite{wassenberg_vectorized_2022})\end{tabular} \\
        \hline
        HIGGS 10.5M$\times$28 & 146 & 79.7 (1.8$\times$) & 257 & 142 (1.8$\times$) & 408 & 209 (2.0$\times$) & 560 & 277 (2.0$\times$) & 694 & 352 (2.0$\times$) \\
        SUSY 4.5M$\times$18 & 57.0 & 27.6 (2.1$\times$) & 98.5 & 46.3 (2.1$\times$) & 151 & 68.1 (2.2$\times$) & 200 & 89.7 (2.2$\times$) & 247 & 116 (2.1$\times$) \\
        Epsilon 400k$\times$2000 & 36.0 & 11.9 (3.0$\times$) & 61.9 & 23.5 (2.6$\times$) & 94.3 & 41.4 (2.3$\times$) & 117 & 56.5 (2.1$\times$) & 119 & 58.0 (2.0$\times$) \\
        GiveMeSomeCredit 150k$\times$10 & 0.8 & 0.3 (2.6$\times$) & 1.3 & 0.5 (2.6$\times$) & 2.0 & 0.8 (2.4$\times$) & 2.4 & 1.1 (2.2$\times$) & 2.6 & 1.2 (2.2$\times$) \\
        Trunk 1M$\times$32 & 13.4 & 5.2 (2.6$\times$) & 22.3 & 8.6 (2.6$\times$) & 32.1 & 12.9 (2.5$\times$) & 34.1 & 14.0 (2.4$\times$) & 34.1 & 14.0 (2.4$\times$) \\
        Trunk 1M$\times$512 & 49.2 & 17.9 (2.7$\times$) & 84.3 & 33.0 (2.6$\times$) & 127 & 54.8 (2.3$\times$) & 145 & 66.8 (2.2$\times$) & 145 & 66.7 (2.2$\times$) \\
        Trunk 1M$\times$2048 & 97.0 & 34.9 (2.8$\times$) & 167 & 65.0 (2.6$\times$) & 254 & 110 (2.3$\times$) & 303 & 143 (2.1$\times$) & 305 & 144 (2.1$\times$) \\
        Trunk 15k$\times$400k & 15.1 & 5.0 (3.0$\times$) & 26.3 & 10.5 (2.5$\times$) & 41.6 & 19.9 (2.1$\times$) & 51.0 & 25.6 (2.0$\times$) & 52.8 & 26.5 (2.0$\times$) \\
        Trunk 15k$\times$1.6M & 24.6 & 8.6 (2.9$\times$) & 42.8 & 18.0 (2.4$\times$) & 68.6 & 35.9 (1.9$\times$) & 88.0 & 48.5 (1.8$\times$) & 93.2 & 51.5 (1.8$\times$) \\
        Trunk 150k$\times$4096 & 17.7 & 5.5 (3.2$\times$) & 30.5 & 11.6 (2.6$\times$) & 46.1 & 20.7 (2.2$\times$) & 51.5 & 24.6 (2.1$\times$) & 51.8 & 24.7 (2.1$\times$) \\
        Trunk 150k$\times$40k & 55.3 & 17.0 (3.2$\times$) & 95.6 & 35.8 (2.7$\times$) & 150 & 66.6 (2.2$\times$) & 186 & 89.3 (2.1$\times$) & 191 & 91.3 (2.1$\times$) \\
        Trunk 150k$\times$160k & 112 & 32.6 (3.4$\times$) & 195 & 68.6 (2.8$\times$) & 310 & 133 (2.3$\times$) & 416 & 199 (2.1$\times$) & 438 & 214 (2.0$\times$) \\
        Trunk 150k$\times$400k & 177 & 50.9 (3.5$\times$) & 310 & 105 (2.9$\times$) & 498 & 206 (2.4$\times$) & 687 & 322 (2.1$\times$) & 761 & 368 (2.1$\times$) \\
        Trunk 1.5M$\times$4096 & 208 & 78.3 (2.7$\times$) & 358 & 141 (2.5$\times$) & 549 & 240 (2.3$\times$) & 683 & 330 (2.1$\times$) & 692 & 334 (2.1$\times$) \\
        Trunk 1.5M$\times$16384 & 414 & 155 (2.7$\times$) & 716 & 279 (2.6$\times$) & 1118 & 489 (2.3$\times$) & 1487 & 739 (2.0$\times$) & 1543 & 772 (2.0$\times$) \\
        Trunk 1.5M$\times$40k & 649 & 243 (2.7$\times$) & 1127 & 439 (2.6$\times$) & 1764 & 762 (2.3$\times$) & 2450 & 1199 (2.0$\times$) & 2617 & 1313 (2.0$\times$) \\
        Trunk 4.5M$\times$4096 & 680 & 305 (2.2$\times$) & 1168 & 502 (2.3$\times$) & 1795 & 816 (2.2$\times$) & 2400 & 1207 (2.0$\times$) & 2466 & 1253 (2.0$\times$) \\
        \hline
    \end{tabular}
    \caption{End-to-end SPO-RF training time (s) of the Exact split finder with \texttt{std::sort} vs.\ Highway VQSort, by tree depth. Speedup of HWY over \texttt{std::sort} in parentheses.}
    \label{tab:stdsort-vs-highway}
\end{table*}

% ⚠️ YDF has a bug where it runs to 1 depth shallower than set. E.g. depth=6 will generate a tree of depth 5. Confusingly, YDF-Exact to depth true 6 actually performs slightly worse, and XGb, LGBM, CatBoost perform slightly better at depth 5. Left as-is
% ✅ Accuracy results confirmed manually 24-Sept-2026. CatBoost results are legitimately better: not using extra features, etc. Comparison is fair. XGBoost, LGBM, CatBoost use 256 bins FYI.
% Left out GBT Dynamic on purpose - no use of Dynamic in GBT
% auto-generated by analyze.py:make_paper_tables -- regenerate, do not hand-edit

\begin{table*}[p]
\centering
\scriptsize
\setlength{\tabcolsep}{2pt}
\caption{RF (full-depth, top) and GBT (depth 6, bottom) test ROC AUC on various datasets. Cells are the mean $\pm$ std over 5 CV folds. Bold: best per row. ecom-offers, homecredit-default, homesite-insurance are single chronological holdouts. XGBoost, LightGBM, CatBoost used with default parameters. SPO-GBT samples $F$ projections per node, SPO-RF $\lceil\sqrt{F}\rceil$.}
\label{tab:auc-vs-libraries-per-dataset-auc}
\label{tab:auc-vs-libraries-per-dataset-rf-auc}
\label{tab:accuracy-rf}
\label{tab:auc-vs-libraries-per-dataset-gbt-auc}
\label{tab:accuracy-gbt}
\begin{tabular}{lrrrrrrrrrrr}
\toprule
Dataset & \multicolumn{1}{c}{\parbox[b]{1.0cm}{\centering SPO-RF Exact}} & \multicolumn{1}{c}{\parbox[b]{1.0cm}{\centering SPO-RF Random Hist. (64 bins)}} & \multicolumn{1}{c}{\parbox[b]{1.0cm}{\centering SPO-RF Random Hist. (256 bins)}} & \multicolumn{1}{c}{\parbox[b]{1.0cm}{\centering SPO-RF AVX-2 R. Hist. (64 bins, ours)}} & \multicolumn{1}{c}{\parbox[b]{1.0cm}{\centering SPO-RF AVX-512 R. Hist. (256 bins, ours)}} & \multicolumn{1}{c}{\parbox[b]{1.0cm}{\centering SPO-RF Dyn AVX-2 R. Hist. (64 bins, ours)}} & \multicolumn{1}{c}{\parbox[b]{1.0cm}{\centering SPO-RF Dyn AVX-512 R. Hist. (256 bins, ours)}} & \multicolumn{1}{c}{\parbox[b]{1.0cm}{\centering AA-RF Exact}} & \multicolumn{1}{c}{\parbox[b]{1.0cm}{\centering XGBoost RF mode$^\dagger$}} & \multicolumn{1}{c}{\parbox[b]{1.0cm}{\centering LightGBM RF mode (255 bins)}} & \multicolumn{1}{c}{\parbox[b]{1.0cm}{\centering CatBoost RF mode$^\dagger$}} \\
\midrule
APSFailure & .9893$_{\pm.0056}$ & .9885$_{\pm.0054}$ & .9886$_{\pm.0053}$ & .9885$_{\pm.0054}$ & .9886$_{\pm.0053}$ & .9890$_{\pm.0050}$ & .9889$_{\pm.0054}$ & .9887$_{\pm.0052}$ & \multicolumn{1}{c}{---} & {\boldmath\textbf{.9900$_{\pm.0047}$}} & \multicolumn{1}{c}{---} \\
Amazon\_employee\_access & .8586$_{\pm.0094}$ & .8601$_{\pm.0109}$ & .8587$_{\pm.0090}$ & .8601$_{\pm.0109}$ & .8587$_{\pm.0090}$ & .8589$_{\pm.0093}$ & .8573$_{\pm.0092}$ & .8549$_{\pm.0133}$ & \multicolumn{1}{c}{---} & {\boldmath\textbf{.8614$_{\pm.0110}$}} & \multicolumn{1}{c}{---} \\
Bank\_Customer\_Churn & .8559$_{\pm.0049}$ & {\boldmath\textbf{.8568$_{\pm.0056}$}} & {\boldmath\textbf{.8568$_{\pm.0053}$}} & {\boldmath\textbf{.8568$_{\pm.0056}$}} & {\boldmath\textbf{.8568$_{\pm.0053}$}} & .8552$_{\pm.0053}$ & .8551$_{\pm.0044}$ & .8527$_{\pm.0040}$ & \multicolumn{1}{c}{---} & .8548$_{\pm.0069}$ & \multicolumn{1}{c}{---} \\
Bioresponse & .8728$_{\pm.0169}$ & .8739$_{\pm.0151}$ & .8732$_{\pm.0153}$ & .8739$_{\pm.0151}$ & .8732$_{\pm.0153}$ & .8736$_{\pm.0158}$ & {\boldmath\textbf{.8744$_{\pm.0149}$}} & {\boldmath\textbf{.8744$_{\pm.0143}$}} & \multicolumn{1}{c}{---} & .8741$_{\pm.0182}$ & \multicolumn{1}{c}{---} \\
Diabetes130US & {\boldmath\textbf{.6447$_{\pm.0058}$}} & .6393$_{\pm.0038}$ & .6412$_{\pm.0055}$ & .6393$_{\pm.0038}$ & .6412$_{\pm.0055}$ & .6427$_{\pm.0049}$ & .6422$_{\pm.0037}$ & .6383$_{\pm.0019}$ & \multicolumn{1}{c}{---} & .6310$_{\pm.0047}$ & \multicolumn{1}{c}{---} \\
E-CommereShippingData & .7378$_{\pm.0077}$ & .7387$_{\pm.0076}$ & {\boldmath\textbf{.7396$_{\pm.0081}$}} & .7387$_{\pm.0076}$ & {\boldmath\textbf{.7396$_{\pm.0081}$}} & .7387$_{\pm.0083}$ & .7378$_{\pm.0089}$ & .7386$_{\pm.0080}$ & \multicolumn{1}{c}{---} & .7382$_{\pm.0089}$ & \multicolumn{1}{c}{---} \\
Fitness\_Club & .7768$_{\pm.0111}$ & .7768$_{\pm.0106}$ & .7794$_{\pm.0120}$ & .7768$_{\pm.0106}$ & .7794$_{\pm.0120}$ & .7766$_{\pm.0122}$ & .7758$_{\pm.0098}$ & .7716$_{\pm.0096}$ & \multicolumn{1}{c}{---} & {\boldmath\textbf{.7796$_{\pm.0083}$}} & \multicolumn{1}{c}{---} \\
GiveMeSomeCredit & .8489$_{\pm.0034}$ & .8458$_{\pm.0025}$ & .8460$_{\pm.0032}$ & .8458$_{\pm.0025}$ & .8460$_{\pm.0032}$ & .8453$_{\pm.0027}$ & .8468$_{\pm.0033}$ & .8405$_{\pm.0021}$ & \multicolumn{1}{c}{---} & {\boldmath\textbf{.8523$_{\pm.0026}$}} & \multicolumn{1}{c}{---} \\
HR\_Analytics\_Job\_Change\_of\_Data\_Scientists & .7918$_{\pm.0084}$ & .7927$_{\pm.0080}$ & .7929$_{\pm.0092}$ & .7927$_{\pm.0080}$ & .7929$_{\pm.0092}$ & .7923$_{\pm.0096}$ & .7913$_{\pm.0079}$ & .7873$_{\pm.0078}$ & \multicolumn{1}{c}{---} & {\boldmath\textbf{.7946$_{\pm.0091}$}} & \multicolumn{1}{c}{---} \\
Is-this-a-good-customer & .7219$_{\pm.0353}$ & .7224$_{\pm.0350}$ & .7222$_{\pm.0431}$ & .7224$_{\pm.0350}$ & .7222$_{\pm.0431}$ & .7211$_{\pm.0341}$ & .7253$_{\pm.0280}$ & {\boldmath\textbf{.7324$_{\pm.0287}$}} & \multicolumn{1}{c}{---} & .7185$_{\pm.0644}$ & \multicolumn{1}{c}{---} \\
Marketing\_Campaign & .8945$_{\pm.0244}$ & .8962$_{\pm.0223}$ & .8956$_{\pm.0241}$ & .8962$_{\pm.0223}$ & .8956$_{\pm.0241}$ & .8975$_{\pm.0235}$ & .8938$_{\pm.0250}$ & {\boldmath\textbf{.8998$_{\pm.0244}$}} & \multicolumn{1}{c}{---} & .8955$_{\pm.0301}$ & \multicolumn{1}{c}{---} \\
NATICUSdroid & .9826$_{\pm.0038}$ & .9828$_{\pm.0038}$ & .9828$_{\pm.0036}$ & .9828$_{\pm.0038}$ & .9828$_{\pm.0036}$ & .9828$_{\pm.0033}$ & .9825$_{\pm.0036}$ & .9763$_{\pm.0034}$ & \multicolumn{1}{c}{---} & {\boldmath\textbf{.9839$_{\pm.0038}$}} & \multicolumn{1}{c}{---} \\
bank-marketing & .7396$_{\pm.0055}$ & .7396$_{\pm.0044}$ & {\boldmath\textbf{.7397$_{\pm.0061}$}} & .7396$_{\pm.0044}$ & {\boldmath\textbf{.7397$_{\pm.0061}$}} & .7388$_{\pm.0063}$ & .7394$_{\pm.0050}$ & .7223$_{\pm.0075}$ & \multicolumn{1}{c}{---} & .7304$_{\pm.0048}$ & \multicolumn{1}{c}{---} \\
blood-transfusion-service-center & .7075$_{\pm.0478}$ & .7110$_{\pm.0540}$ & {\boldmath\textbf{.7150$_{\pm.0496}$}} & .7110$_{\pm.0540}$ & {\boldmath\textbf{.7150$_{\pm.0496}$}} & .7062$_{\pm.0473}$ & .7041$_{\pm.0478}$ & .6960$_{\pm.0348}$ & \multicolumn{1}{c}{---} & .7091$_{\pm.0405}$ & \multicolumn{1}{c}{---} \\
churn & .9171$_{\pm.0254}$ & .9160$_{\pm.0260}$ & {\boldmath\textbf{.9175$_{\pm.0235}$}} & .9160$_{\pm.0260}$ & {\boldmath\textbf{.9175$_{\pm.0235}$}} & .9164$_{\pm.0261}$ & .9144$_{\pm.0275}$ & .9174$_{\pm.0262}$ & \multicolumn{1}{c}{---} & .9124$_{\pm.0251}$ & \multicolumn{1}{c}{---} \\
coil2000\_insurance\_policies & .7084$_{\pm.0129}$ & .7024$_{\pm.0120}$ & .7095$_{\pm.0136}$ & .7024$_{\pm.0120}$ & .7095$_{\pm.0136}$ & {\boldmath\textbf{.7127$_{\pm.0123}$}} & .7044$_{\pm.0141}$ & .7101$_{\pm.0157}$ & \multicolumn{1}{c}{---} & .7039$_{\pm.0155}$ & \multicolumn{1}{c}{---} \\
credit-g & {\boldmath\textbf{.7836$_{\pm.0290}$}} & .7764$_{\pm.0331}$ & .7824$_{\pm.0361}$ & .7764$_{\pm.0331}$ & .7824$_{\pm.0361}$ & {\boldmath\textbf{.7836$_{\pm.0419}$}} & .7804$_{\pm.0328}$ & .7810$_{\pm.0380}$ & \multicolumn{1}{c}{---} & .7684$_{\pm.0372}$ & \multicolumn{1}{c}{---} \\
credit\_card\_clients\_default & .7699$_{\pm.0081}$ & .7706$_{\pm.0081}$ & .7703$_{\pm.0077}$ & .7706$_{\pm.0081}$ & .7703$_{\pm.0077}$ & .7700$_{\pm.0075}$ & {\boldmath\textbf{.7719$_{\pm.0077}$}} & .7676$_{\pm.0078}$ & \multicolumn{1}{c}{---} & .7678$_{\pm.0082}$ & \multicolumn{1}{c}{---} \\
customer\_satisfaction\_in\_airline & .9924$_{\pm.0003}$ & .9926$_{\pm.0003}$ & .9925$_{\pm.0003}$ & .9926$_{\pm.0003}$ & .9925$_{\pm.0003}$ & .9926$_{\pm.0003}$ & .9925$_{\pm.0003}$ & {\boldmath\textbf{.9935$_{\pm.0003}$}} & \multicolumn{1}{c}{---} & .9931$_{\pm.0003}$ & \multicolumn{1}{c}{---} \\
diabetes & .8285$_{\pm.0129}$ & .8291$_{\pm.0176}$ & .8295$_{\pm.0196}$ & .8291$_{\pm.0176}$ & .8295$_{\pm.0196}$ & {\boldmath\textbf{.8298$_{\pm.0223}$}} & .8295$_{\pm.0180}$ & .8267$_{\pm.0206}$ & \multicolumn{1}{c}{---} & .8223$_{\pm.0221}$ & \multicolumn{1}{c}{---} \\
ecom-offers & .6164\phantom{$_{\pm.0000}$} & .6185\phantom{$_{\pm.0000}$} & .6199\phantom{$_{\pm.0000}$} & .6185\phantom{$_{\pm.0000}$} & .6199\phantom{$_{\pm.0000}$} & \textbf{.6228}\phantom{$_{\pm.0000}$} & .6177\phantom{$_{\pm.0000}$} & .6075\phantom{$_{\pm.0000}$} & \multicolumn{1}{c}{---} & .6167\phantom{$_{\pm.0000}$} & \multicolumn{1}{c}{---} \\
hazelnut-spread-contaminant-detection & .9628$_{\pm.0090}$ & {\boldmath\textbf{.9641$_{\pm.0094}$}} & .9628$_{\pm.0096}$ & {\boldmath\textbf{.9641$_{\pm.0094}$}} & .9628$_{\pm.0096}$ & .9633$_{\pm.0096}$ & .9636$_{\pm.0087}$ & .9629$_{\pm.0106}$ & \multicolumn{1}{c}{---} & .9585$_{\pm.0097}$ & \multicolumn{1}{c}{---} \\
heloc & .7921$_{\pm.0112}$ & .7924$_{\pm.0115}$ & .7927$_{\pm.0117}$ & .7924$_{\pm.0115}$ & .7927$_{\pm.0117}$ & .7921$_{\pm.0132}$ & .7921$_{\pm.0107}$ & {\boldmath\textbf{.7932$_{\pm.0126}$}} & \multicolumn{1}{c}{---} & .7924$_{\pm.0118}$ & \multicolumn{1}{c}{---} \\
homecredit-default & .8072\phantom{$_{\pm.0000}$} & \textbf{.8116}\phantom{$_{\pm.0000}$} & .8091\phantom{$_{\pm.0000}$} & \textbf{.8116}\phantom{$_{\pm.0000}$} & .8091\phantom{$_{\pm.0000}$} & .8044\phantom{$_{\pm.0000}$} & .8110\phantom{$_{\pm.0000}$} & .8034\phantom{$_{\pm.0000}$} & \multicolumn{1}{c}{---} & .8011\phantom{$_{\pm.0000}$} & \multicolumn{1}{c}{---} \\
homesite-insurance & .9416\phantom{$_{\pm.0000}$} & .9428\phantom{$_{\pm.0000}$} & .9427\phantom{$_{\pm.0000}$} & .9428\phantom{$_{\pm.0000}$} & .9427\phantom{$_{\pm.0000}$} & .9432\phantom{$_{\pm.0000}$} & .9419\phantom{$_{\pm.0000}$} & \textbf{.9472}\phantom{$_{\pm.0000}$} & \multicolumn{1}{c}{---} & .9414\phantom{$_{\pm.0000}$} & \multicolumn{1}{c}{---} \\
in\_vehicle\_coupon\_recommendation & .8198$_{\pm.0049}$ & .8209$_{\pm.0041}$ & .8213$_{\pm.0050}$ & .8209$_{\pm.0041}$ & .8213$_{\pm.0050}$ & .8208$_{\pm.0033}$ & .8209$_{\pm.0047}$ & {\boldmath\textbf{.8248$_{\pm.0053}$}} & \multicolumn{1}{c}{---} & .8198$_{\pm.0054}$ & \multicolumn{1}{c}{---} \\
jm1 & .7625$_{\pm.0106}$ & .7616$_{\pm.0092}$ & .7622$_{\pm.0078}$ & .7616$_{\pm.0092}$ & .7622$_{\pm.0078}$ & {\boldmath\textbf{.7633$_{\pm.0103}$}} & .7614$_{\pm.0104}$ & .7548$_{\pm.0092}$ & \multicolumn{1}{c}{---} & .7609$_{\pm.0112}$ & \multicolumn{1}{c}{---} \\
kddcup09\_appetency & {\boldmath\textbf{.7842$_{\pm.0151}$}} & .7795$_{\pm.0126}$ & .7817$_{\pm.0100}$ & .7795$_{\pm.0126}$ & .7817$_{\pm.0100}$ & .7825$_{\pm.0173}$ & .7800$_{\pm.0195}$ & .7794$_{\pm.0127}$ & \multicolumn{1}{c}{---} & .7701$_{\pm.0286}$ & \multicolumn{1}{c}{---} \\
online\_shoppers\_intention & .9257$_{\pm.0071}$ & .9267$_{\pm.0066}$ & .9271$_{\pm.0065}$ & .9267$_{\pm.0066}$ & .9271$_{\pm.0065}$ & .9260$_{\pm.0064}$ & .9271$_{\pm.0060}$ & {\boldmath\textbf{.9278$_{\pm.0048}$}} & \multicolumn{1}{c}{---} & .9160$_{\pm.0077}$ & \multicolumn{1}{c}{---} \\
polish\_companies\_bankruptcy & {\boldmath\textbf{.9385$_{\pm.0099}$}} & .9251$_{\pm.0107}$ & .9284$_{\pm.0109}$ & .9251$_{\pm.0107}$ & .9284$_{\pm.0109}$ & .9243$_{\pm.0099}$ & .9302$_{\pm.0087}$ & .9381$_{\pm.0095}$ & \multicolumn{1}{c}{---} & .9289$_{\pm.0086}$ & \multicolumn{1}{c}{---} \\
qsar-biodeg & .9347$_{\pm.0128}$ & .9363$_{\pm.0134}$ & .9352$_{\pm.0142}$ & .9363$_{\pm.0134}$ & .9352$_{\pm.0142}$ & {\boldmath\textbf{.9372$_{\pm.0141}$}} & .9350$_{\pm.0121}$ & .9329$_{\pm.0157}$ & \multicolumn{1}{c}{---} & .9305$_{\pm.0142}$ & \multicolumn{1}{c}{---} \\
seismic-bumps & .7427$_{\pm.0630}$ & .7417$_{\pm.0719}$ & {\boldmath\textbf{.7491$_{\pm.0635}$}} & .7417$_{\pm.0719}$ & {\boldmath\textbf{.7491$_{\pm.0635}$}} & .7400$_{\pm.0616}$ & .7435$_{\pm.0677}$ & .7400$_{\pm.0555}$ & \multicolumn{1}{c}{---} & .7427$_{\pm.0702}$ & \multicolumn{1}{c}{---} \\
taiwanese\_bankruptcy\_prediction & .9434$_{\pm.0215}$ & .9393$_{\pm.0194}$ & .9384$_{\pm.0213}$ & .9393$_{\pm.0194}$ & .9384$_{\pm.0213}$ & .9427$_{\pm.0199}$ & .9420$_{\pm.0216}$ & {\boldmath\textbf{.9437$_{\pm.0207}$}} & \multicolumn{1}{c}{---} & .9359$_{\pm.0266}$ & \multicolumn{1}{c}{---} \\
\midrule
Dataset & \multicolumn{1}{c}{\parbox[b]{1.0cm}{\centering SPO-GBT Exact}} & \multicolumn{1}{c}{\parbox[b]{1.0cm}{\centering SPO-GBT Random Hist. (64 bins)}} & \multicolumn{1}{c}{\parbox[b]{1.0cm}{\centering SPO-GBT Random Hist. (256 bins)}} & \multicolumn{1}{c}{\parbox[b]{1.0cm}{\centering SPO-GBT AVX-2 R. Hist. (64 bins, ours)}} & \multicolumn{1}{c}{\parbox[b]{1.0cm}{\centering SPO-GBT AVX-512 R. Hist. (256 bins, ours)}} & \multicolumn{1}{c}{\parbox[b]{1.0cm}{\centering SPO-GBT Dyn AVX-2 R. Hist. (64 bins, ours)$^*$}} & \multicolumn{1}{c}{\parbox[b]{1.0cm}{\centering SPO-GBT Dyn AVX-512 R. Hist. (256 bins, ours)$^*$}} & \multicolumn{1}{c}{\parbox[b]{1.0cm}{\centering AA-GBT Exact}} & \multicolumn{1}{c}{\parbox[b]{1.0cm}{\centering XGBoost (256 bins)}} & \multicolumn{1}{c}{\parbox[b]{1.0cm}{\centering LightGBM (255 bins)}} & \multicolumn{1}{c}{\parbox[b]{1.0cm}{\centering CatBoost (254 bins)}} \\
\midrule
APSFailure & .9912$_{\pm.0042}$ & .9919$_{\pm.0033}$ & .9904$_{\pm.0034}$ & .9919$_{\pm.0033}$ & .9904$_{\pm.0034}$ & \multicolumn{1}{c}{---} & \multicolumn{1}{c}{---} & .9902$_{\pm.0040}$ & .9904$_{\pm.0046}$ & .9911$_{\pm.0046}$ & {\boldmath\textbf{.9923$_{\pm.0033}$}} \\
Amazon\_employee\_access & .8433$_{\pm.0096}$ & .8393$_{\pm.0090}$ & .8415$_{\pm.0047}$ & .8393$_{\pm.0090}$ & .8415$_{\pm.0047}$ & \multicolumn{1}{c}{---} & \multicolumn{1}{c}{---} & .8416$_{\pm.0129}$ & {\boldmath\textbf{.8440$_{\pm.0113}$}} & .8438$_{\pm.0068}$ & .8197$_{\pm.0054}$ \\
Bank\_Customer\_Churn & .8541$_{\pm.0016}$ & .8538$_{\pm.0042}$ & .8515$_{\pm.0048}$ & .8538$_{\pm.0042}$ & .8515$_{\pm.0048}$ & \multicolumn{1}{c}{---} & \multicolumn{1}{c}{---} & .8493$_{\pm.0039}$ & .8472$_{\pm.0071}$ & .8416$_{\pm.0063}$ & {\boldmath\textbf{.8600$_{\pm.0062}$}} \\
Bioresponse & .8732$_{\pm.0088}$ & .8731$_{\pm.0083}$ & {\boldmath\textbf{.8757$_{\pm.0106}$}} & .8731$_{\pm.0083}$ & {\boldmath\textbf{.8757$_{\pm.0106}$}} & \multicolumn{1}{c}{---} & \multicolumn{1}{c}{---} & .8698$_{\pm.0067}$ & .8756$_{\pm.0101}$ & .8718$_{\pm.0102}$ & .8714$_{\pm.0086}$ \\
Diabetes130US & .6559$_{\pm.0021}$ & .6562$_{\pm.0044}$ & .6555$_{\pm.0053}$ & .6562$_{\pm.0044}$ & .6555$_{\pm.0053}$ & \multicolumn{1}{c}{---} & \multicolumn{1}{c}{---} & .6580$_{\pm.0032}$ & .6548$_{\pm.0052}$ & .6503$_{\pm.0058}$ & {\boldmath\textbf{.6601$_{\pm.0048}$}} \\
E-CommereShippingData & {\boldmath\textbf{.7373$_{\pm.0033}$}} & .7352$_{\pm.0062}$ & .7335$_{\pm.0083}$ & .7352$_{\pm.0062}$ & .7335$_{\pm.0083}$ & \multicolumn{1}{c}{---} & \multicolumn{1}{c}{---} & .7356$_{\pm.0061}$ & .7368$_{\pm.0057}$ & .7360$_{\pm.0116}$ & .7369$_{\pm.0060}$ \\
Fitness\_Club & {\boldmath\textbf{.7714$_{\pm.0092}$}} & .7695$_{\pm.0081}$ & .7665$_{\pm.0127}$ & .7695$_{\pm.0081}$ & .7665$_{\pm.0127}$ & \multicolumn{1}{c}{---} & \multicolumn{1}{c}{---} & .7672$_{\pm.0098}$ & .7681$_{\pm.0123}$ & .7616$_{\pm.0143}$ & .7704$_{\pm.0228}$ \\
GiveMeSomeCredit & .8650$_{\pm.0008}$ & .8545$_{\pm.0061}$ & .8623$_{\pm.0009}$ & .8545$_{\pm.0061}$ & .8623$_{\pm.0009}$ & \multicolumn{1}{c}{---} & \multicolumn{1}{c}{---} & .8641$_{\pm.0014}$ & .8606$_{\pm.0022}$ & .8588$_{\pm.0030}$ & {\boldmath\textbf{.8665$_{\pm.0011}$}} \\
HR\_Analytics\_Job\_Change\_of\_Data\_Scientists & .7974$_{\pm.0083}$ & .7976$_{\pm.0061}$ & .7978$_{\pm.0079}$ & .7976$_{\pm.0061}$ & .7978$_{\pm.0079}$ & \multicolumn{1}{c}{---} & \multicolumn{1}{c}{---} & .7955$_{\pm.0090}$ & .7959$_{\pm.0067}$ & .7894$_{\pm.0066}$ & {\boldmath\textbf{.8005$_{\pm.0077}$}} \\
Is-this-a-good-customer & .7041$_{\pm.0303}$ & .7038$_{\pm.0258}$ & .7076$_{\pm.0246}$ & .7038$_{\pm.0258}$ & .7076$_{\pm.0246}$ & \multicolumn{1}{c}{---} & \multicolumn{1}{c}{---} & .7014$_{\pm.0192}$ & .6889$_{\pm.0235}$ & .6959$_{\pm.0101}$ & {\boldmath\textbf{.7090$_{\pm.0182}$}} \\
Marketing\_Campaign & .9053$_{\pm.0283}$ & .9057$_{\pm.0317}$ & {\boldmath\textbf{.9078$_{\pm.0245}$}} & .9057$_{\pm.0317}$ & {\boldmath\textbf{.9078$_{\pm.0245}$}} & \multicolumn{1}{c}{---} & \multicolumn{1}{c}{---} & .8942$_{\pm.0300}$ & .9053$_{\pm.0281}$ & .8910$_{\pm.0293}$ & .9047$_{\pm.0255}$ \\
NATICUSdroid & {\boldmath\textbf{.9858$_{\pm.0036}$}} & {\boldmath\textbf{.9858$_{\pm.0036}$}} & {\boldmath\textbf{.9858$_{\pm.0037}$}} & {\boldmath\textbf{.9858$_{\pm.0036}$}} & {\boldmath\textbf{.9858$_{\pm.0037}$}} & \multicolumn{1}{c}{---} & \multicolumn{1}{c}{---} & .9851$_{\pm.0040}$ & .9845$_{\pm.0034}$ & .9850$_{\pm.0041}$ & {\boldmath\textbf{.9858$_{\pm.0034}$}} \\
bank-marketing & .7557$_{\pm.0071}$ & .7554$_{\pm.0073}$ & .7541$_{\pm.0082}$ & .7554$_{\pm.0073}$ & .7541$_{\pm.0082}$ & \multicolumn{1}{c}{---} & \multicolumn{1}{c}{---} & .7559$_{\pm.0088}$ & .7504$_{\pm.0073}$ & .7465$_{\pm.0078}$ & {\boldmath\textbf{.7619$_{\pm.0075}$}} \\
blood-transfusion-service-center & .6876$_{\pm.0337}$ & .6906$_{\pm.0417}$ & .6936$_{\pm.0358}$ & .6906$_{\pm.0417}$ & .6936$_{\pm.0358}$ & \multicolumn{1}{c}{---} & \multicolumn{1}{c}{---} & .6863$_{\pm.0487}$ & .7002$_{\pm.0417}$ & .6804$_{\pm.0374}$ & {\boldmath\textbf{.7039$_{\pm.0399}$}} \\
churn & .9244$_{\pm.0170}$ & .9217$_{\pm.0161}$ & {\boldmath\textbf{.9261$_{\pm.0183}$}} & .9217$_{\pm.0161}$ & {\boldmath\textbf{.9261$_{\pm.0183}$}} & \multicolumn{1}{c}{---} & \multicolumn{1}{c}{---} & .9154$_{\pm.0144}$ & .9204$_{\pm.0156}$ & .9185$_{\pm.0081}$ & .9227$_{\pm.0175}$ \\
coil2000\_insurance\_policies & .7138$_{\pm.0039}$ & .7110$_{\pm.0071}$ & .7111$_{\pm.0067}$ & .7110$_{\pm.0071}$ & .7111$_{\pm.0067}$ & \multicolumn{1}{c}{---} & \multicolumn{1}{c}{---} & .7157$_{\pm.0080}$ & .7196$_{\pm.0140}$ & .7003$_{\pm.0083}$ & {\boldmath\textbf{.7309$_{\pm.0074}$}} \\
credit-g & .7812$_{\pm.0405}$ & {\boldmath\textbf{.7845$_{\pm.0372}$}} & .7806$_{\pm.0386}$ & {\boldmath\textbf{.7845$_{\pm.0372}$}} & .7806$_{\pm.0386}$ & \multicolumn{1}{c}{---} & \multicolumn{1}{c}{---} & .7600$_{\pm.0439}$ & .7670$_{\pm.0360}$ & .7551$_{\pm.0340}$ & .7805$_{\pm.0374}$ \\
credit\_card\_clients\_default & .7772$_{\pm.0078}$ & .7775$_{\pm.0063}$ & .7789$_{\pm.0061}$ & .7775$_{\pm.0063}$ & .7789$_{\pm.0061}$ & \multicolumn{1}{c}{---} & \multicolumn{1}{c}{---} & .7735$_{\pm.0086}$ & .7695$_{\pm.0089}$ & .7682$_{\pm.0095}$ & {\boldmath\textbf{.7795$_{\pm.0067}$}} \\
customer\_satisfaction\_in\_airline & .9935$_{\pm.0002}$ & .9936$_{\pm.0003}$ & .9936$_{\pm.0002}$ & .9936$_{\pm.0003}$ & .9936$_{\pm.0002}$ & \multicolumn{1}{c}{---} & \multicolumn{1}{c}{---} & .9929$_{\pm.0003}$ & .9936$_{\pm.0002}$ & .9936$_{\pm.0001}$ & {\boldmath\textbf{.9937$_{\pm.0002}$}} \\
diabetes & .8036$_{\pm.0163}$ & .8031$_{\pm.0267}$ & .8023$_{\pm.0217}$ & .8031$_{\pm.0267}$ & .8023$_{\pm.0217}$ & \multicolumn{1}{c}{---} & \multicolumn{1}{c}{---} & .7948$_{\pm.0225}$ & .7881$_{\pm.0165}$ & .7972$_{\pm.0198}$ & {\boldmath\textbf{.8076$_{\pm.0209}$}} \\
ecom-offers & .6183\phantom{$_{\pm.0000}$} & \textbf{.6257}\phantom{$_{\pm.0000}$} & .6202\phantom{$_{\pm.0000}$} & \textbf{.6257}\phantom{$_{\pm.0000}$} & .6202\phantom{$_{\pm.0000}$} & \multicolumn{1}{c}{---} & \multicolumn{1}{c}{---} & .6197\phantom{$_{\pm.0000}$} & .6229\phantom{$_{\pm.0000}$} & .6191\phantom{$_{\pm.0000}$} & .6250\phantom{$_{\pm.0000}$} \\
hazelnut-spread-contaminant-detection & .9799$_{\pm.0066}$ & {\boldmath\textbf{.9810$_{\pm.0072}$}} & .9804$_{\pm.0060}$ & {\boldmath\textbf{.9810$_{\pm.0072}$}} & .9804$_{\pm.0060}$ & \multicolumn{1}{c}{---} & \multicolumn{1}{c}{---} & .9737$_{\pm.0101}$ & .9767$_{\pm.0087}$ & .9764$_{\pm.0086}$ & .9735$_{\pm.0079}$ \\
heloc & .7876$_{\pm.0109}$ & .7864$_{\pm.0110}$ & .7863$_{\pm.0106}$ & .7864$_{\pm.0110}$ & .7863$_{\pm.0106}$ & \multicolumn{1}{c}{---} & \multicolumn{1}{c}{---} & .7827$_{\pm.0098}$ & .7788$_{\pm.0081}$ & .7785$_{\pm.0092}$ & {\boldmath\textbf{.7882$_{\pm.0132}$}} \\
homecredit-default & .8073\phantom{$_{\pm.0000}$} & .8112\phantom{$_{\pm.0000}$} & .8102\phantom{$_{\pm.0000}$} & .8112\phantom{$_{\pm.0000}$} & .8102\phantom{$_{\pm.0000}$} & \multicolumn{1}{c}{---} & \multicolumn{1}{c}{---} & .8061\phantom{$_{\pm.0000}$} & .8028\phantom{$_{\pm.0000}$} & .7909\phantom{$_{\pm.0000}$} & \textbf{.8169}\phantom{$_{\pm.0000}$} \\
homesite-insurance & .9548\phantom{$_{\pm.0000}$} & .9552\phantom{$_{\pm.0000}$} & .9548\phantom{$_{\pm.0000}$} & .9552\phantom{$_{\pm.0000}$} & .9548\phantom{$_{\pm.0000}$} & \multicolumn{1}{c}{---} & \multicolumn{1}{c}{---} & .9551\phantom{$_{\pm.0000}$} & .9537\phantom{$_{\pm.0000}$} & .9542\phantom{$_{\pm.0000}$} & \textbf{.9554}\phantom{$_{\pm.0000}$} \\
in\_vehicle\_coupon\_recommendation & .8281$_{\pm.0033}$ & .8273$_{\pm.0046}$ & .8283$_{\pm.0051}$ & .8273$_{\pm.0046}$ & .8283$_{\pm.0051}$ & \multicolumn{1}{c}{---} & \multicolumn{1}{c}{---} & .8277$_{\pm.0035}$ & .8310$_{\pm.0044}$ & {\boldmath\textbf{.8325$_{\pm.0042}$}} & .8278$_{\pm.0052}$ \\
jm1 & .7444$_{\pm.0055}$ & .7404$_{\pm.0088}$ & {\boldmath\textbf{.7466$_{\pm.0051}$}} & .7404$_{\pm.0088}$ & {\boldmath\textbf{.7466$_{\pm.0051}$}} & \multicolumn{1}{c}{---} & \multicolumn{1}{c}{---} & .7347$_{\pm.0066}$ & .7330$_{\pm.0073}$ & .7306$_{\pm.0111}$ & .7397$_{\pm.0075}$ \\
kddcup09\_appetency & .8064$_{\pm.0072}$ & .8042$_{\pm.0161}$ & .8074$_{\pm.0088}$ & .8042$_{\pm.0161}$ & .8074$_{\pm.0088}$ & \multicolumn{1}{c}{---} & \multicolumn{1}{c}{---} & .8035$_{\pm.0111}$ & .8063$_{\pm.0187}$ & .7976$_{\pm.0140}$ & {\boldmath\textbf{.8209$_{\pm.0110}$}} \\
online\_shoppers\_intention & .9255$_{\pm.0052}$ & .9247$_{\pm.0066}$ & .9257$_{\pm.0059}$ & .9247$_{\pm.0066}$ & .9257$_{\pm.0059}$ & \multicolumn{1}{c}{---} & \multicolumn{1}{c}{---} & .9232$_{\pm.0070}$ & .9246$_{\pm.0066}$ & .9210$_{\pm.0068}$ & {\boldmath\textbf{.9260$_{\pm.0059}$}} \\
polish\_companies\_bankruptcy & {\boldmath\textbf{.9838$_{\pm.0039}$}} & .9537$_{\pm.0072}$ & .9621$_{\pm.0067}$ & .9537$_{\pm.0072}$ & .9621$_{\pm.0067}$ & \multicolumn{1}{c}{---} & \multicolumn{1}{c}{---} & .9533$_{\pm.0097}$ & .9577$_{\pm.0094}$ & .9555$_{\pm.0099}$ & .9593$_{\pm.0103}$ \\
qsar-biodeg & .9377$_{\pm.0146}$ & {\boldmath\textbf{.9384$_{\pm.0138}$}} & .9383$_{\pm.0123}$ & {\boldmath\textbf{.9384$_{\pm.0138}$}} & .9383$_{\pm.0123}$ & \multicolumn{1}{c}{---} & \multicolumn{1}{c}{---} & .9244$_{\pm.0206}$ & .9304$_{\pm.0158}$ & .9317$_{\pm.0115}$ & .9309$_{\pm.0152}$ \\
seismic-bumps & .7300$_{\pm.0546}$ & .7129$_{\pm.0508}$ & .7181$_{\pm.0375}$ & .7129$_{\pm.0508}$ & .7181$_{\pm.0375}$ & \multicolumn{1}{c}{---} & \multicolumn{1}{c}{---} & .7193$_{\pm.0326}$ & .7156$_{\pm.0400}$ & .7241$_{\pm.0453}$ & {\boldmath\textbf{.7374$_{\pm.0669}$}} \\
taiwanese\_bankruptcy\_prediction & .9413$_{\pm.0162}$ & .9348$_{\pm.0233}$ & .9381$_{\pm.0227}$ & .9348$_{\pm.0233}$ & .9381$_{\pm.0227}$ & \multicolumn{1}{c}{---} & \multicolumn{1}{c}{---} & .9363$_{\pm.0209}$ & .9416$_{\pm.0237}$ & {\boldmath\textbf{.9424$_{\pm.0265}$}} & .9421$_{\pm.0193}$ \\
\bottomrule
\end{tabular}
\\[2pt]
\parbox{\linewidth}{\footnotesize $^\dagger$ RF implementations of this method are not directly comparable. XGBoost 3.4.1 warns that \texttt{XGBRFClassifier} is deprecated and ``does not implement a conventional random forest''. CatBoost does not support Random Forest (Bagged) mode.\\
$^*$ Dynamic methods only benefit trees if they're trained deeper than the GBTs' depth 6. Benefit starts to show at depth 12-16, depending on dataset. As such, they've been excluded from the GBT results.}
\end{table*}

\subsection{\texttt{std::sort} and Highway}
\label{sec:stdsort-vs-highway}

\begin{figure}[b]
    \centering
    \includegraphics[width=0.478\textwidth]{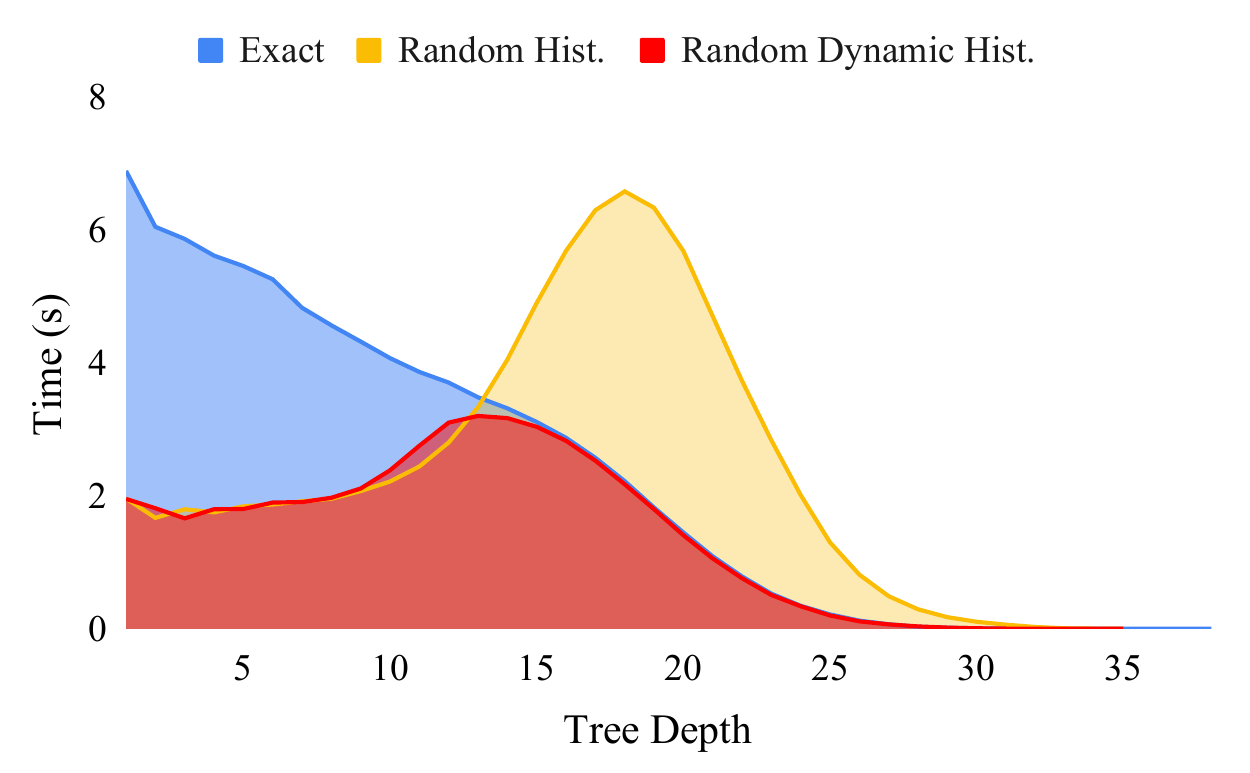}

    \caption{Training runtime by tree depth on a dataset with 1M samples 4096 features between \texttt{std::sort} and scalar Random Hist. Exact was reduced by 2-3$\times$ with the advent of Highway.}

%RB NO -- this kind of interpretation goes in the text.
    %Exact splitting has a clear disadvantage in runtime at higher levels, while histogramming is slower at mid-depth. This is because sorting time rises with the number of active samples, while histogram time tracks the number of active nodes.}
    \label{fig:per-depth-dynamic}
    \Description{Area chart showing the runtime of Exact (sorting-based) splitting and Histogramming methods, at specific tree depths}
\end{figure}

Dynamic histogramming was motivated by the YDF we started from, in which Exact splits sorted with \texttt{std::sort} and Random Hist.\ was the scalar \texttt{std::upper\_bound()} implementation. Figure~\ref{fig:per-depth-dynamic} compares the two in that setting, together with Dynamic switching between them at the per-node threshold of 1350 samples at which that pair breaks even. Sorting dominates at the top of the tree, where nodes are large; the scalar histogram dominates at the bottom, where thousands of small nodes each pay the fixed cost of setting up a histogram. Dynamic uses the histogram above the threshold and \texttt{std::sort} below it, and is the fastest of the three over the whole tree (single tree, single thread): $1.5\times$ faster than \texttt{std::sort} ($1.8\times$ on the split search alone) and $1.1\times$ faster than the scalar histogram, on both the dataset of Figure~\ref{fig:per-depth-dynamic} and HIGGS.

Highway VQSort changes this picture. It trains $2.2\times$ faster than \texttt{std::sort} and, unlike \texttt{std::sort}, it beats the scalar histogram at every depth, including the root node that holds every sample (Figure~\ref{fig:dynamic-motivation-nodetrain}). Dynamic over the scalar histogram therefore no longer wins in any regime: it is $1.45\times$ slower than Highway Exact over the whole tree. Recovering the histogram's advantage requires the vectorized histogram filling of the main text, which is why every speedup in this paper is measured against Highway Exact rather than \texttt{std::sort}.

\subsection{YDF Pull Request Contributions}
\label{app:pull-requests}

\textit{Due to anonymity concerns with exact Pull-request numbers and titles, we only provide a summary below.}

This project has resulted in numerous improvements to the YDF library, contributed as pull requests:

\begin{itemize}
    \item \textbf{Candidate-attribute sampling.} The eager per-node shuffle of all attributes is replaced by a lazy prefix Fisher--Yates shuffle that only draws the few positions actually tested, and is up to 75\% faster on high-dimensional data.
    \item \textbf{Projection-evaluator caching.} We replaced an $O(F)$ object re-initialization with a cached and reused object per thread. Up to 3$\times$ faster on high-dimensional data.
    \item \textbf{Random-histogram split search.} Invariant code hoisted out of the candidate-threshold scan - up to 30\% for sparse oblique speedup.
    \item \textbf{Vectorized bin generation.} \texttt{std::sort} is replaced by Highway VQSort in generating histogram bins; up to 15\%.
    \item \textbf{Histogram split search for oblique forests.} Adds the histogram (approximate) split finder to the sparse-oblique path, which previously supported only exact splits.
    \item \textbf{Equal-width binning.} Samples are assigned to equal-width bins with $O(1)$ arithmetic instead of a binary search over the bin boundaries.
    \item \textbf{Projection sampling.} The per-feature uniform draws that build each sparse projection are replaced by one Binomial draw for the number of non-zeros plus Floyd's sampler for their positions, reducing the cost from linear in the number of features to linear in the number of non-zeros (proof below).
\end{itemize}

\paragraph{Floyd's projection sampling.}
Originally, random sampling to build the projection matrix was a computational bottleneck for SO-YDF. This component comprised 80\% of the runtime, with the math-heavy parts of the code taking <20\%. The previous design assumed small $d$ and was inefficient for wide data.

Existing code calls Unif(0,1) over the projection matrix to create a mask, then populates this mask with weights. However, this requires $\Theta(np)$ calls to {\tt Unif($\cdot$)}, or $p$ calls per each projection, across $n$ projections. We show this can be substituted with a single call to Binomial($np, n\frac{k}{p}$), cutting runtime by 33\% for our settings, proportional to $\Theta(np)$.

Using a variant of an algorithm from Floyd~\cite{10.1145/30401.315746}, we form the projection matrix as follows:

\begin{lstlisting}[language=c++]
projection_density = k/p;
for projection in projections:  // O(n)
    for feature in features:    // O(p)
        if (Unif01(.) < projection_density) {...}
\end{lstlisting}

% Proof commented out for the submission (kept for the final version).
% Let $z^n$ denote the actual number of successes from running this algorithm, i.e. the number of times the inner loop's \texttt{if} condition evaluates to \texttt{True}. We need to show $z^n \sim \text{Binomial}(np,n\frac{k}{p})$:
%
% \begin{proof}
%
% \begin{enumerate}
%     \item The probability of one true evaluation (success) of \texttt{if (Unif01(.) < projection\_density)} $\sim \text{Bernoulli}(\frac{k}{p})$ since \texttt{projection\_density} $=\frac{k}{p}$, and a success only occurs when a Unif(0,1) sample is $< \frac{k}{p}$.
%     \item Let $z^i$ denote the number of successes for each projection, where $i\in [0,n-1]$. $z_i \sim \text{Binomial}(p,\frac{k}{p})$ since by definition, a Binomial($1,\frac{k}{p})=\text{Bernoulli}(\frac{k}{p})$, and by the additive property of the Binomial, $\sum_p \text{Binomial}(1,\frac{k}{p})=\text{Binomial}(p,k)$.
%     \item The same property extends to $z^n=\sum_n z_i=\text{Binomial}(np,nk)$.
% \end{enumerate}
%
% \end{proof}

% This reduces runtime by 30\% for target input parameters. The benefit is linear with the number of projections.
%We PR-ed this function into the YDF repo (PR \#199 - \textit{beware breaking anonymity}).

\end{document}